\documentclass[12pt]{article}

\usepackage[margin=1in]{geometry}
\usepackage{setspace}
\usepackage{graphicx}
\usepackage{subcaption}
\usepackage[
  backend=biber,
  authordate,
  cmsdate=16,
  uniquename=false,
  uniquelist=false
]{biblatex-chicago}

\usepackage{microtype}
\usepackage{hyperref}
\usepackage{booktabs} 
\usepackage{tikz}
\usepackage{comment}
\usetikzlibrary{arrows, positioning}
\usepackage{amssymb}
\usepackage[normalem]{ulem}

\newcommand{\notes}[1]{\par\smallskip
\begin{minipage}{\linewidth}\footnotesize\textit{Notes.}\ #1\end{minipage}}

\hypersetup{
  colorlinks=true,
  linkcolor=black,
  citecolor=black,
  urlcolor=black
}

\title{\textbf{
When F1 Is Not Enough: Auditing Codebooks for LLM-Assisted Annotation}}

\author{Yuehong Cassandra Tai\thanks{Corresponding Author, Assistant Research Professor, Center for Social Data Analytics, Pennsylvania State University, https://orcid.org/0000-0001-7303-7443, United States, \texttt{\href{mailto:yhcasstai@psu.edu}{yhcasstai@psu.edu}}. 
}
\and
Qunhua Li\thanks{Professor, Department of Statistics, Pennsylvania State University, http://orcid.org/0000-0003-0675-7648, United States, \texttt{\href{mailto:qul12@psu.edu}{qul12@psu.edu}}.
}
}

\date{}
\begin{document}
\doublespacing
\maketitle
\vspace{-1cm}
\begin{abstract}
Large language models (LLMs) are increasingly used for text annotation in social science, but standard performance metrics do not explain why errors occur. Low F1 may reflect limitations of the annotation system, defects in the written codebook, or inconsistency in reference labels. We propose a pre-deployment codebook audit for LLM-assisted annotation. The audit uses class-specific metrics and confusion matrices to identify problematic classes and boundaries, structured document-level review to attribute error sources, and targeted codebook revision followed by held-out evaluation. We apply the audit to two published multi-class codebooks from distinct domains. In both applications, the audit identifies under-specified or conflicting coding rules, and targeted revisions reduce the diagnosed errors on held-out data. Analyses with two LLM annotators show that some problematic boundaries and codebook revisions transfer across models, whereas other revision effects are annotator-specific. The results show that LLM annotation errors can help identify weaknesses in the written measurement instrument, not only in the model. Codebook auditing complements prompting, retrieval, and fine-tuning by clarifying whether persistent errors call for model improvement, codebook revision, or renewed scrutiny of reference labels.
\end{abstract}

\section{Introduction}
\label{sec:intro}

Large language models are becoming part of the measurement infrastructure of social science. Researchers increasingly use LLMs to annotate stance, frames, topics, attribution, and political events at a scale that would be costly or infeasible with manual coding alone. Recent methodological work has largely focused on how to obtain reliable model outputs: whether zero- or few-shot prompting is sufficient, when structured prompting or retrieval helps, and when supervised fine-tuning is worth the cost \parencite{gilardi2023chatgpt,Ornstein2025,Halterman2025codebook,Benoit2025,Griswold2025,Alizadeh2025}. Performance is typically evaluated against human-coded reference labels using F1 or related metrics.

This focus leaves an important diagnostic question underdeveloped. When an LLM disagrees with a reference label, what is the source of the disagreement? The model may fail to apply a clear rule; the written codebook may be incomplete, ambiguous, or internally inconsistent; or the released codebook may not support the reference label. Aggregate F1 cannot distinguish among these possibilities. It measures agreement with reference labels, not why disagreement occurs.

The distinction matters because the remedies differ. Prompt design, retrieval, calibration, or fine-tuning may help when the annotation system fails to implement a clear rule. If the written codebook does not state the operative criterion clearly, however, model-side changes may improve agreement with existing labels without resolving the underlying measurement problem. If the reference labels themselves are ambiguous or inconsistent, higher agreement may not imply better measurement. Codebook-based LLM annotation therefore requires evaluating the annotation system and the written instrument together.

We propose a pre-deployment codebook audit for LLM-assisted annotation. The audit treats the codebook and annotator as a \emph{codebook--annotator system}: observed errors depend on both the written operationalization and the model applying it. Class-specific metrics and confusion matrices identify weak classes and systematic confusions, and structured document-level review then distinguishes among LLM errors, repairable codebook defects, reference-label issues, and disagreements that would require a new substantive coding rule. Only repairable codebook defects motivate targeted revisions, which are evaluated on held-out data. The goal is not simply to increase F1, but to determine what an error pattern implies about implementation, codebook specification, and reference labels. 

We demonstrate the audit with two published multi-class codebooks from distinct substantive domains: the BFRS political-violence event taxonomy \parencite{bueno2015measuring,Halterman2025codebook} and the EU credit-claiming codebook \parencite{Hunter2025}. In both applications, targeted revisions reduce the diagnosed errors on held-out data. Analyses with Llama and Gemma further show that some problematic boundaries recur across LLMs and some revisions transfer across LLMs, while other revision effects depend on the annotator. 

This diagnostic contribution complements recent work showing that LLMs may rely on label semantics or internalized priors rather than supplied definitions and can be sensitive to codebook presentation \parencite{Halterman2025codebook,casanova2026,he2026better}. These studies examine whether models follow a supplied codebook; our audit adds a procedure for diagnosing whether persistent errors reveal a repairable problem in the written codebook itself and for evaluating targeted revisions. More broadly, the audit connects LLM evaluation to the construct-validity concern that measurement quality depends not only on predictive accuracy, but also on whether an operationalization represents the intended concept \parencite{adcock2001measurement}.

\section{Background and Related Work}
\label{sec:related}

Recent work shows that LLMs can perform many text-annotation tasks used in political science, with zero- or few-shot prompting sometimes achieving high agreement with expert or crowd labels \parencite{gilardi2023chatgpt,Ornstein2025,Benoit2025}. Performance, however, is sensitive to prompt design, task complexity, label semantics, and supervised examples \parencite{Halterman2025codebook,Alizadeh2025,Griswold2025}. This literature has clarified important model-side design choices and reinforced the need for task-specific validation against human labels \parencite{pangakis2025keeping,Tornberg2024B}.

Agreement with reference labels does not, however, reveal whether the model applied the written coding rules as intended. LLMs may rely on label names, surface cues, or familiar semantic associations rather than the supplied definitions \parencite{Halterman2025codebook}, model-internalized conceptual priors can limit how written definitions steer annotation behavior \parencite{casanova2026}, and richer guidance can improve accuracy without ensuring adherence to the supplied rules \parencite{he2026better}. A model can therefore achieve the same level of agreement either by supplying the written operationalization or by reaching the reference labels through other cues.

A separate problem can arise from the written codebook itself. Codebooks translate abstract constructs into observable coding rules and are therefore part of the measurement instrument, not merely an input to an LLM prompt. Human coders often learn difficult distinctions through training, examples, discussion, feedback, and adjudication, establishing conventions that are understood by the original coding team but are not fully recorded in the written document. Independent annotators---and especially a zero-shot LLM---must rely much more heavily on what the written document actually specifies. Research on annotation practice similarly shows that unclear or incomplete guidelines can impede consistent coding \parencite{rottger2022}, while guideline development and annotation quality management are often under-reported \parencite{klie2024analyzing}. Such omissions can have substantive consequences because incomplete conceptualization can generate downstream bias that better scoring or statistical correction cannot repair \parencite{halterman2026protest}.

The credit-claiming codebook provides a concrete example. \textcite{Hunter2025} reports Krippendorff's $\alpha=0.78$ after multiple rounds of coder training and codebook refinement, whereas independent coders applying the released codebook achieve approximately $\alpha=0.43$
\parencite{Yang2025}. This difference suggests that some coding boundaries may depend on training, discussion, or conventions established during the original coding process that are not fully recoverable from the released text.

Reference labels create a further complication. Evaluation against human-coded labels is necessary for measuring agreement, but it also treats those labels as the target the model should reproduce. In difficult cases, however, released reference sets may contain annotation errors or reflect coding conventions that are not recoverable from the written rules, and high inter-annotator agreement does not by itself establish that every individual label is supported by the codebook \parencite{klie2024analyzing}. 

A model--reference-label mismatch can therefore arise because the model fails to apply a clear rule, because the written codebook does not specify the relevant distinction, or because the reference label is ambiguous or inconsistent with the released coding materials. Class-specific precision, recall, and confusion matrices can locate where such disagreements concentrate, but they cannot determine whether the written codebook itself warrants repair. We therefore introduce a pre-deployment codebook audit that uses localized error patterns and document-level review to identify repairable defects in the written rules, distinguish them from patterns that should not trigger codebook revision, and evaluate targeted revisions on held-out data.

\section{Method}
\label{sec:pipeline}

\subsection{Codebook Audit Pipeline}

Let $\mathcal{Y}=\{1,\ldots,K\}$ denote the set of coding categories, and let
$C^{(0)}=\{c^{(0)}_1,\ldots,c^{(0)}_K\}$ denote the original codebook, where
$c^{(0)}_k$ contains the written definition, inclusion criteria, exclusion criteria, and examples
for class $k$. We assume access to a labeled development set
\[
D_{\mathrm{dev}}=\{(x_i,y_i)\}_{i=1}^{n_{\mathrm{dev}}},
\]
and a disjoint held-out validation set,
\[
D_{\mathrm{test}}=\{(x_i,y_i)\}_{i=1}^{n_{\mathrm{test}}}.
\]
The development set is used in Steps~1--3 to identify the classes and boundaries that warrant review and to specify the planned revision sequence. The disjoint held-out set is used only to evaluate the original and successive revised codebooks. Held-out results do not determine revision targets, their order, or the wording of subsequent revisions. 

Given a codebook $C$, an LLM annotator produces a predicted label
\[
\hat y_i = f_\theta(x_i; C, s),
\]
where $f_\theta$ denotes the model and annotation prompt, and $s$ denotes the annotation setting, including decoding parameters and, when applicable, retrieval. These settings are held fixed throughout the audit, so that changes in performance can be attributed to changes in the codebook rather than changes in the model pipeline.

For many-class codebooks, the prompt may include a retrieval step. Let
\[
R_m(x_i,C)\subset C
\]
denote the $m$ codebook entries retrieved for document $x_i$. The annotator then conditions on $R_m(x_i,C)$ rather than the full codebook. For smaller codebooks, the full codebook is included in the prompt. In either case, the model returns a predicted label and, optionally, a short rationale. 

\begin{figure}
\includegraphics[width=1\textwidth]{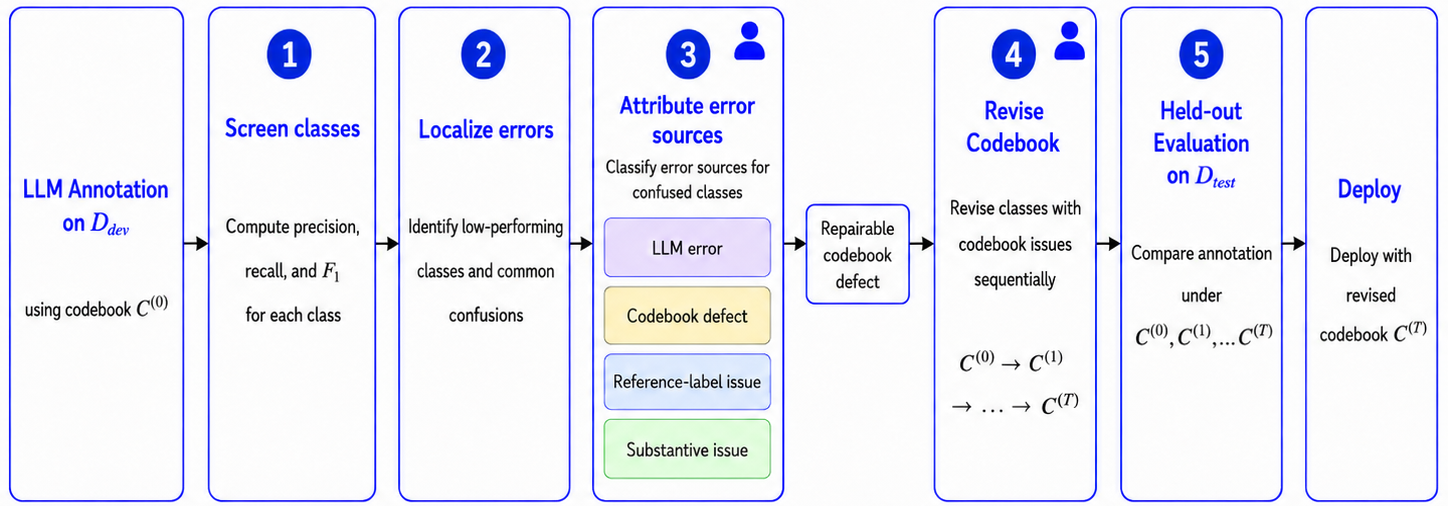}
\caption{\textbf{Codebook audit pipeline.}}
\notes{Steps 1–3 diagnose error patterns on the development set; Step 4 revises only repairable codebook defects; Step 5 evaluates the prespecified revisions on held-out data. Human icons indicate researcher judgment. Held-out performance does not determine subsequent revisions.}
\label{fig:pipeline}
\end{figure}

The audit proceeds in five steps: (1) screen classes, (2) localize errors, (3) attribute error sources, (4) revise the diagnosed codebook entries sequentially, and (5) evaluate each resulting codebook version on held-out data (Figure~\ref{fig:pipeline}). Steps~1--3 form a diagnostic stage conducted on $D_{\mathrm{dev}}$ using the original codebook. This stage identifies the classes and boundaries that warrant review and determines the disposition of the corresponding error patterns. Patterns attributed to LLM error, reference-label ambiguity or inconsistency, or a substantive judgment not specified by the existing operationalization are documented but do not proceed to revision. Patterns attributed to recurring, repairable codebook defects become revision targets in Step~4. When more than one target is identified, the revisions are implemented sequentially. Step~5 evaluates the original and successive revised codebooks on $D_{\mathrm{test}}$; these held-out results assess generalization but do not feed back into the diagnostic or revision decisions.

\subsubsection{Step 1: Screen Classes}

We first run the annotator on $D_{\mathrm{dev}}$ using the original codebook $C^{(0)}$. For each class $k$, we compute precision, recall, and F1:
\[
P_k=\frac{\sum_i \mathbf{1}\{\hat y_i=k,\,y_i=k\}}{\sum_i \mathbf{1}\{\hat y_i=k\}},\qquad
R_k=\frac{\sum_i \mathbf{1}\{\hat y_i=k,\,y_i=k\}}{\sum_i \mathbf{1}\{y_i=k\}},
\]
and $F1_k=2P_kR_k/(P_k+R_k)$. Uncertainty is quantified with a nonparametric document-level bootstrap.

To make screening reproducible, we use two pre-specified signals on the original-codebook development results. For the \emph{level} signal, let $k_{(1)}$ and $k_{(2)}$ denote the classes with the lowest and second-lowest F1. The lowest-F1 class is nominated when the paired bootstrap 95\% confidence interval for $F1_{k_{(2)}}-F1_{k_{(1)}}$ lies above zero. For the \emph{asymmetry} signal, define $A_k=P_k-R_k$ and let $k_A=\arg\max_k |A_k|$. Class $k_A$ is nominated when the bootstrap 95\% confidence interval for $A_{k_A}$ excludes zero. Positive $A_k$ indicates relatively low recall and under-prediction; negative $A_k$ indicates relatively low precision and over-prediction. If the same class triggers both signals it is reviewed once; exact ties are retained. The intervals are used as screen criteria for prioritizing diagnostic review rather than testing a substantive hypothesis.

\subsubsection{Step 2: Localize Errors}

For classes nominated in Step~1, we inspect the confusion matrix
\[
M_{ab}
=
\sum_i \mathbf{1}\{y_i=a,\,\hat y_i=b\},
\]
where rows are reference labels and columns are model predictions. We use the off-diagonal entries to identify three patterns: over-predicted classes with many false positives, under-predicted classes with many false negatives, and class pairs with frequent bidirectional confusion. These classes and class pairs are carried forward to document-level review in Step~3. We pay particular attention to fallback categories because classes defined largely by exclusion can be over-predicted, under-predicted, or both when their assignment conditions are not explicit.

\subsubsection{Step 3: Attribute Error Sources}

For each major error pattern carried forward from Step~2, we initially review five affected documents and expand the sample to as many as ten when the source remains mixed or unclear. Each case is considered together with the document text, the relevant codebook entries, the reference label, and the model prediction. Model rationales, when available, are treated only as supplementary information about model behavior.

Attribution follows a fixed four-way decision procedure. The reviewer first asks whether the written codebook, interpreted according to its stated definitions and coding guidance, provides a sufficiently clear basis for assigning the document. If the written rule clearly supports the reference label but not the model prediction, the case is classified as an \emph{LLM error}. If the written rule clearly supports the model prediction rather than the reference label, or if substantively similar documents receive inconsistent reference labels, the case is classified as a \emph{reference-label issue}. If the written entries are incomplete, ambiguous, or internally inconsistent in a way that can be addressed by clarifying the existing operationalization, the case is classified as a \emph{codebook defect}. If resolving the disagreement would instead require introducing a new substantive criterion not determined by the existing construct definition, the case is classified as a \emph{substantive issue} rather than as a repairable wording defect.

The attribution determines the disposition of each diagnosed error pattern. A pattern becomes a revision target only when review identifies a recurring, repairable problem in the written rules; in our implementation, the same explicit clause-level problem must recur in a majority of the reviewed cases. Patterns better explained by LLM error, reference-label ambiguity or inconsistency, substantive judgment, or mixed evidence are documented without codebook revision. Appendix~\ref{app:stoprule} gives the detailed disposition rules and examples from the two applications.

\subsubsection{Step 4: Revise the Codebook}

Using the diagnostic results from Steps~1--3, we revise only entries associated with repairable codebook defects. The revision targets and their substantive rationale are fixed by the development-stage diagnosis. When more than one entry or boundary is implicated, we implement the revisions sequentially so that the effect of each clarification can be examined separately. This produces a revision sequence
\[
C^{(0)} \rightarrow C^{(1)} \rightarrow \cdots \rightarrow C^{(T)},
\]
where $C^{(0)}$ is the original codebook and each $C^{(t)}$ incorporates one additional targeted revision. Previously adopted revisions are retained as the next diagnosed entry or boundary is changed.

We also assess whether the observed pattern depends on semantic class names by rerunning annotation with all class names replaced by anonymous labels, such as $L001,\ldots,L00K$, while leaving the written definitions and coding guidance unchanged. Let $F1^{\mathrm{text}}_k$ and $F1^{\mathrm{anon}}_k$ denote class-$k$ F1 under semantic and anonymous labels:
\[
\Delta^{\mathrm{name}}_k
=
F1^{\mathrm{text}}_k-F1^{\mathrm{anon}}_k.
\]
A positive value indicates better performance when the semantic class name is available, whereas a negative value indicates better performance when it is removed. We use this ablation as a diagnostic of label-name dependence and as a signal that the class name may warrant substantive review; it does not determine revision targets and is not, by itself, evidence that a name is conceptually appropriate or inappropriate.

Each written revision targets the specific defect identified during the development-stage diagnosis. When possible, the revised wording states affirmative assignment conditions, makes relevant inclusion and exclusion boundaries explicit, and resolves inconsistencies among definitions, clarifications, and examples. Each change is logged with the original wording, revised wording, rationale, targeted error pattern, and expected effect. Steps~1--3 are not rerun after each revision; the sequential revisions implement the targets identified in the initial diagnostic stage.

\subsubsection{Step 5: Evaluate on Held-Out Data}

The held-out sample $D_{\mathrm{test}}$ is used to evaluate whether the targeted codebook revisions identified on $D_{\mathrm{dev}}$ generalize to previously unseen documents. We evaluate the original codebook and each successive revised version,
\[
C^{(0)},C^{(1)},\ldots,C^{(T)},
\]
under the same model, prompt, retrieval configuration, decoding parameters, and evaluation metrics. For each transition from $C^{(t-1)}$ to $C^{(t)}$, we examine changes in class-specific precision, recall, and F1, with particular attention to the error pattern targeted by that revision. Reporting the full sequence shows whether the expected effect of each clarification also appears out of sample. For the overall effect of the completed audit, we additionally compare the original and final codebooks:
\[
\Delta F1_k=F1_k(C^{(T)})-F1_k(C^{(0)}).
\]
Uncertainty is quantified with a paired document-level bootstrap using the same resampled documents under the codebook versions being compared.

Held-out performance is evaluative rather than diagnostic. In our applications it is examined after each planned update, but it is not used to add or remove revision targets, change their order, alter subsequent wording, or rerun Steps~1--3. The revision path is determined by the development-stage diagnosis and substantive review. Appendix~\ref{app:power} discusses considerations for planning the size of the held-out evaluation sample.

\section{Setting and Data}
\label{sec:setting}

\subsection{Data and Case Selection}
\label{sec:data}

We apply the audit to two published multi-class codebooks. The BFRS codebook \parencite{bueno2015measuring}, compiled for LLM annotation by \textcite{Halterman2025codebook}, classifies short news reports about political violence in Pakistan into twelve event types, including the fallback category \textsc{Other}. The credit-claiming codebook \parencite{Hunter2025} classifies statements about European Council summit outcomes into four categories: \textsc{Credit-claiming}, \textsc{Credit-sharing}, \textsc{Blame-shifting}, and \textsc{Descriptive}. 

The two applications provide contrasting annotation settings. \textsc{Other} among twelve categories accounts for $6.5\%$ of BFRS corpus, whereas \textsc{Descriptive} among four categories accounts for $60.5\%$ of the credit-claiming corpus. The cases therefore differ in both the complexity of the label space and the prevalence of the fallback category, but both provide detailed coding guidances that support document-level diagnostic review.

For each codebook, we draw a class-stratified development sample and a disjoint held-out  sample. For BFRS, $D_{\mathrm{dev}}$ contains 573 documents and $D_{\mathrm{test}}$ contains 1,045; for credit claiming, the corresponding samples contain 172 and 321 documents. The held-out samples target approximately 100 documents per class where the corpus limits. They combine two nonoverlapping class-stratified draws from the same held-out pool: an initial draw of 50 documents per class and a second disjoint draw added to improve evaluation precision. We analyze the two draws together as a single $D_{\mathrm{test}}$. We also use a 20-per-class subsampling of the development data to assess the stability of nomination and localization decisions. Appendix~\ref{app:sampling} provides sampling details.

\subsection{Model Setup}
\label{sec:setting-models}

Both codebooks are annotated with the same zero-shot model, Llama-3.3-70B-Instruct, using deterministic decoding. Original and revised codebooks are evaluated under identical settings within each application. BFRS uses retrieval-augmented generation because its twelve categories contain longer class-specific guidance, while the four-class credit-claiming codebook is included in full in the prompt. Appendix~\ref{app:annotator} provides the full model details. Claude (Anthropic, Claude Code, Opus 5) was used to assist with generating and refining code used to produce figures; all code and outputs were reviewed by the authors.

\section{Results}
\label{sec:results}

We apply the audit to the BFRS event codebook and the credit-claiming codebook. For each case, we report  the development-stage diagnosis, the resulting codebook revision, and its evaluation on held-out data. We then examine the stability of the diagnostic patterns across samples and LLM annotators.

\subsection{Case 1: BFRS Codebook}
\label{sec:results-bfrs}

\paragraph{LLM annotation and class-wise performance.}
The model annotates $D_{\mathrm{dev}}$ using the original BFRS codebook $C^{(0)}$ with retrieval. Performance varies substantially across the twelve classes. \textsc{Other} has the lowest F1 ($0.404$), followed by \textsc{Gov\_Attack\_on\_Civilian} (F1 $=0.597$). \textsc{Gov\_Attack\_on\_Civilian} also has the strongest precision--recall asymmetry (precision $=0.85$, recall $=0.46$). These patterns nominate \textsc{Other} through the level screen and \textsc{Gov\_Attack\_on\_Civilian} through the asymmetry screen. The paired-bootstrap 95\% confidence interval for the F1 gap between the second-lowest class and \textsc{Other} lies above zero, and the interval for \textsc{Gov\_Attack\_on\_Civilian}'s asymmetry excludes zero. Full results are reported in Appendix~Table~\ref{tab:per-class-f1} to~\ref{tab:bfrs-nomination-screen}.

\begin{figure}[!htbp]
    \centering
    \includegraphics[width=0.9\linewidth]{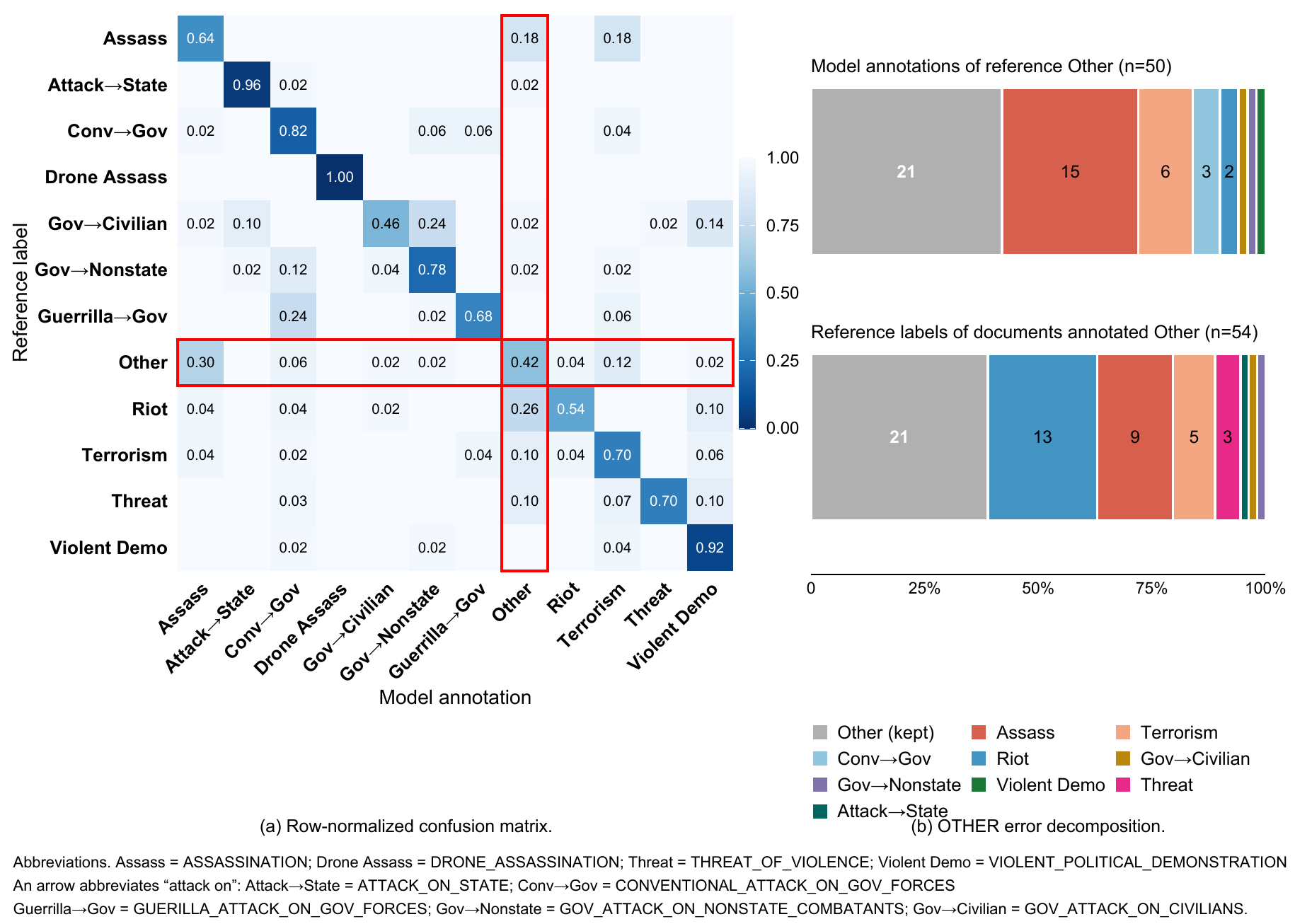}
    \caption{\textbf{BFRS diagnosis under the original codebook on $D_{\mathrm{dev}}$}}
    \label{fig:bfrs-diagnosis}
    \notes{Development sample (n = 573). In panel (a), diagonal entries are class-specific recall and shares below $0.01$ are omitted. In Panel (b), the top bar shows model annotations for documents with reference label \textsc{Other} and the bottom bar shows the reference-label composition of documents annotated as \textsc{Other}.}
\end{figure}

\paragraph{Problematic classes and common confusions.}
Figure~\ref{fig:bfrs-diagnosis} summarizes the BFRS error pattern. Only $42\%$ of reference \textsc{Other} cases are predicted correctly. The remaining cases are assigned to several substantive categories, most often \textsc{Assassination} ($30\%$), \textsc{Terrorism} ($12\%$), and \textsc{Conventional\_Attack} ($6\%$). Errors also occur in the opposite direction: reference \textsc{Riot}, \textsc{Assassination}, and \textsc{Terrorism} cases are sometimes predicted as \textsc{Other}. Thus, the \textsc{Other} errors are distributed across several neighboring substantive classes rather than concentrated in a single confusion.

The next-lowest recall class is \textsc{Gov\_Attack\_on\_Civilian}: $46\%$ of its reference cases are predicted correctly. Its largest confusion is with \textsc{Gov\_Attack\_on\_Nonstate} ($24\%$), followed by \textsc{Violent\_Political\_Demonstration} ($14\%$) and \textsc{Attack\_on\_State} ($10\%$). These errors concentrate in substantively adjacent state-violence categories.

\paragraph{Error-source attribution.}
Review of the \textsc{Other} errors identifies two codebook defects. First, the original \textsc{Other} entry defines the class mainly by exclusion: it covers incidents that are ``not easily categorized'' or whose source of violence is unclear, and directs annotators to use \textsc{Other} when an event ``does not fit into the other categories.'' The entry therefore provides few positive conditions for assigning the fallback class, consistent with true \textsc{Other} cases being pulled into more specific categories.

Second, the \textsc{Other}--\textsc{Assassination} boundary is underspecified. Although the \textsc{Assassination} definition requires deliberate targeting, its clarification states that the label applies when ``a body is found with bullet wounds or other marks of violence and/or torture.'' Without an explicit targeting condition, unexplained killings can be assigned to \textsc{Assassination} even when the report does not establish that the victim was deliberately singled out.

The review also explains why other candidate boundaries are not revised. Many \textsc{Riot} cases predicted as \textsc{Other} reflect the same open-ended fallback rule rather than a defect in the \textsc{Riot} entry. For \textsc{Gov\_Attack\_on\_Civilian}, sampled errors instead reveal ambiguous reference cases: near-identical reports of security forces destroying suspected militants' houses or hideouts receive different reference labels, sometimes as \textsc{Gov\_Attack\_on\_Civilian} and sometimes as \textsc{Gov\_Attack\_on\_Nonstate} (See examples in Appendix~\ref{app:goldcases}). This pattern is also consistent with the manual error analysis in \textcite{Halterman2025codebook}. We therefore carry only the \textsc{Other} and \textsc{Assassination} entries forward to revision.

\paragraph{Codebook revision.}
The label-name ablation first assesses whether performance depends on the \textsc{Other} name. F1 is higher with the semantic name than with an anonymous label (Appendix~Figure~\ref{fig:label-name-ablation}; Appendix ~\ref{app:label-name-ablation}), providing no evidence that the observed error pattern is caused by a misleading class name. We therefore retain the name and focus the revision on the written assignment rules.

The first revision, $C^{(1)}$, changes only \textsc{Other}. It replaces the open fallback rule with two affirmative assignment conditions: the report lacks enough information about perpetrator and target to assign a substantive class, or the event falls outside all substantive categories. It also states that \textsc{Other} should not be used when a substantive class plausibly applies.

\begin{figure}[!htbp]
\centering
\begin{subfigure}[t]{0.54\textwidth}
    \centering
    \includegraphics[width=\linewidth]{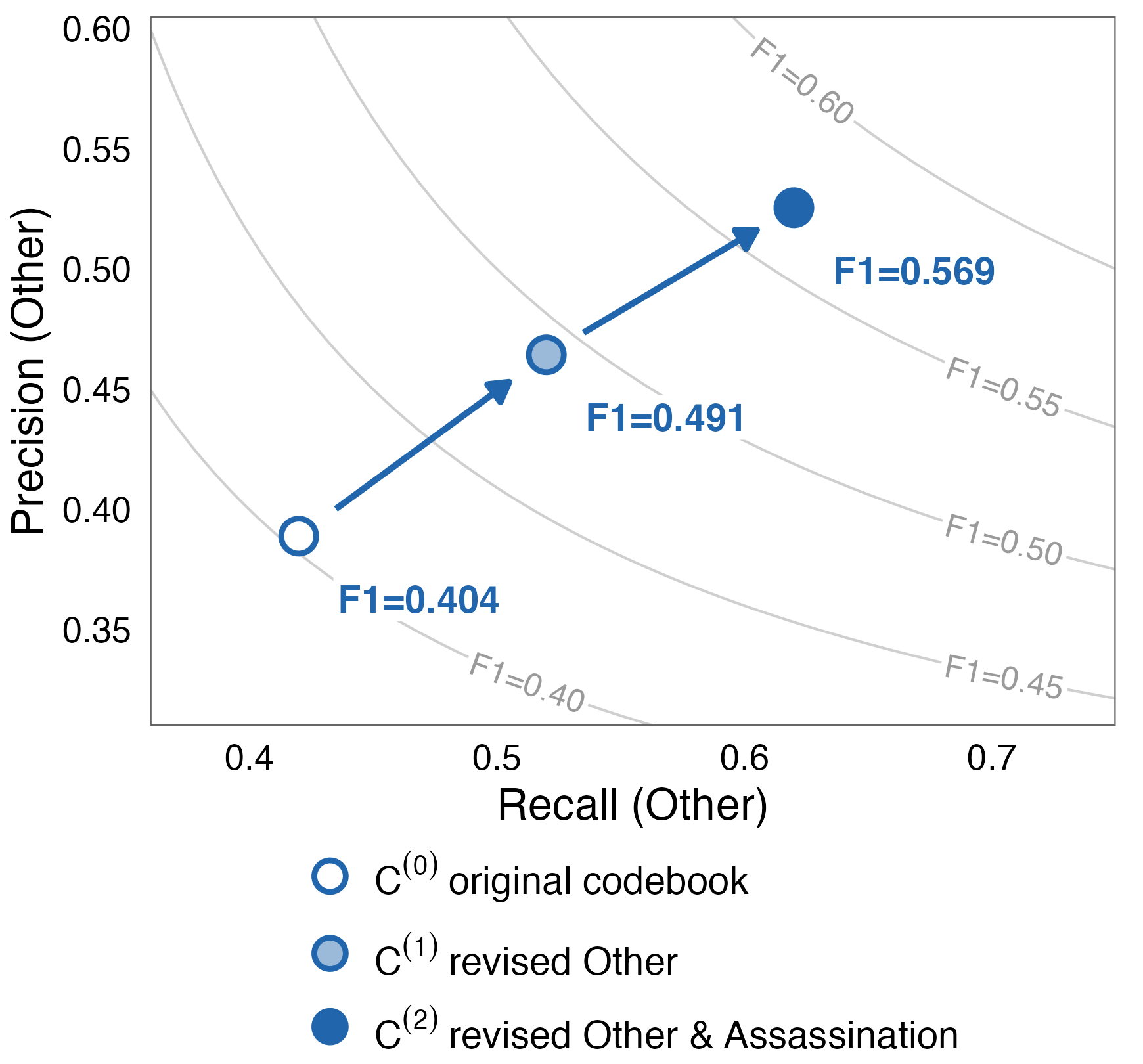}
    \caption{Precision--recall trajectory for \textsc{Other} on $D_{\mathrm{dev}}$.}
\end{subfigure}\hfill
\begin{subfigure}[t]{0.42\textwidth}
    \centering
    \includegraphics[width=\linewidth]{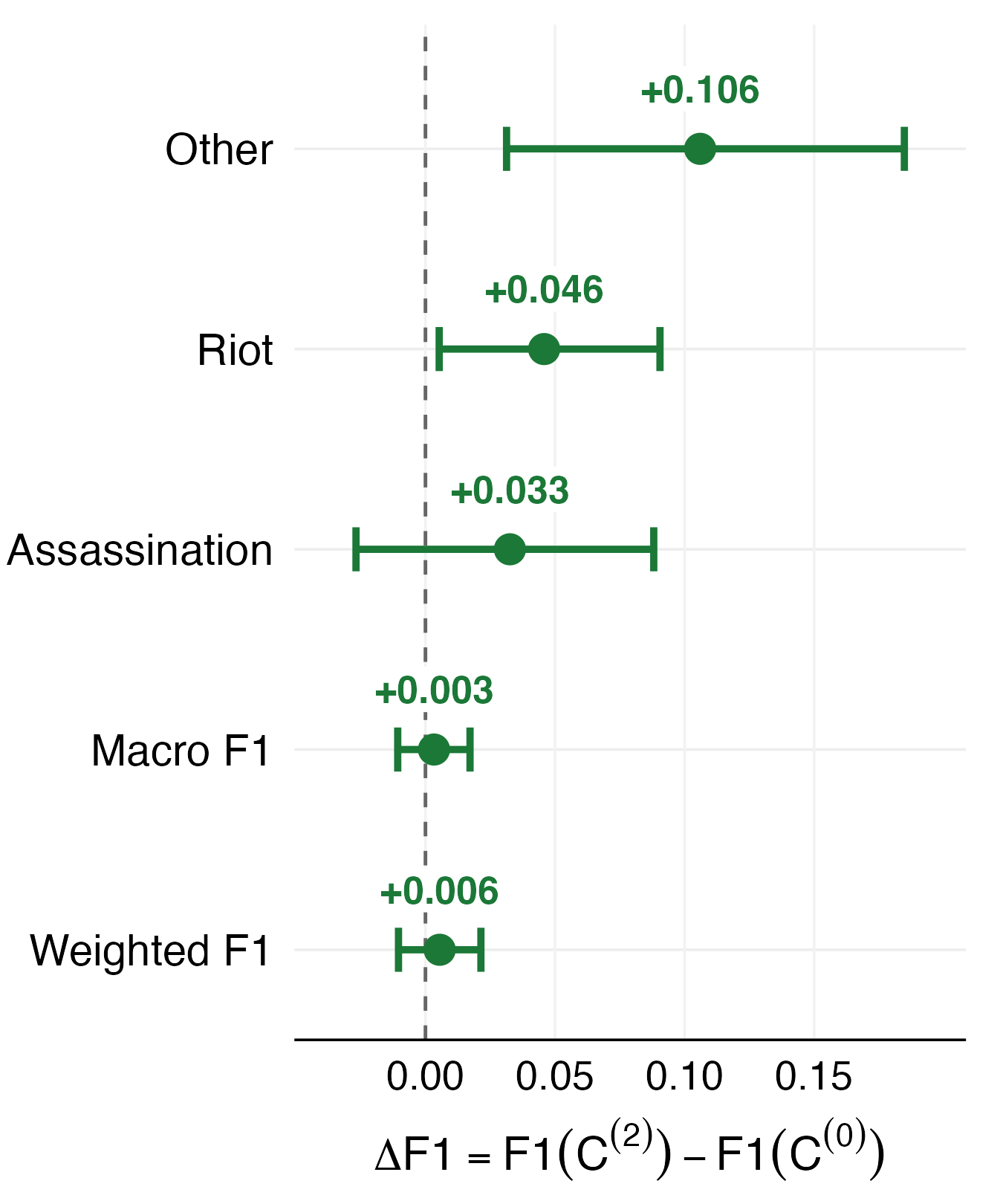}
    \caption{Held-out $\Delta F1$ with paired-bootstrap 95\% CIs.}
\end{subfigure}
\caption{\textbf{BFRS revision trajectory and held-out evaluation.}}
\notes{Panel (a) shows precision and recall for \textsc{Other} on $D_{\mathrm{dev}}$ (n = 573). $C^{(0)}$ is the original codebook, $C^{(1)}$  revises \textsc{Other}, $C^{(2)}$ revises \textsc{Other} plus \textsc{Assassination}. Gray curves are constant-F1 contours, and arrows trace the revision sequence. Panel (b) reports held-out $\Delta F1=F1(C^{(2)})-F1(C^{(0)})$ ($n=1{,}045$), with paired-bootstrap 95\% confidence intervals ($B=5{,}000$).}
\label{fig:bfrs-revision-eval}
\end{figure}

The second revision, $C^{(2)}$, retains the revised \textsc{Other} entry and adds an explicit targeting condition to the \textsc{Assassination} clarification. A body found with signs of violence is coded as \textsc{Assassination} only when the report indicates that the victim was deliberately singled out; otherwise, unexplained cases without evidence of responsibility or targeting are directed to \textsc{Other}. Full wording is provided in Appendix~\ref{app:revised-codebook}. Steps~1--3 had identified these two entries as the repairable BFRS targets; \textsc{Gov\_Attack\_on\_Civilian} was documented as a reference-label issue rather than a codebook defect. We therefore implement the two planned revisions sequentially, yielding $C^{(1)}$ and $C^{(2)}$.
\newpage
\paragraph{Held-out evaluation.}
We evaluate $C^{(0)}$, $C^{(1)}$, and $C^{(2)}$ under the same Llama configuration. Figure~\ref{fig:bfrs-revision-eval} separates the development-stage revision trajectory from evaluation on the held-out validation sample (Numerical results in Appendix~Table~\ref{tab:bfrs-aggregate}).

On $D_{\mathrm{dev}}$, \textsc{Other} F1 increases from $0.40$ under $C^{(0)}$ to $0.49$ under $C^{(1)}$ and $0.57$ under $C^{(2)}$, driven primarily by higher recall. \textsc{Assassination} F1 increases from $0.62$ to $0.69$. Although \textsc{Riot} is never edited, its F1 increases from $0.67$ to $0.76$, and the number of \textsc{Riot} cases predicted as \textsc{Other} falls from 13 to 6. The 95\% confidence intervals for the gains in \textsc{Other}, \textsc{Riot}, macro F1, and weighted F1 exclude zero.

The targeted pattern also improves on the held-out sample. From $C^{(0)}$ to $C^{(2)}$, F1 increases from $0.429$ to $0.535$ for \textsc{Other} ($\Delta=+0.106$, 95\% CI $[+0.031,+0.185]$) and from $0.725$ to $0.771$ for \textsc{Riot} ($\Delta=+0.046$, 95\% CI $[+0.005,+0.090]$). \textsc{Assassination} also increases from $0.746$ to $0.779$, although its confidence interval includes zero. Macro and weighted F1 change little, consistent with a targeted boundary revision rather than a broad shift in the taxonomy (Appendix Table ~\ref{tab:bfrs-aggregate}). Overall, the BFRS case shows that clarifying a fallback class and one neighboring boundary improves the intended error pattern on unseen data without revising every low-performing class.

\subsection{Case 2: Credit-Claiming Codebook}
\label{sec:results-credit}

\paragraph{LLM annotation and class-wise performance.}
Under the original codebook $C^{(0)}$, class F1 values range from approximately $0.67$ to $0.75$, so the lowest-F1 class is not clearly separated from the others. \textsc{Descriptive}, however, has the largest absolute precision--recall asymmetry: precision is $0.59$ and recall is $0.82$, with its 95\% confidence interval excluding zero (Appendix Table~\ref{tab:credit-nomination-screen}). \textsc{Descriptive} is therefore nominated through the asymmetry screen.

\begin{figure}[!htbp]
\centering
\begin{subfigure}[t]{0.52\textwidth}
    \centering
    \includegraphics[width=\linewidth]{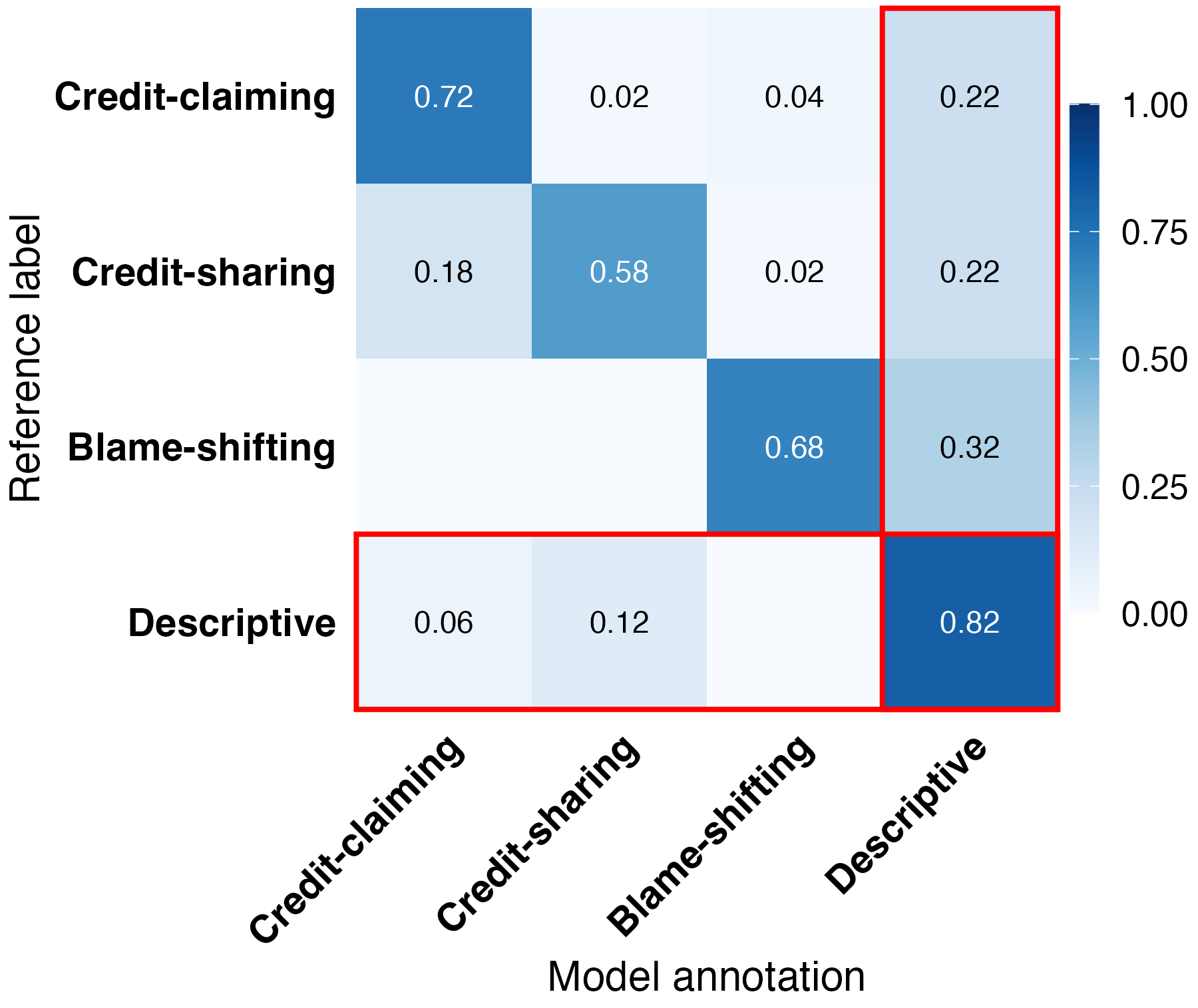}
    \caption{Row-normalized confusion matrix.}
\end{subfigure}\hfill
\begin{subfigure}[t]{0.44\textwidth}
    \centering
    \includegraphics[width=\linewidth,trim = 0cm 0cm 0cm 0.8cm, clip=true]{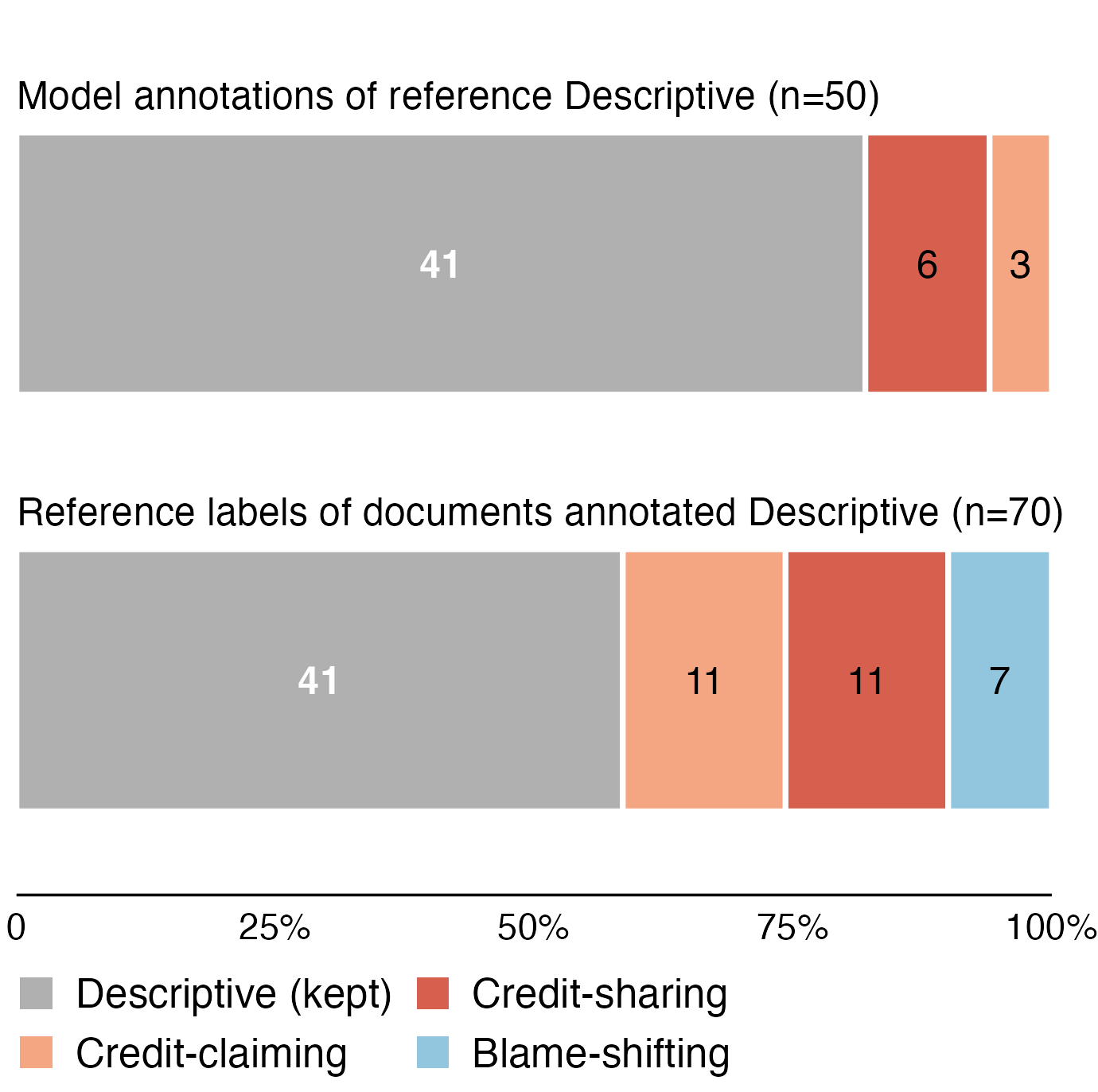}
    \caption{\textsc{Descriptive} error decomposition.}
\end{subfigure}
\caption{\textbf{Credit-claiming diagnosis under the original codebook on $D_{\mathrm{dev}}$.}}
\notes{Development sample ($D_{\mathrm{dev}}$, n = 172). In panel (a), diagonal entries are class-specific recall and shares below $0.01$ are omitted. In panel (b), the top bar shows model annotations for documents with reference label \textsc{Descriptive} and the bottom bar shows the reference-label composition of documents annotated as \textsc{Descriptive}.}
\label{fig:hunter-diagnosis}
\end{figure}

\paragraph{Problematic classes and common confusions.}
Figure~\ref{fig:hunter-diagnosis} shows that \textsc{Descriptive} is over-predicted. Of 50 reference \textsc{Descriptive} cases in $D_{\mathrm{dev}}$, 41 are predicted correctly. However, among the 70 cases predicted as \textsc{Descriptive}, 29 are reference-labeled as \textsc{Credit-claiming}, \textsc{Credit-sharing}, or \textsc{Blame-shifting}. Thus, the main error is assignment of substantive attribution statements to the fallback class.

\paragraph{Error-source attribution.}
Document review traces this pattern to conflicting guidance for future-oriented statements. \textsc{Descriptive} is defined as containing no credit or blame attribution, but one clarification includes desired actions or outcomes, such as statements that Europe ``must'' or ``should'' do something. Elsewhere, the coding guidance allows both prospective and retrospective attribution. A future-oriented statement that credits a specific actor can therefore satisfy competing rules: its modality points toward \textsc{Descriptive}, while its attribution points toward \textsc{Credit-claiming} or \textsc{Credit-sharing}.

\paragraph{Codebook revision.}
The label-name ablation shows little change when class names are anonymized (Appendix~Figure~\ref{fig:label-name-ablation}), providing little evidence that label-name semantics drive the observed confusion. We therefore retain the class names and revise the written clarification. The revised rule clarifies that future-oriented language does not by itself make a statement \textsc{Descriptive}. A ``must'' or ``should'' statement is \textsc{Descriptive} only when it credits no specific actor. Statements crediting the speaker's own government or country are \textsc{Credit-claiming}, and those crediting the EU, EU institutions, or partner governments are \textsc{Credit-sharing}. The remaining entries are unchanged. Because the diagnosis identified no other recurring clause-level problem that could be clarified without imposing a new substantive rule, $C^{(1)}$ is the sole planned revision (See full wording in Appendix~\ref{app:revised-codebook}).

\begin{figure}[!htbp]
\centering
\begin{subfigure}[t]{0.5\textwidth}
    \centering    \includegraphics[width=\linewidth]{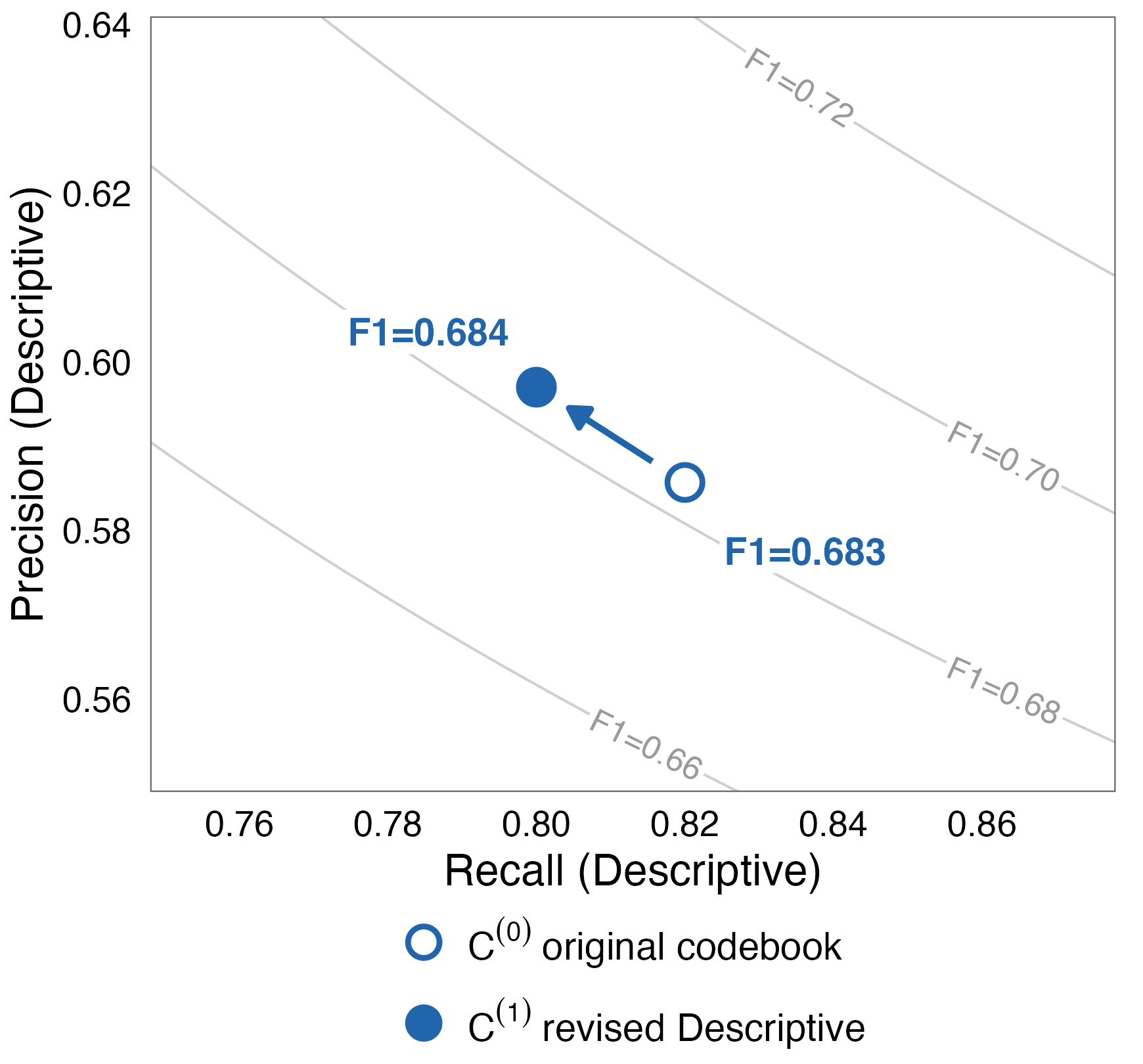}
    \caption{Precision--recall tradeoff for \textsc{Descriptive} on $D_{\mathrm{dev}}$.}
\end{subfigure}\hfill
\begin{subfigure}[t]{0.42\textwidth}
    \centering
    \includegraphics[width=\linewidth]{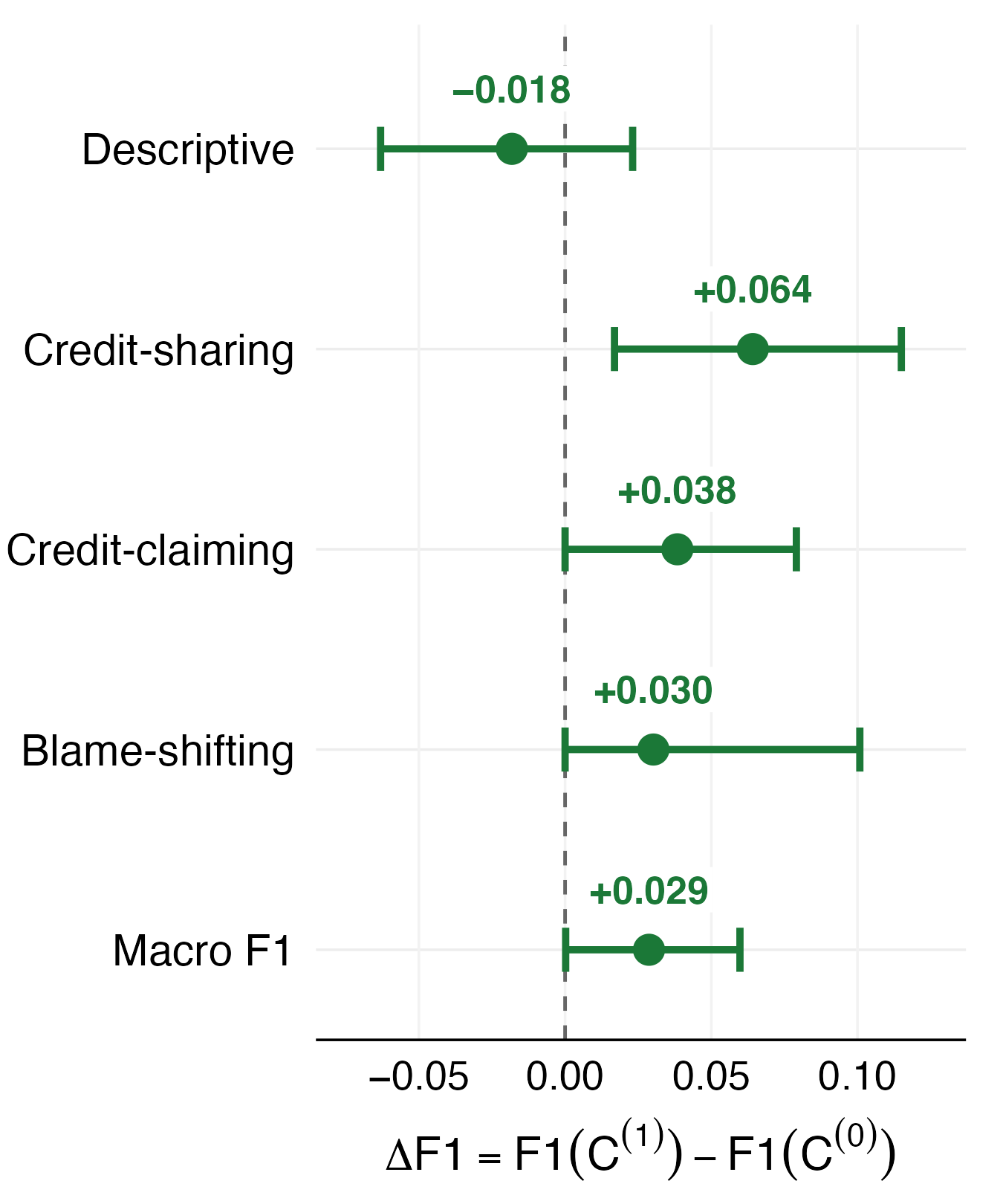}
    \caption{Held-out $\Delta F1$ with paired-bootstrap 95\% CIs.}
\end{subfigure}
\caption{\textbf{Credit-claiming revision trajectory and held-out evaluation.}}
\notes{Panel (a) shows precision and recall for \textsc{Descriptive} on $D_{\mathrm{dev}}$ (n = 172). $C^{(0)}$ is original codebook and $C^{(1)}$ revises \textsc{Descriptive}. Gray curves are constant-F1 contours, and the arrow traces the revision. Panel (b) reports held-out $\Delta F1=F1(C^{(1)})-F1(C^{(0)})$ ($n=321$), with paired-bootstrap 95\% confidence intervals ($B=5{,}000$).}
\label{fig:hunter-revision-eval}
\end{figure}

\paragraph{Held-out evaluation.}
We evaluate $C^{(0)}$ and the revised codebook $C^{(1)}$ under the same annotation configuration (Figure~\ref{fig:hunter-revision-eval}; Appendix~Table~\ref{tab:hunter-aggregate}). As expected, narrowing \textsc{Descriptive} trades recall for precision and reduces the number of substantive attribution statements assigned to the fallback class. On the held-out sample, F1 increases from $0.589$ to $0.653$ for \textsc{Credit-sharing} ($\Delta=+0.064$, 95\% CI $[+0.017,+0.115]$), from $0.709$ to $0.748$ for \textsc{Credit-claiming}, and from $0.682$ to $0.711$ for \textsc{Blame-shifting}. \textsc{Descriptive} decreases slightly from $0.623$ to $0.605$, consistent with the intended precision--recall tradeoff, while macro F1 increases from $0.651$ to $0.679$. Thus, the revision improves the targeted substantive-class errors even though the fallback class's own F1 does not increase (Appendix Table~\ref{tab:hunter-aggregate}). Additional inter-rater reliability analyses are reported in Appendix~\ref{sec:irr}.

Overall, the credit-claiming case shows that the same audit can identify a different fallback-class failure: over-prediction caused by conflicting written guidance rather than an open-ended residual rule.

\subsection{Cross-Case Lessons}
\label{sec:results-crosscase}

The two applications show why aggregate performance alone is insufficient for codebook-based annotation. In BFRS, poor performance on \textsc{Other} has limited effect on weighted F1 because the class accounts for only $6.5\%$ of the corpus. In the credit-claiming case, \textsc{Descriptive} accounts for $60.5\%$, so errors in smaller substantive classes can be obscured by the dominant fallback category. In both applications, fallback categories show different patterns: \textsc{Other} has both false positives and false negatives, whereas \textsc{Descriptive} is mainly over-predicted. The diagnostic patterns are stable under 20-per-class subsampling. Across 200 draws, BFRS \textsc{Other} is the lowest-F1 class in 96\% of samples and the \textsc{Other}--\textsc{Assassination} localization is reproduced in 97\%. For credit claiming, the \textsc{Descriptive}'s over-prediction pattern is reproduced in 100\% of draws (Appendix~\ref{app:stabtest}). These are application-specific stability results rather than a universal sample-size requirement.

\subsection{Behavior Across LLM Annotators}
\label{sec:gemma}

Because different LLMs can interpret the same written codebook differently, we repeat the analysis with Gemma-4-31B using the same samples, prompts, and codebook versions (Appendix~\ref{app:gemma}). We therefore treat the object of the audit as the \emph{codebook--annotator system}, rather than assuming that every diagnosed pattern is model-independent.

\begin{figure}[!htbp]
\centering
\includegraphics[width=0.95\textwidth]{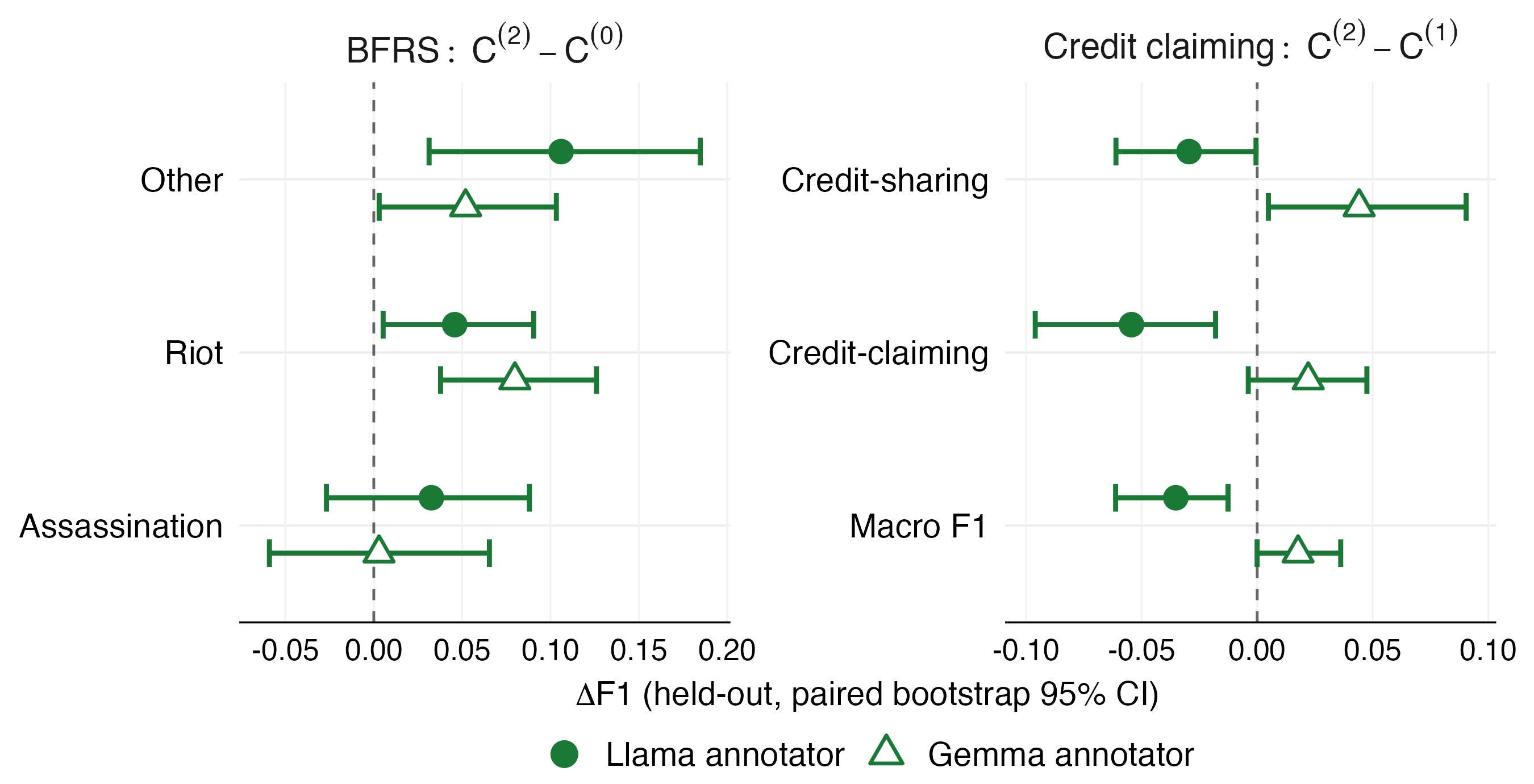}
\caption{\textbf{Revision effects across LLM annotators.}}
\notes{Held-out samples: BFRS, $n=1{,}045$; credit claiming, $n=321$ ($B=5{,}000$). For BFRS, $C^{(2)}-C^{(0)}$ compares the revised \textsc{Other} plus \textsc{Assassination} codebook with the original. For credit claiming, $C^{(2)}-C^{(1)}$ evaluates the Gemma-derived \textsc{Credit-sharing} revision relative to the revised \textsc{Descriptive} codebook.}
\label{fig:cross-llm-effects}
\end{figure}

For BFRS, the main diagnosis is stable across LLMs. Both Llama and Gemma identify \textsc{Other} as the weakest class and localize the main problem to the \textsc{Other}--\textsc{Assassination} boundary. The revisions developed with Llama also improve Gemma's held-out performance: under ($C^{(2)}$), \textsc{Other} increases by $0.052$ and \textsc{Riot} by up to $0.080$, while \textsc{Assassination} remains approximately stable. This provides evidence that some codebook clarifications transfer across annotators. The credit-claiming case shows a complementary pattern. Both models concentrate errors in the boundary region involving \textsc{Credit-sharing}, \textsc{Credit-claiming}, and \textsc{Descriptive}, but Llama nominates \textsc{Descriptive} whereas Gemma nominates \textsc{Credit-sharing}. The \textsc{Descriptive} revision ($C^{(1)}$) developed with Llama moves Gemma's classes in the development sample in the same general direction, but does not reproduce the same held-out gains. We therefore test a separate Gemma-derived \textsc{Credit-sharing} revision ($C^{(2)}$), not part of the Llama revision sequence. It raises Gemma's held-out \textsc{Credit-sharing} F1 by $0.044$, yet lowers Llama's \textsc{Credit-sharing} by $0.029$, \textsc{Credit-claiming} by $0.054$, and macro F1 by $0.035$. Taken together, the cross-LLM results show that some revision effects transfer across annotators, whereas others are annotator-specific. Figure~\ref{fig:cross-llm-effects} summarizes this contrast.

\section{Discussion}
\label{sec:results-limits}

Taken together, the two applications show that annotation errors can reveal different problems in the codebook–annotator system, and that those problems require different responses. Codebook auditing complements model-side improvement strategies. Prompt refinement, retrieval design, calibration, and fine-tuning can improve how well an LLM implements a supplied codebook. The audit addresses the related question of whether the written target is sufficiently specified. As the BFRS case shows, fine-tuning can substantially improve overall BFRS performance \parencite{Halterman2025codebook}, while our results show that a specific written boundary can still warrant revision (See Appendix ~\ref{app:finetune}). The two approaches can therefore be combined rather than treated as alternatives. In addition, low performance does not by itself justify revision. In BFRS, \textsc{Gov\_Attack\_on\_Civilian} has low recall, but document review points to ambiguous or inconsistent reference cases rather than a clear written defect. Such cases are reported as reference-label issues rather than used to justify another wording change. This guards against revising a codebook simply to increase F1.

The cross-LLM results also show that some codebook problems recur across annotators, whereas others are annotator-specific. In BFRS, Llama and Gemma identify the same problematic \textsc{Other}--\textsc{Assassination} boundary, and the Llama-derived revisions improve performance for both annotators. In the credit-claiming case, the same broader boundary region is difficult for both models, but the nominated class and the beneficial repair differ. This suggests that some weaknesses are properties of the written instrument that recur across annotators, while their observable error pattern and the most effective clarification can depend on an LLM's prior semantics and decision behavior. A revision should therefore be verified on the codebook--annotator system that will actually be used.

A practical advantage of the audit is that it concentrates human effort on the parts of the codebook where substantive judgment is most informative. Nomination and localization are computed automatically from model predictions. Researchers then inspect a small purposive sample, typically 5--10 documents per pattern, together with the relevant codebook entries and labels, to attribute the error source and decide whether revision is warranted. Our 20-per-class subsampling analysis suggests that the initial nomination and localization can be stable with relatively few labeled cases in these applications, although larger samples give more precise class-specific estimates and are preferable for held-out evaluation. The main human investment is therefore targeted substantive judgment rather than large-scale manual reannotation.

Improved held-out F1 should be interpreted narrowly. It shows that the revised wording makes the codebook easier for the evaluated annotator to apply in a way that better reproduces the reference labels. It does not establish that the revision is substantively correct or improves construct validity. Independent human application of the original and revised codebooks would provide a stronger test of whether the revision also improves reproducibility for human coders. Final decisions about construct meaning therefore remain with domain experts.

\section{Conclusion}
\label{sec:conclusion}

LLM annotation changes the role of the codebook from background documentation to an executable instruction set. Human coders may supplement written rules with training, discussion, adjudication, and tacit conventions; an LLM annotator relies much more directly on the released text. When that text is incomplete or internally inconsistent, annotation errors can help identify weaknesses in the operationalization as well as limitations of the annotation system.

We introduce a pre-deployment audit that combines reproducible class-level screening, confusion-matrix localization, structured human attribution of error sources, targeted codebook revision, and held-out evaluation. Across two applications, targeted revisions reduce the targeted errors, while cross-LLM analyses show that some clarifications transfer across annotators and others are system-specific.

The contribution is diagnostic. Rather than treating higher F1 as the only goal, the audit asks what disagreement reveals about the codebook--annotator system and the reference labels. This complements model-side improvements by helping researchers determine whether to improve the model, clarify the codebook, or revisit the reference labels before large-scale deployment.

\noindent\textbf{Funding.}
This work was supported by the Center for Socially Responsible Artificial Intelligence (CSRAI) at Penn State through its seed funding program.

\vspace{0.6em}
\small

\printbibliography

@article{bueno2015measuring,
  title={Measuring political violence in Pakistan: Insights from the BFRS Dataset},
  author={Bueno de Mesquita, Ethan and Fair, C Christine and Jordan, Jenna and Rais, Rasul Bakhsh and Shapiro, Jacob N},
  journal={Conflict Management and Peace Science},
  volume={32},
  number={5},
  pages={536--558},
  year={2015},
  doi ={10.1177/0738894214542401},
  publisher={SAGE Publications Sage UK: London, England}
}

@article{gilardi2023chatgpt,
  title={ChatGPT outperforms crowd workers for text-annotation tasks},
  author={Gilardi, Fabrizio and Alizadeh, Meysam and Kubli, Ma{\"e}l},
  journal={Proceedings of the National Academy of Sciences},
  volume={120},
  number={30},
  pages={e2305016120},
  year={2023},
  publisher={National Acad Sciences}
}

@article{Halterman2025codebook,
title={Codebook llms: Evaluating llms as measurement tools for political science concepts},
  author={Halterman, Andrew and Keith, Katherine A},
  journal={Political Analysis},
  volume={34},
  number={2},
  pages={188--204},
  doi={10.1017/pan.2025.10017},
  year={2026},
  publisher={Cambridge University Press}
}

@article{adcock2001measurement,
  title={Measurement validity: A shared standard for qualitative and quantitative research},
  author={Adcock, Robert and Collier, David},
  journal={American political science review},
  volume={95},
  number={3},
  pages={529--546},
  year={2001},
  publisher={Cambridge University Press}
}

@article{Hunter2025,
  title={Credit claiming in the European Union},
  author={Hunter, Tom},
  journal={The Journal of Politics},
  volume={87},
  number={3},
  pages={889--904},
  doi ={10.1086/732970},
  year={2025},
  publisher={The University of Chicago Press Chicago, IL}
}

@unpublished{Yang2025,
  title={Data Annotation with Large Language Models: Lessons from a Large Empirical Evaluation},
  author={Yang, Eddie and Wang, Zoey and Zhou, Carl and Xu, Yaosheng},
  year={2026}
}

@article{Alizadeh2025,
  author  = {Alizadeh, M. and Kubli, M. and Samei, Z. and Dehghani, S. and Zahedivafa, M. and Bermeo, J. D. and Korobeynikova, M. and Gilardi, F.},
  title   = {Open-source {LLMs} for Text Annotation: A Practical Guide for Model Setting and Fine-tuning},
  journal = {Journal of Computational Social Science},
  volume  = {8},
  pages   = {17},
  year    = {2025}
}

@article{Benoit2025,
  author  = {Benoit, K. and De Marchi, S. and Laver, C. and Laver, M. and Ma, J.},
  title   = {Using Large Language Models to Analyze Political Texts through Natural Language Understanding},
  journal = {American Journal of Political Science},
  year    = {2025}
}

@article{Griswold2025,
  author  = {Griswold, M. and Robbins, M. W. and Pollard, M. S.},
  title   = {Stay Tuned: Improving Sentiment Analysis and Stance Detection Using Large Language Models},
  journal = {Political Analysis},
  pages   = {1--20},
  year    = {2025}
}

@article{Ornstein2025,
  author  = {Ornstein, J. T. and Blasingame, E. N. and Truscott, J. S.},
  title   = {How to Train Your Stochastic Parrot: Large Language Models for Political Texts},
  journal = {Political Science Research and Methods},
  year    = {2025}
}

@article{casanova2026,
  title={On the Limits of LLM Adaptability: Impact of Model-Internalized Priors on Annotation Task Performance},
  author={Casanova, Etienne and Kocielnik, Rafal and Alvarez, R Michael},
  journal={ICML 2026 (Oral \& Spotlight); PMLR vol. 306.},
  year={2026}
}

@inproceedings{pangakis2025keeping,
  title={Keeping humans in the loop: human-centered automated annotation with generative AI},
  author={Pangakis, Nick and Wolken, Sam},
  booktitle={Proceedings of the International AAAI Conference on Web and Social Media},
  volume={19},
  pages={1471--1492},
  year={2025}
}

@article{Tornberg2024B, 
title={Best Practices for Text Annotation with Large Language Models}, volume={18}, url={https://sociologica.unibo.it/article/view/19461}, DOI={10.6092/issn.1971-8853/19461}, abstractNote={&amp;lt;p style=&amp;quot;font-weight: 400;&amp;quot;&amp;gt;Large Language Models (LLMs) have ushered in a new era of text annotation, as their ease-of-use, high accuracy, and relatively low costs have meant that their use has exploded in recent months. However, the rapid growth of the field has meant that LLM-based annotation has become something of an academic Wild West: the lack of established practices and standards has led to concerns about the quality and validity of research. Researchers have warned that the ostensible simplicity of LLMs can be misleading, as they are prone to bias, misunderstandings, and unreliable results. Recognizing the transformative potential of LLMs, this essay proposes a comprehensive set of standards and best practices for their reliable, reproducible, and ethical use. These guidelines span critical areas such as model selection, prompt engineering, structured prompting, prompt stability analysis, rigorous model validation, and the consideration of ethical and legal implications. The essay emphasizes the need for a structured, directed, and formalized approach to using LLMs, aiming to ensure the integrity and robustness of text annotation practices, and advocates for a nuanced and critical engagement with LLMs in social scientific research.&amp;lt;/p&amp;gt;}, number={2}, journal={Sociologica}, author={Törnberg, Petter}, year={2024}, month={Jan.}, pages={67–85} }

@inproceedings{halterman2026protest,
  author = {Halterman, Andrew and Keith, Katherine A.},
  title  = {What is a protest anyway? Codebook conceptualization is still a first-order concern in LLM-era classification},
  booktitle = {
Proceedings of the 64th Annual Meeting of the Association for Computational Linguistics (Volume 1: Long Papers)},
  year   = {2026}
}

@inproceedings{rottger2022,
  author    = {R{\"o}ttger, Paul and Vidgen, Bertie and Hovy, Dirk and Pierrehumbert, Janet B.},
  title={Two contrasting data annotation paradigms for subjective NLP tasks},
  booktitle={Proceedings of the 2022 conference of the North American chapter of the association for computational linguistics: human language technologies},
  pages={175--190},
  year={2022}
}

@article{klie2024analyzing,
  author  = {Klie, Jan-Christoph and Eckart de Castilho, Richard and Gurevych, Iryna},
  title={Analyzing dataset annotation quality management in the wild},
  journal={Computational Linguistics},
  volume={50},
  number={3},
  pages={817--866},
  year={2024},
  publisher={MIT Press 255 Main Street, 9th Floor, Cambridge, Massachusetts 02142, USA~…}
}

@article{he2026better,
  title={When Better Codebooks Are Not Enough: Predictive Performance and Behavioral Reliability in LLM Political Event Coding},
  author={He, Zixian and Murugesan, Bharath Raahul and Brandt, Patrick and Hu, Yibo},
  journal={arXiv preprint arXiv:2606.06781},
  year={2026}
}

\newpage
\appendix
\clearpage
\setcounter{page}{1}

\textbf{\huge{\centering{Supplementary Appendix for ``When F1 Is Not Enough: Auditing Codebooks for LLM-Assisted Annotation"}}}

{\large{\centering{Yuehong Cassandra Tai, Qunhua Li}}}

\section{Dataset Sampling}
\label{app:sampling}
We use a stratified \textbf{development} sample for screening, localization, attribution, and specification of the revision targets, and a disjoint \textbf{held-out validation} sample for evaluating the resulting codebook versions. For BFRS, $D_{\mathrm{dev}}$ contains $573$ documents. The final held-out set contains $1{,}045$ documents, targeting approximately $100$ per class; \textsc{drone\_assassination}, \textsc{threat\_of\_violence}, and \textsc{gov\_attack\_on\_civilians} are capped at the number of available documents. For credit claiming, $D_{\mathrm{dev}}$ contains $172$ documents and the final held-out set contains $321$ documents, with approximately $100$ each for \textsc{Credit-claiming}, \textsc{Credit-sharing}, and \textsc{Descriptive}; the smaller \textsc{Blame-shifting} population contributes $21$ held-out documents.

\paragraph{Construction of the held-out validation sample.}
The final $D_{\mathrm{test}}$ for each application is the union of two nonoverlapping class-stratified draws from the same held-out pool. The first draw targeted approximately $50$ documents per class. A second disjoint draw was subsequently added to increase the precision of the evaluation, bringing the target to approximately $100$ documents per class where the corpus permits. Sampling is without replacement, and all held-out documents are disjoint from $D_{\mathrm{dev}}$. The revision targets and wording are determined from the development sample; results on either held-out draw are not used to add, remove, reorder, or rewrite subsequent planned revisions. We therefore treat the combined documents as a single held-out validation sample in all reported analyses.
\paragraph{Decision stability at 20 documents per class.} The audit's decisions are made on the development sample, so it is worth asking how many labeled documents that sample needs. Specifically, we test whether the audit would have reached the same decisions from a smaller sample. From BFRS dev sample and Credit-claiming dev we draw $20$ documents per class without replacement, re-score, and re-apply the nomination and localization rules,  repeating $200$ times.  We use $20$ because it is the largest value at which every class in both schemes is genuinely subsampled. For BFRS, \textsc{threat\_of\_violence} has $30$ documents and for credit-claiming, \textsc{Blame-shifting} has $22$ documents.

\section{Annotator and Retrieval Configuration}
\label{app:annotator}

Both codebooks are annotated by Llama-3.3-70B-Instruct, served through Ollama in $4$-bit quantization on two A100-40GB node, run fully deterministically ($T=0$, \texttt{max\_tokens}${}=200$). The annotator emits a JSON object with a \texttt{label\_text} field and a two-sentence \texttt{rationale} (the chosen class and why; the strongest competitor and the codebook field that rules it out). Output parsed cleanly on all documents in both corpora. 

For BFRS, we use retrieval-augmented generation because the twelve-class codebook contains long, mutually confusable class-specific guidance. Each document is embedded with \texttt{all-mpnet-base-v2}; the top $k=6$ codebook records are retrieved by cosine similarity and expanded by up to three \texttt{confusable\_with} neighbors, capped at nine records. Retrieval recall, the proportion of documents for which the reference class appears in the retrieved set, is $0.969$ overall, with no class below $0.90$. Retrieval failure is therefore unlikely to explain the main diagnosed error patterns. The same retrieval configuration is held fixed across BFRS codebook versions. For \textcite{Hunter2025}'s four-class codebook, we use no retrieval and provide the full codebook as a single structured JSON object for every document.

\section{Pattern Disposition Rules}
\label{app:stoprule}

Steps~1--3 identify error patterns on $D_{\mathrm{dev}}$ and determine how each pattern should be handled. The disposition rules distinguish errors that warrant a codebook revision from those that should be documented but not addressed through wording changes. These dispositions are determined during the development-stage diagnosis; held-out performance does not add or remove revision targets.

Each diagnosed error pattern is assigned one of the following dispositions.
\begin{enumerate}
    \item \textbf{Revise the codebook.} Document-level review identifies a recurring defect in the written codebook, such as an incomplete assignment condition, an ambiguous class boundary, or internally conflicting guidance. The problem can be addressed by clarifying the existing operationalization without changing the intended construct. The implicated entry or boundary becomes a revision target in Step~4.

    \item \textbf{LLM error.} The written codebook provides a sufficiently clear rule that supports the reference label, but the annotator does not apply that rule correctly. The pattern is documented but does not motivate a codebook revision.

    \item \textbf{Reference-label issue.} The written codebook does not clearly support the reference label, or substantively similar documents receive inconsistent reference labels. The pattern is documented rather than used to revise the codebook solely to reproduce the existing labels.

    \item \textbf{Substantive issue.} The disagreement cannot be resolved by clarifying the existing written rules because the codebook does not specify the substantive criterion needed to distinguish the cases. Adding such a rule would extend or alter the operationalization rather than clarify it, so any change requires additional domain-expert justification.
\end{enumerate}

A pattern is assigned to codebook revision only when document review identifies a recurring, repairable problem in the written rules. Patterns that are mixed or indeterminate are documented but are not revised unless the review supports a specific codebook defect. The dispositions from Steps~1--3 define the revision targets used in Step~4. When more than one entry or boundary is implicated, the revisions are implemented sequentially so that the effect of each change can be examined separately. Steps~1--3 are not rerun after each revision, and held-out performance is not used to add or remove revision targets. After all diagnosed repairable defects have been addressed, the resulting codebook is denoted $C^{(T)}$.

The two applications illustrate different dispositions. In BFRS, the open-ended \textsc{Other} rule and the \textsc{Other}--\textsc{Assassination} boundary are classified as repairable codebook defects and therefore revised. In contrast, the \textsc{Gov\_Attack\_on\_Civilian} confusion is not revised because document review reveals substantively similar cases with inconsistent reference labels. In the credit-claiming application, conflicting guidance for future-oriented attribution is classified as a repairable codebook defect and motivates the revision to \textsc{Descriptive}.

\section{Validation Sample-Size Considerations}
\label{app:power}

The audit determines whether a codebook revision is warranted using the development sample. A separate practical question is how large the held-out sample should be to evaluate the resulting revision with reasonable precision. Because the effect of a targeted revision depends on how often the diagnosed error occurs, we use a simple paired calculation as a sample-size planning heuristic.

Consider a class with $n$ reference documents in the held-out sample. Comparing predictions under the original and revised codebooks, let $d_{+}$ denote the number of documents that change from incorrectly classified to correctly classified, and let $d_{-}$ denote the number that change from correctly classified to incorrectly classified. For the recall component of class performance, the paired change in the proportion correctly classified is
\[
\Delta=\frac{d_{+}-d_{-}}{n}.
\]
Let
\[
p_d=\frac{d_{+}+d_{-}}{n}
\]
denote the proportion of documents whose correctness changes between the two codebook versions. The variance of the document-level paired difference is $p_d-\Delta^2$, giving the conservative approximation
\[
SE(\Delta)\approx\sqrt{\frac{p_d}{n}}.
\]

To obtain an interpretable planning value, consider the favorable case in which a revision corrects errors without introducing new ones, so that $d_{-}=0$. Then $\Delta=p_d=d_{+}/n$. Under a normal approximation, a two-sided comparison with $\alpha=.05$ and approximately 80\% power requires
\[
\frac{\Delta}{SE(\Delta)}
\gtrsim z_{.975}+z_{.80}
\approx 1.96+0.84=2.80.
\]
Substitution gives
\[
d_{+}\gtrsim (2.80)^2\approx 7.8.
\]
Thus, under this favorable scenario, an effect corresponding to roughly eight corrected documents is large enough to be detectable under conventional significance and power levels. If the targeted error occurs in a proportion $p$ of documents from a class, a rough planning value is therefore
\[
n\approx\frac{8}{p}.
\]
For example, if the diagnosed error occurs in approximately 8\% of documents, a held-out sample of about 100 documents from that class would contain roughly eight affected cases on average. This calculation motivated our target of approximately 100 held-out documents per class where the available corpus permitted.

This calculation is only a planning heuristic. It concerns the recall component of class performance; changes in precision depend on predictions for documents belonging to other reference classes. Actual revisions may both correct existing errors and introduce new ones, and the prevalence of an error pattern observed on $D_{\mathrm{dev}}$ need not be identical in $D_{\mathrm{test}}$. Most importantly, this is not a formal power calculation for class-specific F1 and is not used to determine whether a codebook defect should be revised or whether a revision has succeeded. Revision decisions are based on $D_{\mathrm{dev}}$ and substantive review; revision effects are estimated on $D_{\mathrm{test}}$ using paired document-level bootstrap confidence intervals for the reported performance measures.

\newpage
\section{BFRS per-class F1 and Confusion Matrix}
\label{app:perclass}

\begin{table}[!htbp]
\centering
\caption{BFRS Annotator per-class F1 on $D_{\mathrm{dev}}$ (original codebook,  ($C^{0}$))}
\label{tab:per-class-f1}
\resizebox{\textwidth}{!}{%
\begin{tabular}{lrrrrr}
\toprule
Class & $n$ & precision & recall & F1 & 95\% CI \\
\midrule
\textsc{drone\_assassination}                 & 43 & 1.00 & 1.00 & 1.000 & $[1.00, 1.00]$ \\
\textsc{attack\_on\_state}                    & 50 & 0.89 & 0.96 & 0.923 & $[0.86, 0.97]$ \\
\textsc{threat\_of\_violence}                 & 30 & 0.96 & 0.70 & 0.808 & $[0.67, 0.91]$ \\
\textsc{violent\_political\_demonstration}    & 50 & 0.71 & 0.92 & 0.800 & $[0.71, 0.88]$ \\
\textsc{guerilla\_attack\_on\_gov\_forces}    & 50 & 0.87 & 0.68 & 0.764 & $[0.66, 0.86]$ \\
\textsc{gov\_attack\_on\_nonstate\_combatants}& 50 & 0.68 & 0.78 & 0.729 & $[0.62, 0.82]$ \\
\textsc{conventional\_attack\_on\_gov\_forces}& 50 & 0.60 & 0.82 & 0.695 & $[0.60, 0.78]$ \\
\textsc{riot}                                 & 50 & 0.87 & 0.54 & 0.667 & $[0.54, 0.77]$ \\
\textsc{terrorism}                            & 50 & 0.58 & 0.70 & 0.636 & $[0.52, 0.73]$ \\
\textsc{assassination}                        & 50 & 0.60 & 0.64 & 0.621 & $[0.50, 0.72]$ \\
\textsc{gov\_attack\_on\_civilians}           & 50 & 0.85 & 0.46 & 0.597 & $[0.45, 0.72]$ \\
\textsc{other}                                & 50 & 0.39 & 0.42 & 0.404 & $[0.29, 0.52]$ \\
\bottomrule
\notes{sorted by F1, bootstrap $95\%$ CIs over $2000$ resamples. The residual catch-all \textsc{other} is the worst class by a clear point-estimate margin (${\sim}0.19$ below the next), the unambiguous Step-1 nominee.}
\end{tabular} %
}
\end{table}

\begin{table}[!htbp]
\centering
\caption{Step-1 nomination screen on BFRS $D_{\mathrm{dev}}$ under the original codebook $C^{(0)}$}
\label{tab:bfrs-nomination-screen}
\resizebox{\textwidth}{!}{%
\begin{tabular}{lrr}
\toprule
Quantity & Estimate & 95\% CI \\
\midrule
\multicolumn{3}{l}{BFRS (dev, $n=573$)} \\
\addlinespace
Level gap: F1(\textsc{gov\_attack\_on\_civilians}) $-$ F1(\textsc{other}) & $\mathbf{+0.194}$ & $[+0.010, +0.367]$ \\
\addlinespace
Asymmetry $A_k$: \textsc{gov\_attack\_on\_civilians} & $\mathbf{+0.392}$ & $[+0.230, +0.557]$ \\
Asymmetry $A_k$: \textsc{riot} & $\mathbf{+0.331}$ & $[+0.171, +0.492]$ \\
Asymmetry $A_k$: \textsc{threat\_of\_violence} & $\mathbf{+0.255}$ & $[+0.083, +0.429]$ \\
Asymmetry $A_k$: \textsc{conventional\_attack\_on\_gov\_forces} & $\mathbf{-0.217}$ & $[-0.360, -0.084]$ \\
Asymmetry $A_k$: \textsc{violent\_political\_demonstration} & $\mathbf{-0.212}$ & $[-0.336, -0.089]$ \\
Asymmetry $A_k$: \textsc{guerilla\_attack\_on\_gov\_forces} & $\mathbf{+0.192}$ & $[+0.042, +0.346]$ \\
Asymmetry $A_k$: \textsc{terrorism} & $-0.117$ & $[-0.259, +0.026]$ \\
Asymmetry $A_k$: \textsc{gov\_attack\_on\_nonstate\_combatants} & $-0.096$ & $[-0.243, +0.053]$ \\
Asymmetry $A_k$: \textsc{attack\_on\_state} & $-0.071$ & $[-0.170, +0.021]$ \\
Asymmetry $A_k$: \textsc{assassination} & $-0.036$ & $[-0.188, +0.119]$ \\
Asymmetry $A_k$: \textsc{other} & $-0.031$ & $[-0.156, +0.089]$ \\
Asymmetry $A_k$: \textsc{drone\_assassination} & $0.000$ & $[0.000, 0.000]$ \\
\bottomrule
\end{tabular}%
}
\notes{Development sample, original codebook $C^{(0)}$, Llama annotator. The level signal is the gap $F1_{k_{(2)}}-F1_{k_{(1)}}$ between the second-lowest and the lowest F1 class; the asymmetry signal is $A_k=P_k-R_k$, reported for every class and sorted by $|A_k|$ descending within each panel. Positive $A_k$ indicates under-prediction, negative $A_k$ over-prediction. Intervals are percentile $95\%$ confidence intervals from a paired document-level bootstrap with replacement ($B=5{,}000$).}
\end{table}

\begin{table}[!htbp]
\centering
\caption{BFRS Annotator per-class F1 on $D_{\mathrm{dev}}$ (revised \textsc{other} codebook, $C^{(1)}$)}
\label{tab:per-class-f1-c1}
\resizebox{\textwidth}{!}{%
\begin{tabular}{lrrrrr}
\toprule
Class & $n$ & precision & recall & F1 & 95\% CI \\
\midrule
\textsc{drone\_assassination}                 & 43 & 1.00 & 1.00 & 1.000 & $[1.00, 1.00]$ \\
\textsc{attack\_on\_state}                    & 50 & 0.89 & 0.96 & 0.923 & $[0.86, 0.97]$ \\
\textsc{threat\_of\_violence}                 & 30 & 0.95 & 0.70 & 0.808 & $[0.67, 0.92]$ \\
\textsc{violent\_political\_demonstration}    & 50 & 0.69 & 0.90 & 0.783 & $[0.69, 0.86]$ \\
\textsc{guerilla\_attack\_on\_gov\_forces}    & 50 & 0.89 & 0.68 & 0.773 & $[0.67, 0.86]$ \\
\textsc{gov\_attack\_on\_nonstate\_combatants}& 50 & 0.71 & 0.80 & 0.755 & $[0.66, 0.84]$ \\
\textsc{riot}                                 & 50 & 0.87 & 0.66 & 0.750 & $[0.63, 0.84]$ \\
\textsc{conventional\_attack\_on\_gov\_forces}& 50 & 0.63 & 0.88 & 0.733 & $[0.63, 0.82]$ \\
\textsc{terrorism}                            & 50 & 0.62 & 0.70 & 0.660 & $[0.54, 0.76]$ \\
\textsc{assassination}                        & 50 & 0.65 & 0.66 & 0.653 & $[0.53, 0.75]$ \\
\textsc{gov\_attack\_on\_civilians}           & 50 & 0.88 & 0.42 & 0.568 & $[0.41, 0.69]$ \\
\textsc{other}                                & 50 & 0.46 & 0.52 & 0.491 & $[0.37, 0.61]$ \\
\bottomrule
\notes{sorted by F1, bootstrap $95\%$ CIs over $2000$ resamples; same dev sample and metrics as Table~\ref{tab:per-class-f1}. \textsc{other} rises from $0.404$ to $0.491$ and remains the worst class; \textsc{Riot} improves even though its entry is not edited.}
\end{tabular} %
}
\end{table}

\begin{table}[!htbp]
\centering
\caption{BFRS Annotator per-class F1 on $D_{\mathrm{dev}}$ (revised \textsc{other} plus \textsc{assass} codebook, $C^{(2)}$)}
\label{tab:per-class-f1-c2}
\resizebox{\textwidth}{!}{%
\begin{tabular}{lrrrrr}
\toprule
Class & $n$ & precision & recall & F1 & 95\% CI \\
\midrule
\textsc{drone\_assassination}                 & 43 & 1.00 & 1.00 & 1.000 & $[1.00, 1.00]$ \\
\textsc{attack\_on\_state}                    & 50 & 0.89 & 0.96 & 0.923 & $[0.86, 0.97]$ \\
\textsc{threat\_of\_violence}                 & 30 & 0.95 & 0.70 & 0.808 & $[0.67, 0.92]$ \\
\textsc{guerilla\_attack\_on\_gov\_forces}    & 50 & 0.88 & 0.72 & 0.791 & $[0.69, 0.87]$ \\
\textsc{violent\_political\_demonstration}    & 50 & 0.68 & 0.90 & 0.776 & $[0.69, 0.85]$ \\
\textsc{riot}                                 & 50 & 0.87 & 0.68 & 0.764 & $[0.65, 0.85]$ \\
\textsc{gov\_attack\_on\_nonstate\_combatants}& 50 & 0.74 & 0.78 & 0.757 & $[0.66, 0.84]$ \\
\textsc{conventional\_attack\_on\_gov\_forces}& 50 & 0.62 & 0.88 & 0.727 & $[0.63, 0.81]$ \\
\textsc{assassination}                        & 50 & 0.67 & 0.70 & 0.686 & $[0.56, 0.78]$ \\
\textsc{terrorism}                            & 50 & 0.67 & 0.68 & 0.673 & $[0.56, 0.77]$ \\
\textsc{other}                                & 50 & 0.53 & 0.62 & 0.569 & $[0.45, 0.67]$ \\
\textsc{gov\_attack\_on\_civilians}           & 50 & 0.91 & 0.40 & 0.556 & $[0.40, 0.68]$ \\
\bottomrule
\notes{sorted by F1, bootstrap $95\%$ CIs over $2000$ resamples; same dev sample and metrics as Table~\ref{tab:per-class-f1}. \textsc{other} reaches $0.569$ and \textsc{assassination} $0.686$: the added targeting condition improves both sides of the revised boundary, and \textsc{other} is no longer the worst class.}
\end{tabular} %
}
\end{table}

\begin{table}[!htbp]
\centering
\caption{BFRS per-class precision and recall at every revision ladder, $D_{\mathrm{dev}}$}
\label{tab:pr-rung-grid-bfrs-dev}
\resizebox{\textwidth}{!}{%
\begin{tabular}{lrrrrrrr}
\toprule
 & & \multicolumn{2}{c}{$C^{(0)}$} & \multicolumn{2}{c}{$C^{(1)}$} & \multicolumn{2}{c}{$C^{(2)}$} \\
\cmidrule(lr){3-4}\cmidrule(lr){5-6}\cmidrule(lr){7-8}
Class & $n$ & $P$ & $R$ & $P$ & $R$ & $P$ & $R$ \\
\midrule
\textsc{other}                                & 50 & \textbf{0.39} & \textbf{0.42} & \textbf{0.46} & \textbf{0.52} & 0.53 & 0.62 \\
\textsc{gov\_attack\_on\_civilians}           & 50 & 0.85 & 0.46 & 0.88 & 0.42 & \textbf{0.91} & \textbf{0.40} \\
\textsc{assassination}                        & 50 & 0.60 & 0.64 & 0.65 & 0.66 & 0.67 & 0.70 \\
\textsc{terrorism}                            & 50 & 0.58 & 0.70 & 0.62 & 0.70 & 0.67 & 0.68 \\
\textsc{riot}                                 & 50 & 0.87 & 0.54 & 0.87 & 0.66 & 0.87 & 0.68 \\
\textsc{conventional\_attack\_on\_gov\_forces}& 50 & 0.60 & 0.82 & 0.63 & 0.88 & 0.62 & 0.88 \\
\textsc{gov\_attack\_on\_nonstate\_combatants}& 50 & 0.68 & 0.78 & 0.71 & 0.80 & 0.74 & 0.78 \\
\textsc{guerilla\_attack\_on\_gov\_forces}    & 50 & 0.87 & 0.68 & 0.89 & 0.68 & 0.88 & 0.72 \\
\textsc{violent\_political\_demonstration}    & 50 & 0.71 & 0.92 & 0.69 & 0.90 & 0.68 & 0.90 \\
\textsc{threat\_of\_violence}                 & 30 & 0.95 & 0.70 & 0.95 & 0.70 & 0.95 & 0.70 \\
\textsc{attack\_on\_state}                    & 50 & 0.89 & 0.96 & 0.89 & 0.96 & 0.89 & 0.96 \\
\textsc{drone\_assassination}                 & 43 & 1.00 & 1.00 & 1.00 & 1.00 & 1.00 & 1.00 \\
\bottomrule
\end{tabular} %
}
\notes{Rows are sorted by $C^{(0)}$ F1. The audit's revision targets are determined from the original-codebook diagnosis in Steps~1--3; the later precision and recall values are reported descriptively to show how the planned revisions change annotation behavior. \textsc{Other} is the level-screen nominee under $C^{(0)}$, while \textsc{Gov\_Attack\_on\_Civilian} is nominated by the precision--recall asymmetry and is reviewed during the initial attribution stage. Because the latter pattern is traced to ambiguous or inconsistent reference labels (Appendix~\ref{app:goldcases}), it is not a codebook-revision target.}
\end{table}

\newpage
\begin{table}[!htbp]
\centering
\caption{BFRS per-class precision and recall across codebook versions on the held-out validation sample ($n=1{,}045$)}
\label{tab:pr-rung-grid-bfrs-test}
\resizebox{\textwidth}{!}{%
\begin{tabular}{lrrrrrrr}
\toprule
 & & \multicolumn{2}{c}{$C^{(0)}$} & \multicolumn{2}{c}{$C^{(1)}$} & \multicolumn{2}{c}{$C^{(2)}$} \\
\cmidrule(lr){3-4}\cmidrule(lr){5-6}\cmidrule(lr){7-8}
Class & $n$ & $P$ & $R$ & $P$ & $R$ & $P$ & $R$ \\
\midrule
\textsc{other}                                & 100 & \textbf{0.44} & \textbf{0.42} & 0.51 & 0.47 & 0.54 & 0.53 \\
\textsc{gov\_attack\_on\_civilians}           & 78 & 0.81 & 0.33 & \textbf{0.80} & \textbf{0.26} & \textbf{0.83} & \textbf{0.26} \\
\textsc{threat\_of\_violence}                 & 32 & 1.00 & 0.56 & 1.00 & 0.56 & 1.00 & 0.56 \\
\textsc{terrorism}                            & 100 & 0.63 & 0.84 & 0.63 & 0.83 & 0.63 & 0.81 \\
\textsc{riot}                                 & 100 & 0.80 & 0.66 & 0.80 & 0.75 & 0.80 & 0.74 \\
\textsc{guerilla\_attack\_on\_gov\_forces}    & 100 & 0.84 & 0.64 & 0.85 & 0.64 & 0.85 & 0.62 \\
\textsc{conventional\_attack\_on\_gov\_forces}& 100 & 0.65 & 0.84 & 0.63 & 0.85 & 0.63 & 0.84 \\
\textsc{assassination}                        & 100 & 0.74 & 0.75 & 0.78 & 0.73 & 0.80 & 0.76 \\
\textsc{violent\_political\_demonstration}    & 100 & 0.72 & 0.86 & 0.71 & 0.87 & 0.70 & 0.87 \\
\textsc{gov\_attack\_on\_nonstate\_combatants}& 100 & 0.72 & 0.87 & 0.70 & 0.85 & 0.70 & 0.85 \\
\textsc{attack\_on\_state}                    & 100 & 0.95 & 0.96 & 0.95 & 0.96 & 0.95 & 0.96 \\
\textsc{drone\_assassination}                 & 35 & 0.97 & 0.97 & 0.97 & 0.97 & 0.97 & 0.97 \\
\bottomrule
\end{tabular} %
}
\end{table}
\newpage

\begin{table}[!htbp]
\centering
\caption{BFRS per-class and aggregate F1 on the development and held-out validation samples}
\label{tab:bfrs-aggregate}
\resizebox{\textwidth}{!}{%
\begin{tabular}{llll}
\toprule
F1 & $C^{(0)}$ & $C^{(1)}$ ($\Delta$) & $C^{(2)}$ ($\Delta$) \\
\midrule
\multicolumn{4}{l}{Dev ($n=573$, $50$/class)}\\
\textsc{other}         & $0.404$ & $0.491$ ($\mathbf{+0.087}$ $[+0.007,+0.171]$) & $0.569$ ($\mathbf{+0.164}$ $[+0.062,+0.272]$) \\
\textsc{riot}          & $0.667$ & $0.750$ ($\mathbf{+0.083}$ $[+0.000,+0.173]$) & $0.764$ ($\mathbf{+0.098}$ $[+0.011,+0.193]$) \\
\textsc{assassination} & $0.621$ & $0.653$ ($+0.033$ $[-0.030,+0.100]$)          & $0.686$ ($+0.065$ $[-0.022,+0.157]$) \\
macro                  & $0.720$ & $0.741$ ($\mathbf{+0.021}$ $[+0.003,+0.039]$) & $0.753$ ($\mathbf{+0.032}$ $[+0.010,+0.054]$) \\
weighted               & $0.714$ & $0.736$ ($\mathbf{+0.022}$ $[+0.003,+0.041]$) & $0.748$ ($\mathbf{+0.034}$ $[+0.011,+0.056]$) \\
\midrule
\multicolumn{4}{l}{Held-out validation ($n=1{,}045$, ${\sim}100$/class; \textsc{drone} $35$, \textsc{threat} $32$, \textsc{gov\_civ} $78$)}\\
\textsc{other}         & $0.429$ & $0.487$ ($+0.058$ $[-0.008,+0.129]$) & $0.535$ ($\mathbf{+0.106}$ $[+0.031,+0.185]$) \\
\textsc{riot}          & $0.725$ & $0.773$ ($\mathbf{+0.048}$ $[+0.009,+0.092]$) & $0.771$ ($\mathbf{+0.046}$ $[+0.005,+0.090]$) \\
\textsc{assassination} & $0.746$ & $0.753$ ($+0.006$ $[-0.039,+0.049]$) & $0.779$ ($+0.033$ $[-0.027,+0.088]$) \\
macro                  & $0.731$ & $0.731$ ($+0.000$ $[-0.013,+0.013]$) & $0.734$ ($+0.003$ $[-0.011,+0.017]$) \\
weighted               & $0.722$ & $0.724$ ($+0.002$ $[-0.013,+0.016]$) & $0.727$ ($+0.005$ $[-0.010,+0.021]$) \\
\bottomrule
\end{tabular}%
}
\notes{First column is the F1 metric; \textsc{consistent} is the baseline level, and each revision column reports its F1 level and paired with-replacement bootstrap difference ($B=5000$) from the original codebook on the same documents, with 95\% confidence intervals in brackets. Macro and weighted F1 are computed over all twelve classes. The held-out panel is the final validation sample used for out-of-sample evaluation.}
\end{table}
\newpage
\newpage

\section{Credit-Claiming per-class F1 and Confusion Matrix}
\label{app:hunter-perclass}
The per-class F1 with bootstrap CIs and the confusion matrix, on the unrevised base codebook (dev, text). Unlike BFRS, the four classes cluster (F1 $0.67$--$0.75$), so no class is clearly separated at the bottom of the F1 distribution. The fallback class \textsc{Descriptive} has the lowest precision and highest recall, indicating over-prediction, but its F1 is not clearly separated from the other classes.

\begin{table}[!htbp]
\centering
\caption{Credit-claiming Annotator per-class F1 on $D_{\mathrm{dev}}$ (original codebook $C^{0}$)}
\label{tab:hunter-perclass}
\begin{tabular}{lrrrrr}
\toprule
Category & $n$ & precision & recall & F1 & 95\% CI \\
\midrule
Blame-shifting        & 22 & 0.83 & 0.68 & 0.750 & $[0.58, 0.88]$ \\
Claiming              & 50 & 0.75 & 0.72 & 0.735 & $[0.63, 0.82]$ \\
Descriptive & 50 & 0.59 & 0.82 & 0.683 & $[0.58, 0.78]$ \\
Sharing              & 50 & 0.81 & 0.58 & 0.674 & $[0.54, 0.78]$ \\
\bottomrule
\end{tabular}
\end{table}

\begin{table}[!htbp]
\centering
\caption{Step-1 nomination screen on Credit-Claim $D_{\mathrm{dev}}$ under the original codebook $C^{(0)}$}
\label{tab:credit-nomination-screen}
\resizebox{\textwidth}{!}{%
\begin{tabular}{lrr}
\toprule
Quantity & Estimate & 95\% CI \\
\midrule
\multicolumn{3}{l}{Credit claiming (dev, $n=172$)} \\
\addlinespace
Level gap: F1(Descriptive) $-$ F1(Sharing) & $+0.009$ & $[-0.124, +0.144]$ \\
\addlinespace
Asymmetry $A_k$: Descriptive & $\mathbf{-0.234}$ & $[-0.370, -0.096]$ \\
Asymmetry $A_k$: Sharing & $\mathbf{+0.226}$ & $[+0.067, +0.385]$ \\
Asymmetry $A_k$: Blame-shifting & $+0.152$ & $[-0.083, +0.385]$ \\
Asymmetry $A_k$: Claiming & $+0.030$ & $[-0.117, +0.188]$ \\
\bottomrule
\end{tabular}%
}
\notes{Development sample, original codebook $C^{(0)}$, Llama annotator. The level signal is the gap $F1_{k_{(2)}}-F1_{k_{(1)}}$ between the second-lowest and the lowest F1 class; the asymmetry signal is $A_k=P_k-R_k$, reported for every class and sorted by $|A_k|$ descending within each panel. Positive $A_k$ indicates under-prediction, negative $A_k$ over-prediction. Intervals are percentile $95\%$ confidence intervals from a paired document-level bootstrap with replacement ($B=5{,}000$).}
\end{table}

\begin{table}[!htbp]
\centering
\caption{Credit-claiming Annotator per-class F1 on $D_{\mathrm{dev}}$ (revised Descriptive, $C^{(1)}$)}
\label{tab:hunter-perclass-c1}
\begin{tabular}{lrrrrr}
\toprule
Category & $n$ & precision & recall & F1 & 95\% CI \\
\midrule
\emph{Sharing}                                & 50 & 0.87 & 0.66 & 0.750 & $[0.63, 0.84]$ \\
\emph{Blame-shifting}                         & 22 & 0.83 & 0.68 & 0.750 & $[0.58, 0.88]$ \\
\emph{Claiming}                               & 50 & 0.76 & 0.74 & 0.747 & $[0.65, 0.83]$ \\
\emph{Descriptive (residual)}                 & 50 & 0.60 & 0.80 & 0.684 & $[0.58, 0.77]$ \\
\bottomrule
\end{tabular} %
\end{table}

\begin{table}[h]
\centering
\caption{Credit-claiming per-class precision and recall at every version, $D_{\mathrm{dev}}$}
\label{tab:pr-rung-grid-hunter-dev}
\resizebox{\textwidth}{!}{%
\begin{tabular}{lrrrrr}
\toprule
 & & \multicolumn{2}{c}{$C^{(0)}$} & \multicolumn{2}{c}{$C^{(1)}$} \\
\cmidrule(lr){3-4}\cmidrule(lr){5-6}
Class & $n$ & $P$ & $R$ & $P$ & $R$ \\
\midrule
\emph{Sharing}                                & 50 & \textbf{0.81} & \textbf{0.58} & 0.87 & 0.66 \\
\emph{Descriptive (residual)}                 & 50 & 0.59 & 0.82 & \textbf{0.60} & \textbf{0.80} \\
\emph{Claiming}                               & 50 & 0.75 & 0.72 & 0.76 & 0.74 \\
\emph{Blame-shifting}                         & 22 & 0.83 & 0.68 & 0.83 & 0.68 \\
\bottomrule
\notes{At $C^{(0)}$ Credit-sharing is nominally the lowest-F1 class ($0.674$ against Descriptive's $0.683$, a gap of $0.009$), while Descriptive shows the strongest over-prediction pattern: it has the lowest precision ($0.59$) and highest recall ($0.82$) in the table, triggering the asymmetry criterion. The revision trades that recall for precision by design: at $C^{(1)}$ Credit-sharing rises to $0.750$ while Descriptive stays flat ($0.683 \to 0.684$) and remains the residual.}
\end{tabular} %
}
\end{table}

\begin{table}[h]
\centering
\caption{Credit-claiming per-class precision and recall across codebook versions on the held-out validation sample ($n=321$)}
\label{tab:pr-rung-grid-hunter-test}
\resizebox{\textwidth}{!}{%
\begin{tabular}{lrrrrr}
\toprule
 & & \multicolumn{2}{c}{$C^{(0)}$} & \multicolumn{2}{c}{$C^{(1)}$} \\
\cmidrule(lr){3-4}\cmidrule(lr){5-6}
Class & $n$ & $P$ & $R$ & $P$ & $R$ \\
\midrule
\emph{Sharing}                                & 100 & \textbf{0.66} & \textbf{0.53} & 0.67 & 0.64 \\
\emph{Descriptive (residual)}                 & 100 & 0.58 & 0.67 & \textbf{0.62} & \textbf{0.59} \\
\emph{Blame-shifting}                         & 21 & 0.65 & 0.71 & 0.67 & 0.76 \\
\emph{Claiming}                               & 100 & 0.70 & 0.72 & 0.73 & 0.77 \\
\bottomrule

\end{tabular} %
}
\end{table}
\newpage

\begin{table}[!htbp]
\centering
\small
\caption{Credit-claiming per-class and aggregate F1 on the development and held-out validation samples}
\label{tab:hunter-aggregate}
\begin{tabular}{lll}
\toprule
F1 & $C{^(0)}$ & $C{^(1)}$ ($\Delta$) \\
\midrule
\multicolumn{3}{l}{Dev ($n=172$)}\\
Descriptive & $0.683$ & $0.684$ ($+0.001$ $[-0.051,+0.052]$) \\
Claiming         & $0.735$ & $0.747$ ($+0.013$ $[-0.030,+0.057]$) \\
Sharing           & $0.674$ & $0.750$ ($\mathbf{+0.077}$ $[+0.015,+0.150]$) \\
Blame-shifting     & $0.750$ & $0.750$ ($+0.000$ $[+0.000,+0.000]$) \\
macro                     & $0.711$ & $0.733$ ($+0.023$ $[-0.008,+0.055]$) \\
weighted                  & $0.704$ & $0.730$ ($+0.026$ $[-0.009,+0.063]$) \\
\midrule
\multicolumn{3}{l}{Held-out validation ($n=321$, ${\sim}100$/class; Blame-shifting $21$)}\\
Descriptive & $0.623$ & $0.605$ ($-0.018$ $[-0.063,+0.023]$) \\
Claiming          & $0.709$ & $0.748$ ($+0.038$ $[+0.000,+0.079]$) \\
Sharing           & $0.589$ & $0.653$ ($\mathbf{+0.064}$ $[+0.017,+0.115]$) \\
Blame-shifting    & $0.682$ & $0.711$ ($+0.030$ $[+0.000,+0.101]$) \\
macro                     & $0.651$ & $0.679$ ($\mathbf{+0.029}$ $[+0.000,+0.060]$) \\
weighted                  & $0.643$ & $0.671$ ($+0.028$ $[-0.003,+0.060]$) \\
\bottomrule
\end{tabular}
\notes{Each $\Delta$ is the mean of the paired with-replacement bootstrap difference distribution ($B=5{,}000$), not the difference of the two rounded F1 columns.}
\end{table}

\newpage

\section{Label-Name Ablation}
\label{app:label-name-ablation}

\begin{figure}[!htbp]
\centering
\begin{subfigure}[t]{0.49\textwidth}
    \centering
    \includegraphics[width=\linewidth,trim = 1.2cm 0cm 0cm 0.8cm, clip=true]{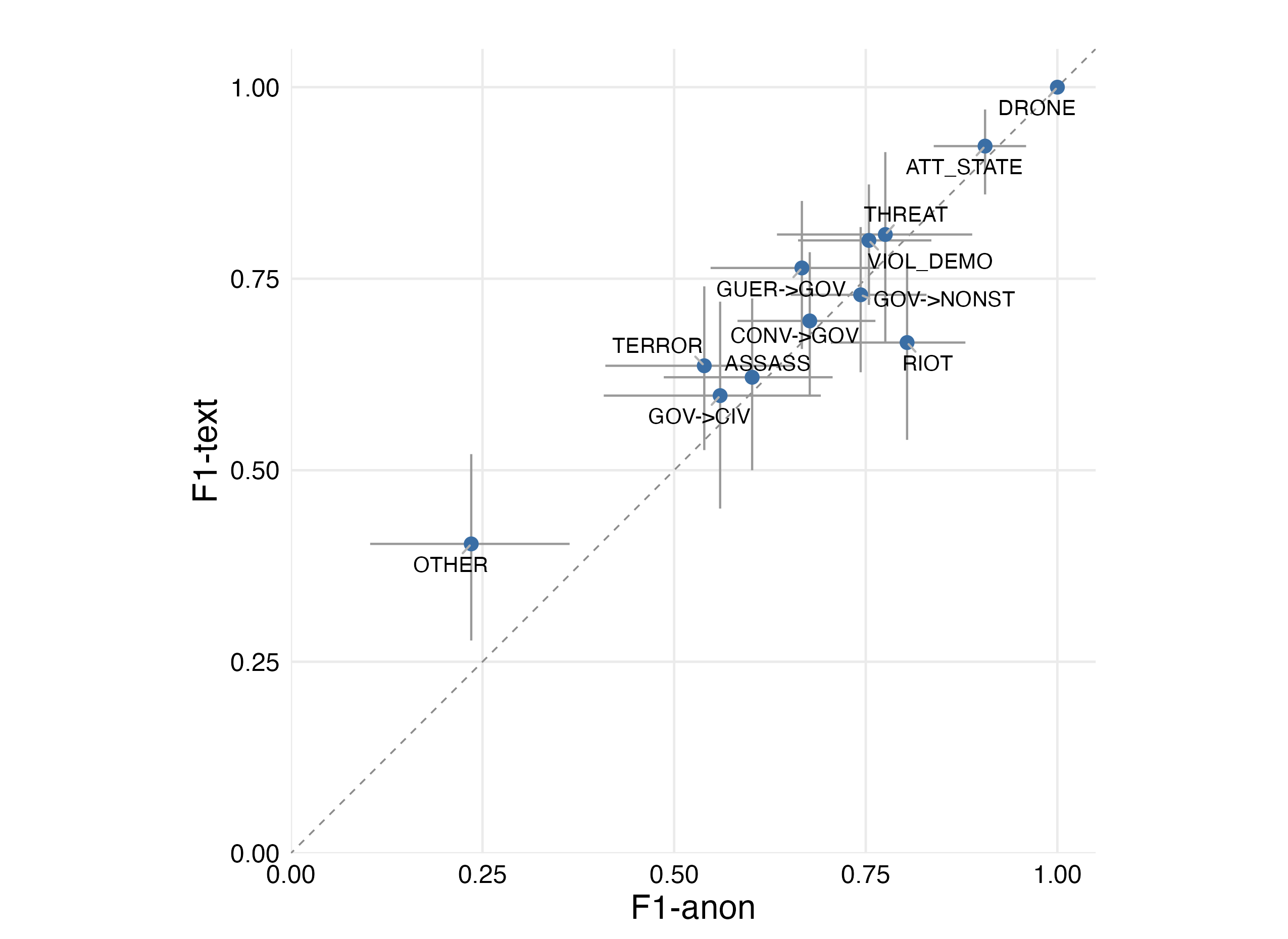}
    \caption{BFRS.}
\end{subfigure}\hfill
\begin{subfigure}[t]{0.49\textwidth}
    \centering
    \includegraphics[width=\linewidth,trim = 0.5cm 0cm 0cm 0.5cm, clip=true]{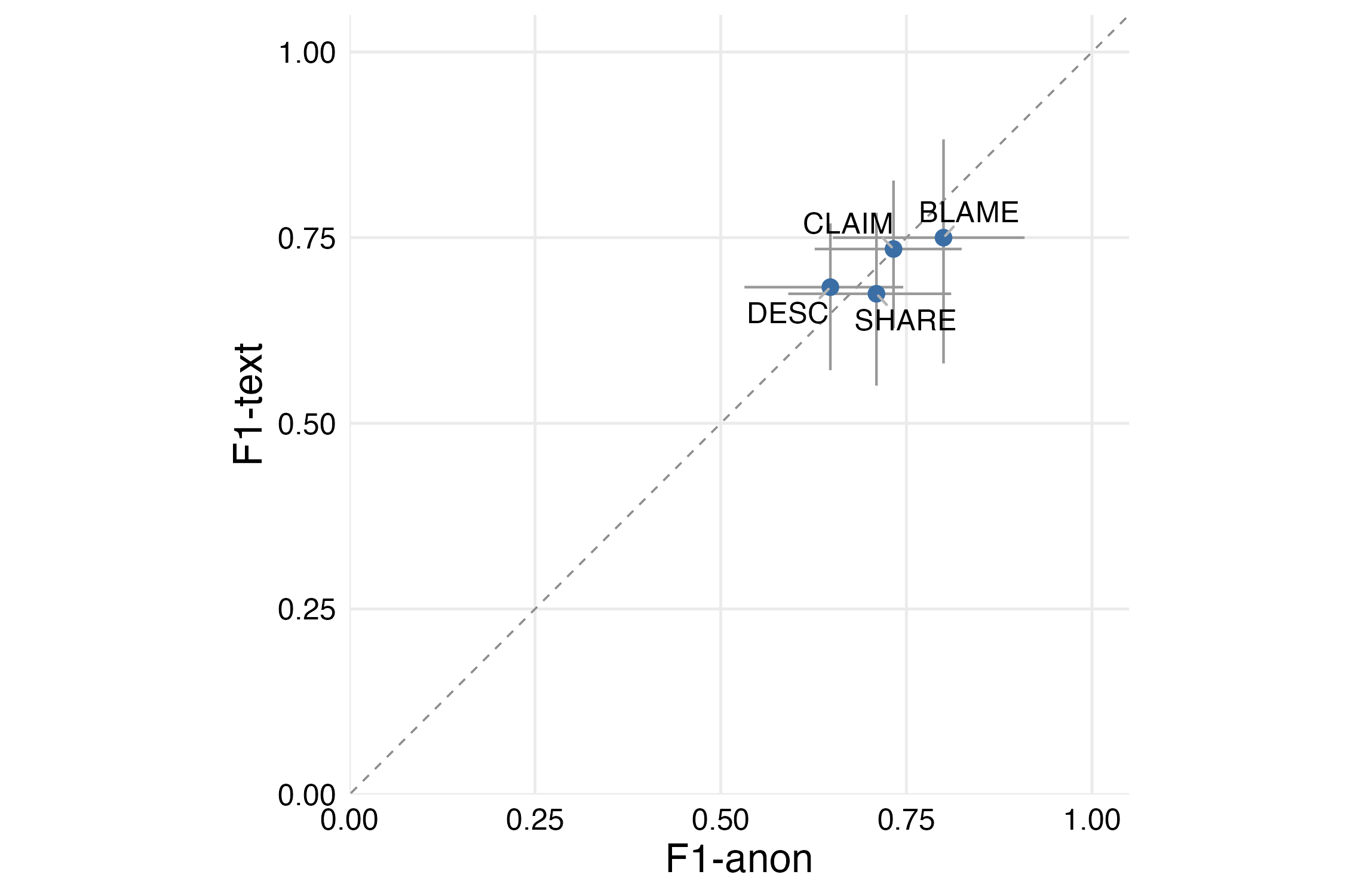}
    \caption{Credit claiming.}
\end{subfigure}
\caption{\textbf{Label-name ablation for the original codebooks on $D_{\mathrm{dev}}$.}}
\notes{Each panel compares per-class F1 with semantic class names, $F1^{\mathrm{text}}_k$, against F1 with anonymized labels, $F1^{\mathrm{anon}}_k$. The diagonal marks equal performance; points above it indicate higher F1 when semantic class names are available. Error bars show bootstrap 95\% confidence intervals. In BFRS, \textsc{Other} lies above the diagonal, indicating that its semantic name is helpful rather than clearly misleading. The credit-claiming classes lie closer to the diagonal, providing little evidence that label-name semantics drive the main confusion pattern. The ablation is used as a diagnostic of label-name dependence, not as a test of conceptual validity.}
\label{fig:label-name-ablation}
\end{figure}

\subsection{BFRS per-class F1 with anonymous label names}
\label{app:bfrs-id-mode}

\begin{figure}
    \centering
    \includegraphics[width=1\linewidth]{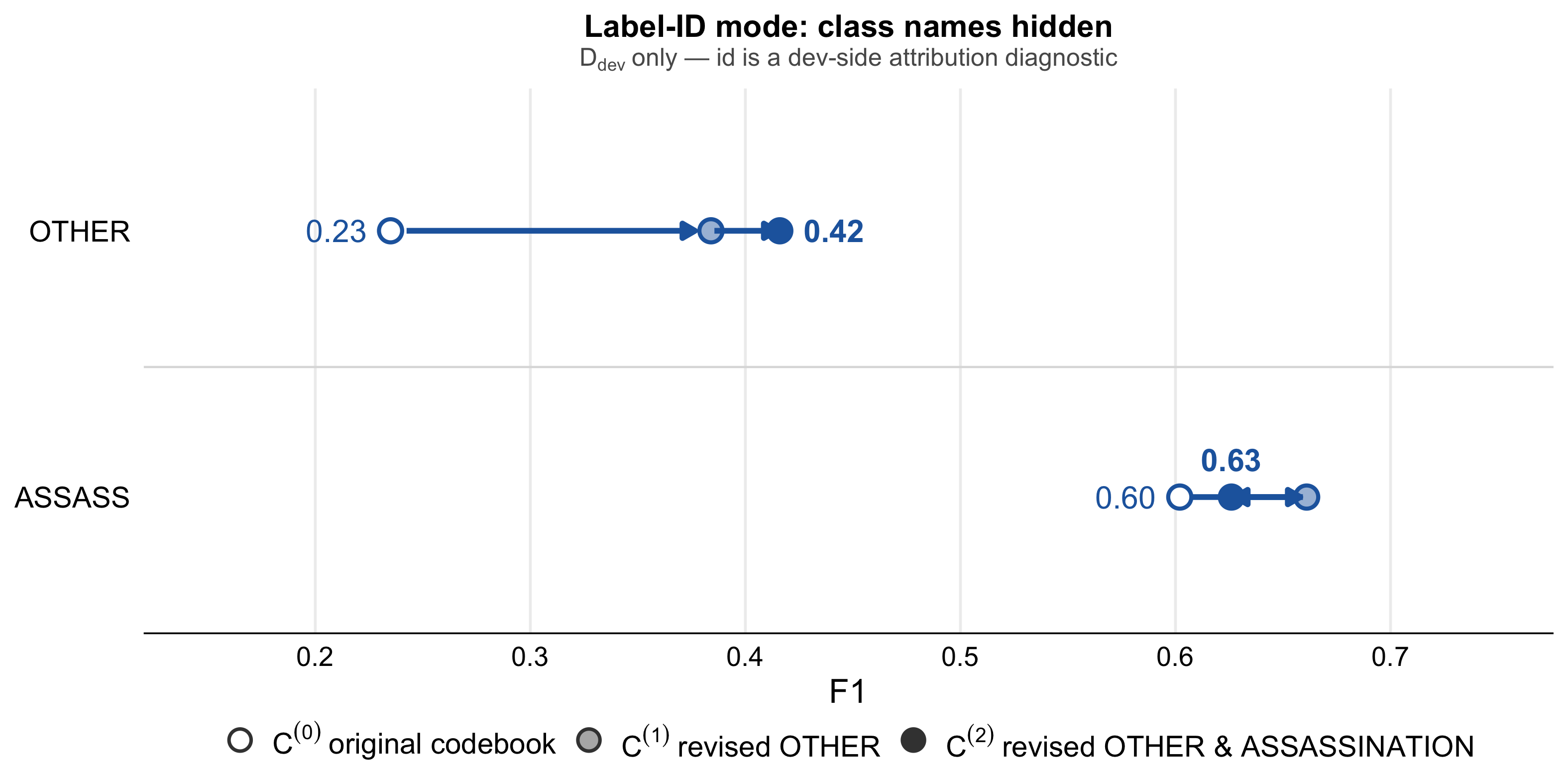}
    \caption{F1 Changes in OTHER and ASSASSINATION in ID-Mode across Codebook Revisions}
    \label{fig:bfrs_idladder}
\end{figure}

Table ~\ref{tab:bfrs-id} displays the changes of per-class F1 in anonymous label name. \textsc{other} climbs furthest and its interval clears zero, so the recovery is carried by the revised definition and clarification text rather than by the class name. \textsc{guerilla\_attack\_on\_gov\_forces} also clears zero although its entry was never edited, a spillover from the \textsc{other} rewrite. \textsc{assassination} barely moves. The $C^{(2)}$ edit to \textsc{assassination} plays a different role in the audit. Its purpose is to stop the \textsc{assassination} entry from capturing \textsc{other} documents, so its intended effect appears in \textsc{other}'s recall; \textsc{assassination} itself is expected only to hold steady. \textsc{riot} does not increase and its text-condition recovery is a spillover of the \textsc{other} edit. Figure \ref{fig:bfrs_idladder} highlights F1 changes for revised \textsc{other} and \textsc{assassination}.

\begin{table}[!htbp]
\centering
\small
\caption{BFRS Per-Class F1 with anonymous label name, $D_{\mathrm{dev}}$, $n=573$)}
\label{tab:bfrs-id}
\begin{tabular}{l r r r r l}
\toprule
Class & $n$ & $C^{(0)}$ & $C^{(1)}$ & $C^{(2)}$ & $\Delta$ id ($95\%$ CI) \\
\midrule
\textsc{other} & 50 & $0.235$ & $0.384$ & $0.416$ & $\mathbf{+0.180}$ $[+0.050,+0.319]$ \\
\textsc{guerilla\_attack\_on\_gov\_forces} & 50 & $0.667$ & $0.674$ & $0.745$ & $\mathbf{+0.078}$ $[+0.021,+0.151]$ \\
\textsc{terrorism} & 50 & $0.539$ & $0.587$ & $0.607$ & $+0.067$ $[-0.010,+0.150]$ \\
\textsc{conventional\_attack\_on\_gov\_forces} & 50 & $0.677$ & $0.682$ & $0.710$ & $+0.033$ $[-0.005,+0.074]$ \\
\textsc{assassination} & 50 & $0.602$ & $0.661$ & $0.626$ & $+0.024$ $[-0.062,+0.110]$ \\
\textsc{attack\_on\_state} & 50 & $0.906$ & $0.914$ & $0.914$ & $+0.009$ $[+0.000,+0.029]$ \\
\textsc{gov\_attack\_on\_nonstate\_combatants} & 50 & $0.743$ & $0.757$ & $0.750$ & $+0.007$ $[+0.000,+0.023]$ \\
\textsc{drone\_assassination} & 43 & $1.000$ & $1.000$ & $1.000$ & $+0.000$ $[+0.000,+0.000]$ \\
\textsc{gov\_attack\_on\_civilians} & 50 & $0.560$ & $0.568$ & $0.560$ & $+0.000$ $[+0.000,+0.000]$ \\
\textsc{violent\_political\_demonstration} & 50 & $0.754$ & $0.732$ & $0.732$ & $-0.022$ $[-0.058,+0.007]$ \\
\textsc{threat\_of\_violence} & 30 & $0.776$ & $0.776$ & $0.750$ & $-0.026$ $[-0.090,+0.000]$ \\
\textsc{riot} & 50 & $0.804$ & $0.800$ & $0.776$ & $-0.028$ $[-0.075,+0.011]$ \\
\midrule
macro & 573 & $0.689$ & $0.711$ & $0.715$ & $+0.027$ \\
\bottomrule
\end{tabular}
\notes{Deployment-model (Llama) F1 with every class \emph{name} hidden. $\Delta$ is the paired document bootstrap of the last revision against the first ($B=2000$), $95\%$ CI in brackets, \textbf{bold} where the interval strictly excludes zero.}
\end{table}

\newpage
\subsection{Credit-Claiming per-class F1 with anonymous label names}
\label{app:credit-id-mode}

\begin{figure}[!htbp]
    \centering
    \includegraphics[width=0.8\linewidth]{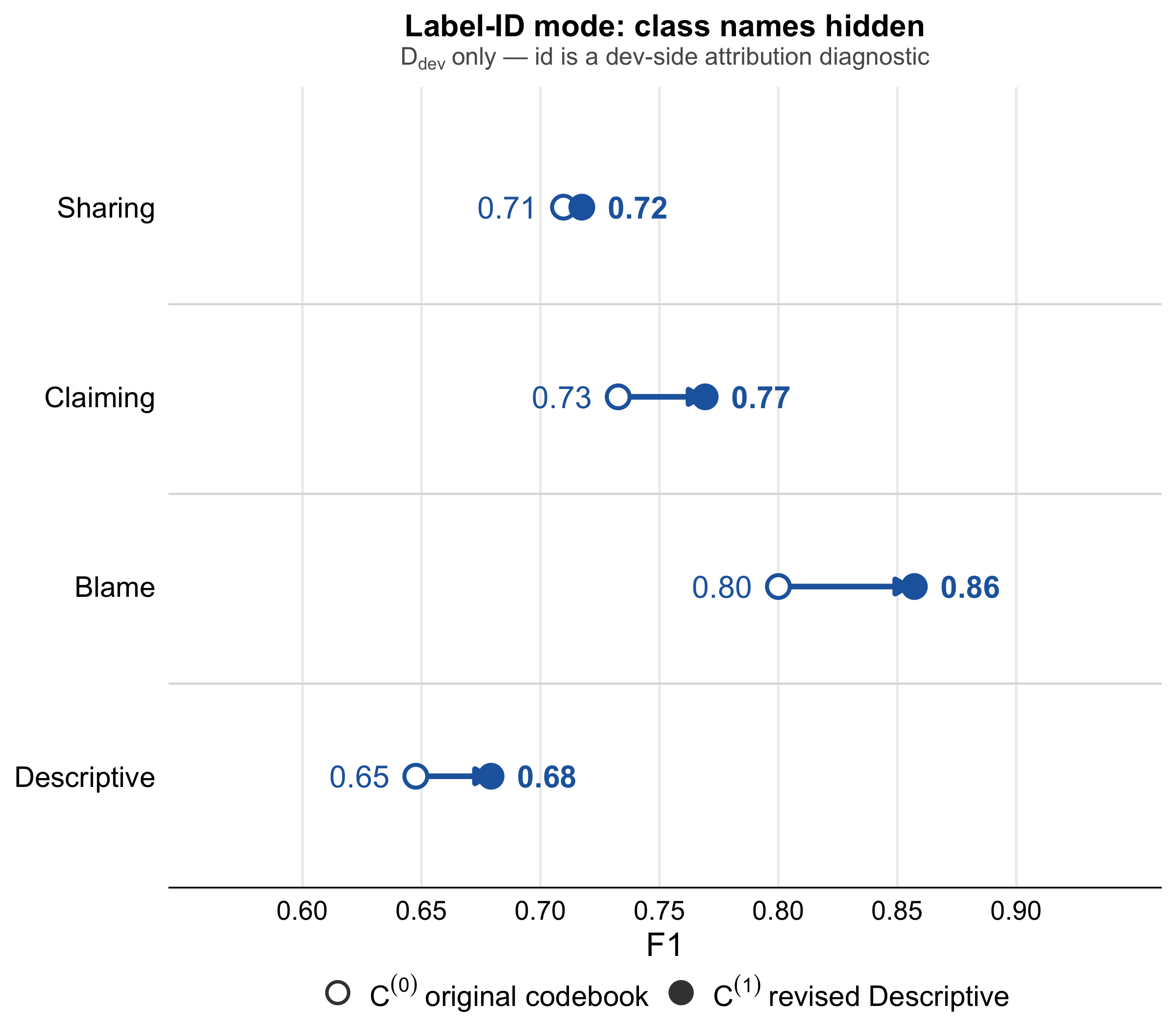}
    \caption{F1 Changes in ID-Mode across Codebook Revisions}
    \label{fig:credit_idladder}
\end{figure}

Table ~\ref{tab:hunter-id} and Figure ~\ref{fig:credit_idladder}  display the changes of per-class F1 in anonymous label name. F1 changes in all classes show the expected direction. 

\begin{table}[!htbp]
\centering
\small
\caption{Credit-claiming ladder with anonymous label names, $D_{\mathrm{dev}}$, $n=172$)}
\label{tab:hunter-id}
\begin{tabular}{l r r r l}
\toprule
Class & $n$ & $C^{(0)}$ & $C^{(1)}$ & $\Delta$ id ($95\%$ CI) \\
\midrule
Blame-shifting & 22 & $0.800$ & $0.857$ & $+0.057$ $[+0.000,+0.133]$ \\
Claiming & 50 & $0.733$ & $0.769$ & $+0.037$ $[+0.000,+0.082]$ \\
Descriptive & 50 & $0.648$ & $0.679$ & $+0.032$ $[-0.006,+0.076]$ \\
Sharing & 50 & $0.710$ & $0.717$ & $+0.008$ $[+0.000,+0.026]$ \\
\midrule
macro & 172 & $0.722$ & $0.756$ & $+0.033$ \\
\bottomrule
\end{tabular}
\notes{Deployment-model (Llama) F1 with anonymous label names. $\Delta$ is the paired document bootstrap of the last revision against the first ($B=2000$), $95\%$ CI in brackets, \textbf{bold} where the interval strictly excludes zero.}
\end{table}

\section{BFRS Per-class Decomposition}
\label{app:finetune}
Table~\ref{tab:hk-3codebooks-573} reports the full per-class decomposition, F1, precision, and recall, for the fine-tuned model and for our annotator under the four rungs of the \textsc{other} ladder (full codebook with no RAG; \texttt{+RAG} baseline; \texttt{+RAG} OTHER revision, and \texttt{+RAG} OTHER + Assassination revision) on that shared sample.

\begin{table}[!htbp]
\centering
\caption{Fine-tuned Mistral-7B vs.\ our annotator}
\label{tab:hk-3codebooks-573}
\resizebox{\linewidth}{!}{%
\begin{tabular}{lrrrrrrrrrrrrrrrr}
\toprule
 & & \multicolumn{3}{c}{ft (fine-tuned)} & \multicolumn{3}{c}{full codebook (no RAG)} & \multicolumn{3}{c}{\textsc{base} (+RAG)} & \multicolumn{3}{c}{\textsc{other\_rev} (+RAG)} & \multicolumn{3}{c}{\textsc{other\_assass} (+RAG)} \\
\cmidrule(lr){3-5}\cmidrule(lr){6-8}\cmidrule(lr){9-11}\cmidrule(lr){12-14}\cmidrule(lr){15-17}
Category & $n$ & F1 & P & R & F1 & P & R & F1 & P & R & F1 & P & R & F1 & P & R \\
\midrule
\textsc{drone\_assassination}                  & 43 & 1.000 & 1.00 & 1.00 & 1.000 & 1.00 & 1.00 & 1.000 & 1.00 & 1.00 & 1.000 & 1.00 & 1.00 & 1.000 & 1.00 & 1.00 \\
\textsc{attack\_on\_state}                     & 50 & 0.970 & 0.96 & 0.98 & 0.923 & 0.89 & 0.96 & 0.923 & 0.89 & 0.96 & 0.923 & 0.89 & 0.96 & 0.923 & 0.89 & 0.96 \\
\textsc{riot}                                  & 50 & 0.854 & 0.83 & 0.88 & 0.709 & 0.97 & 0.56 & 0.667 & 0.87 & 0.54 & 0.750 & 0.87 & 0.66 & 0.764 & 0.87 & 0.68 \\
\textsc{guerilla\_attack\_on\_gov\_forces}     & 50 & 0.851 & 0.91 & 0.80 & 0.760 & 0.76 & 0.76 & 0.764 & 0.87 & 0.68 & 0.773 & 0.89 & 0.68 & 0.791 & 0.88 & 0.72 \\
\textsc{gov\_attack\_on\_nonstate\_combatants} & 50 & 0.841 & 0.79 & 0.90 & 0.702 & 0.75 & 0.66 & 0.729 & 0.68 & 0.78 & 0.755 & 0.71 & 0.80 & 0.757 & 0.74 & 0.78 \\
\textsc{violent\_political\_demonstration}     & 50 & 0.815 & 0.76 & 0.88 & 0.803 & 0.70 & 0.94 & 0.800 & 0.71 & 0.92 & 0.783 & 0.69 & 0.90 & 0.776 & 0.68 & 0.90 \\
\textsc{conventional\_attack\_on\_gov\_forces} & 50 & 0.809 & 0.86 & 0.76 & 0.672 & 0.59 & 0.78 & 0.695 & 0.60 & 0.82 & 0.733 & 0.63 & 0.88 & 0.727 & 0.62 & 0.88 \\
\textsc{gov\_attack\_on\_civilians}            & 50 & 0.805 & 0.95 & 0.70 & 0.587 & 0.88 & 0.44 & 0.597 & 0.85 & 0.46 & 0.568 & 0.88 & 0.42 & 0.556 & 0.91 & 0.40 \\
\textsc{threat\_of\_violence}                  & 30 & 0.800 & 1.00 & 0.67 & 0.830 & 0.96 & 0.73 & 0.808 & 0.95 & 0.70 & 0.808 & 0.95 & 0.70 & 0.808 & 0.95 & 0.70 \\
\textsc{terrorism}                             & 50 & 0.783 & 0.69 & 0.90 & 0.631 & 0.51 & 0.82 & 0.636 & 0.58 & 0.70 & 0.660 & 0.62 & 0.70 & 0.673 & 0.67 & 0.68 \\
\textsc{assassination}                         & 50 & 0.682 & 0.55 & 0.90 & 0.614 & 0.61 & 0.62 & 0.621 & 0.60 & 0.64 & 0.653 & 0.65 & 0.66 & 0.686 & 0.67 & 0.70 \\
\textsc{other}                                 & 50 & \textbf{0.232} & 0.42 & 0.16 & 0.352 & 0.39 & 0.32 & 0.404 & 0.39 & 0.42 & 0.491 & 0.46 & 0.52 & \textbf{0.569} & 0.53 & 0.62 \\
\midrule
\textbf{Weighted-F1} & 573 & \textbf{0.784} & & & 0.708 & & & 0.714 & & & 0.736 & & & 0.748 & & \\
\textbf{Macro-F1}    & 573 & \textbf{0.787} & & & \textbf{0.715} & & & \textbf{0.720} & & & \textbf{0.741} & & & \textbf{0.753} & & \\
\bottomrule
\end{tabular}%
}
\notes{ft is QLoRA Mistral-7B-Instruct-v0.2 rank${}={}$16/all (full-dev weighted-F1 $0.821$) from \textcite{Halterman2025codebook}, scored on our $573$. The four right-hand groups are all Llama-3.3-70B, zero-shot, no weight updates, and differ only in retrieval and revision: full codebook (no RAG) injects the whole codebook with no retrieval; \textsc{base} adds RAG on the unrevised baseline;\textsc{other\_rev} (+RAG) adds RAG on the one-sided revision; \textsc{other\_assass} adds RAG on the two-sided revision. }
\end{table}

\newpage

\section{Credit-claiming per class IRR changes across different codebooks}
\label{sec:irr}

\begin{figure}[!htbp]
\centering
\includegraphics[width=0.75\textwidth,trim = 0.2cm 0cm 0cm 0cm, clip=true]{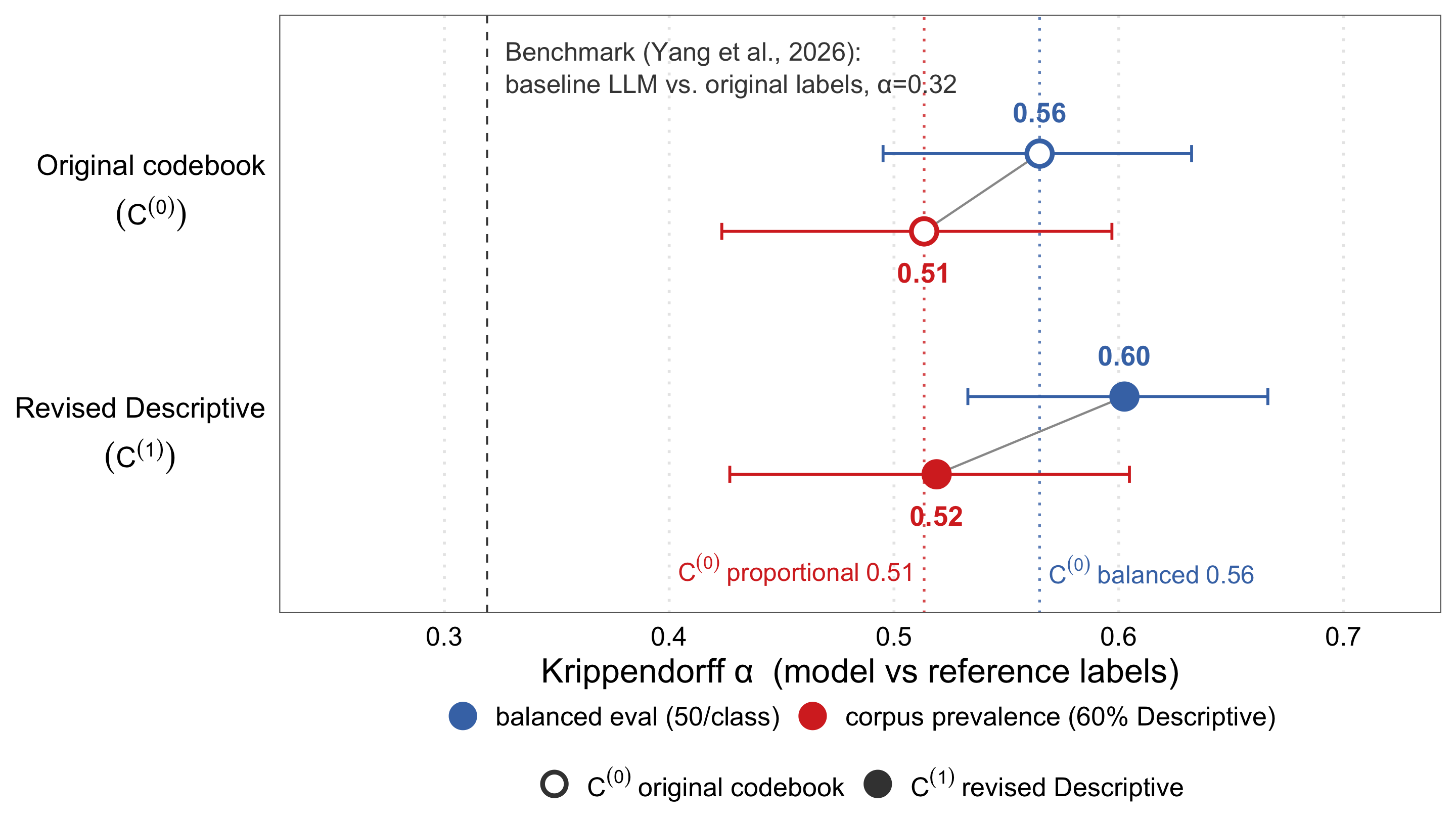}
\caption{Credit-claiming revision on IRR}
\notes{Open and filled points denote performance under the original codebook ($C^{(0)}$), and the revised \textsc{Descriptive} codebook ($C^{(1)}$). Color gives
  the class weighting. Blue is the evaluation on sample with balanced classes; red reweights the same predictions to the corpus class distribution. The dashed black vertical line is the benchmark from \textcite{Yang2025}: a baseline LLM against the original labels at corpus prevalence. Whiskers are bootstrap $95\%$ confidence intervals ($5{,}000$ resamples).}
\label{fig:hunter-alpha}
\end{figure}

\section{Stability Test Using 20 Documents Per Class}
\label{app:stabtest}

We assess whether the main diagnostic decisions from Steps~1--2 remain stable when the development sample is reduced to 20 labeled documents per class. For each application, we draw 20 documents per class without replacement from $D_{\mathrm{dev}}$, recompute the screening metrics and confusion patterns, and repeat the procedure 200 times.

\begin{table}[!htbp]
  \centering
  \caption{Diagnostic-pattern reproduction at $20$ labeled documents per class.}
  \label{tab:stability20}
  \begin{tabular}{llr}
  \toprule
  Codebook & Audit decision & Reproduced \\
  \midrule
  BFRS & \textsc{other} (lowest F1)                           & $96.0\%$ \\
       & Localize: gold \textsc{other} flows to \textsc{assassination}  & $97.0\%$ \\
  \midrule
  Credit-claim &  \textsc{Descriptive} (largest recall-over-precision excess) & $100\%$ \\
       & Localize: \textsc{Descriptive} receives more than it loses      & $100\%$ \\
  \bottomrule
  \end{tabular}
  \end{table}
  
\begin{figure}[!htbp]
\centering
\includegraphics[width=0.8\textwidth]{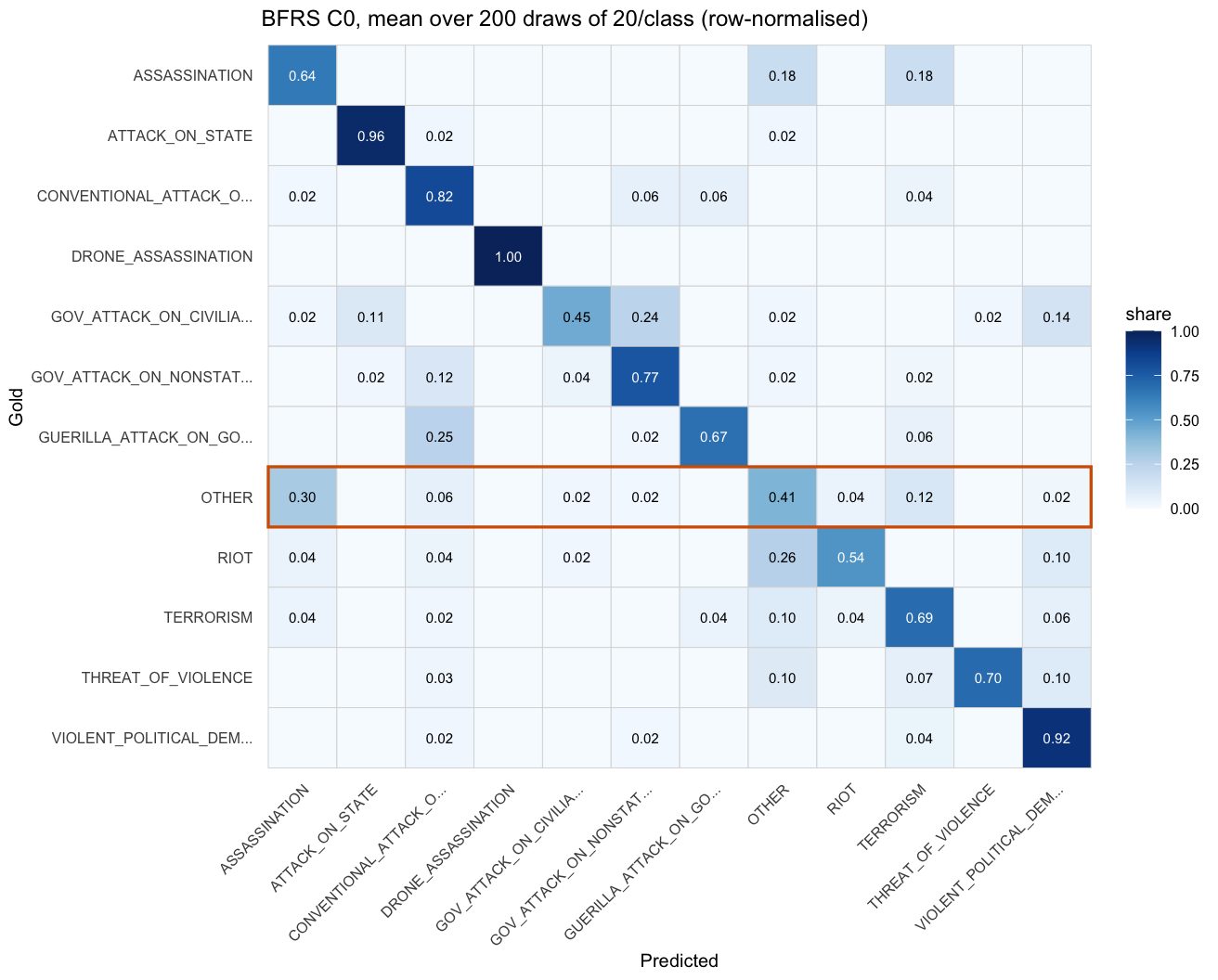} \\
\includegraphics[width=0.8\textwidth]{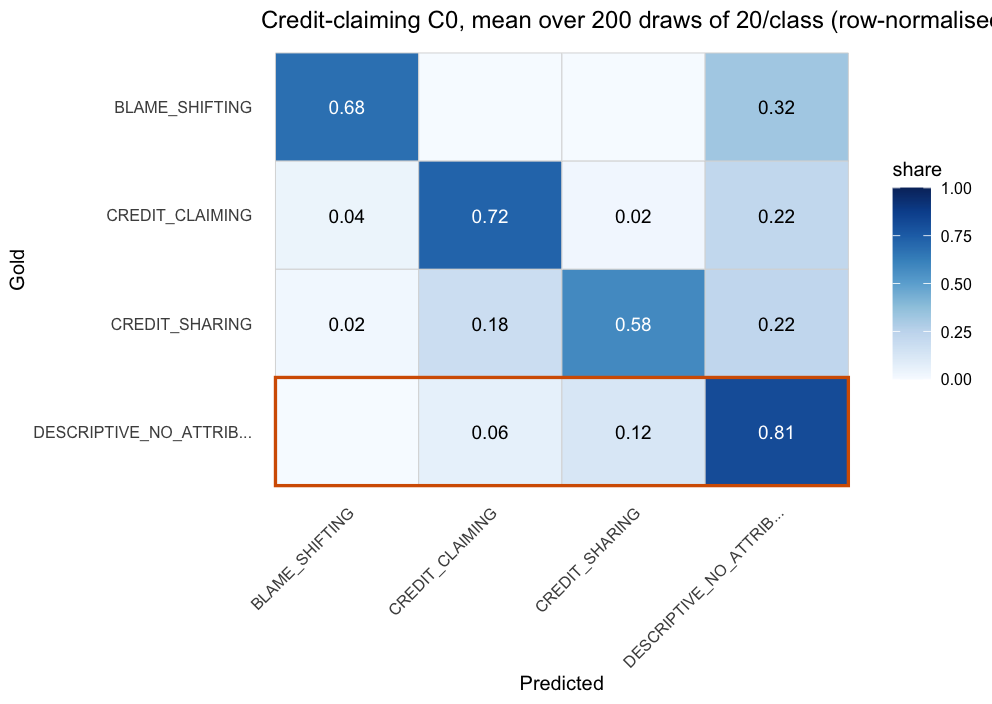}
\caption{Stability under 20 labeled documents per class for BFRS and credit claiming.}
\label{fig:dev20}
\end{figure}

For BFRS, \textsc{Other} remains the lowest-F1 class in 96.0\% of draws, with a median F1 gap of 0.156 from the second-lowest class. The main localization also persists: the concentration of reference \textsc{Other} errors in \textsc{Assassination}, the boundary targeted by $C^{(2)}$, is reproduced in 97.0\% of draws. For credit claiming, \textsc{Descriptive} retains the largest recall-over-precision asymmetry in 100\% of draws, and its over-prediction pattern is localized in 100\% of draws. Table~\ref{tab:stability20} summarizes these results and Figure~\ref{fig:dev20} displays the per-class mean precision-recall metrics over 200 draws.

These results show that the principal nomination and localization decisions are stable under substantial subsampling in our two applications. They should not be interpreted as a general claim that 20 documents per class are sufficient for every codebook: class-specific estimates are necessarily less precise at this size, and larger held-out samples are preferable for evaluating revision effects. Appendix~\ref{app:power} discusses validation sample-size considerations.

\newpage

\section{Codebook Revision Wordings}
\label{app:revised-codebook}

The two BFRS variants share identical revised \textsc{other} text, so \texttt{other\_assass}${-}$\texttt{other\_rev} isolates the \textsc{assassination} clause. \textsc{Riot} is not edited; its improvement is a spillover from clarifying the \textsc{Other} boundary. Each edit touches only its named class's \texttt{definition} and \texttt{clarification}; the positive and negative examples and every other class entry are identical, so the per-class delta is attributable to the wording change alone. 

The three BFRS versions at a glance. 
\begin{center}
\begin{tabular}{lll}
\toprule
version & \textsc{other} entry & \textsc{assassination} \texttt{clarification} \\
\midrule
\texttt{base} (original) & original & original \\
\texttt{other\_rev} & revised (affirmative assignment rule) & original (unchanged) \\
\texttt{other\_assass} & revised ($=$ \texttt{other\_rev}) & revised (targeting condition) \\
\bottomrule
\end{tabular}
\end{center}

\subsection{BFRS \textsc{other} (\texttt{other\_rev})}

The audit nominated \textsc{other} as a weak fallback class with errors in both directions. The revision replaces the open ``does not fit the other categories'' default with an affirmative construct-absent condition. 

\paragraph{\textsc{other} in \texttt{base} (original, verbatim).}
\begin{description}
\item[\texttt{definition}] Incidents that are not easily categorized such as
people being hurt by unexploded ordnance or when the source of the violence is
unclear.
\item[\texttt{clarification}] An event that does not fit into the other categories (ASSASSINATION, ATTACK\_ON\_STATE, CONVENTIONAL\_ATTACK\_ON\_GOV\_FORCES, \\GUERILLA\_ATTACK\_ON\_GOV\_FORCES, GOV\_ATTACK\_ON\_CIVILIANS) should be coded as OTHER.
\end{description}

\paragraph{\textsc{other} in \texttt{other\_rev} and \texttt{other\_assass}
(revised).}
\begin{description}
\item[\texttt{definition}] An incident that matches no substantive class. This happens in two situations. First, when the report does not say enough about the perpetrator and the target to assign any class, for example when the source of the violence is unclear, or when a body is found with no sign of who was responsible or that the victim was targeted. Second, when the event falls outside every substantive category, for example when people are hurt by unexploded ordnance. Do not use OTHER as a default when a substantive class plausibly applies.

\item[\texttt{clarification}] Code OTHER only when no substantive class fits. Two cases qualify. (1)~The report does not establish the kind of perpetrator and gives no indication that the victim was deliberately targeted. For example, when a body is found but the report does not say who was responsible and does not show that the victim was singled out, code OTHER. When the kind of actor and target is clear, use the substantive class instead: if unidentified militants bombed a police post, the actor (militants) and the target (security forces) are clear, so it is GUERILLA\_ATTACK\_ON\_GOV\_FORCES. (2)~The event does not belong to any substantive category, for example harm from unexploded ordnance or an accidental detonation. Note: a clash between groups is RIOT even if it is not political, and a deliberate killing of a specifically targeted individual is ASSASSINATION; neither is OTHER.

\item[\texttt{negative\_clarification}] An event that does not fit into the other categories (
ASSASSINATION, ATTACK\_ON\_STATE, CONVENTIONAL\_ATTACK\_ON\_GOV\_FORCES, \\GUERILLA\_ATTACK\_ON\_GOV\_FORCES, GOV\_ATTACK\_ON\_CIVILIANS) should be coded as OTHER.
\end{description}

\subsection{BFRS - (\textsc{assassination})}

The \textsc{Other}--\textsc{Assassination} confusion is partly driven by \textsc{Assassination}'s own \texttt{clarification}, whose final sentence (``a body found with bullet wounds\ldots'') is broader than its targeting \texttt{definition}. This edit conditions only that last sentence on deliberate targeting; the \texttt{definition}, \texttt{negative\_clarification}, and examples are unchanged. 

\paragraph{\textsc{assassination} \texttt{clarification} in \texttt{base} and
\texttt{other\_rev} (original --- unchanged in both).}
\begin{quote}
Targets may include military, political, civil society or civilian state or federal leadership. In some cases, there may be a suspected government agency behind the assassination. This Label applies when a body is found with bullet wounds or other marks of violence and/or torture.
\end{quote}

\paragraph{\textsc{assassination} \texttt{clarification} in \texttt{other\_assass}
(revised --- \texttt{definition} unchanged; only the last sentence conditioned).}
\begin{quote} Targets may include military, political, civil society or civilian state or federal leadership. In some cases, there may be a suspected government agency behind the assassination. A body found with bullet wounds or other marks of violence indicates ASSASSINATION when the report shows the victim was deliberately singled out, for example a named or identified person, a victim attacked at home or workplace, or a person killed for a stated personal, family, or political reason. When a body is found in unexplained circumstances, with no information about who was responsible and no indication the victim was deliberately targeted, code OTHER instead.
\end{quote}

\subsection{Credit-claim- Descriptive (no attribution): the tense discriminator}

The audit nominated Descriptive because it is over-predicted and its \texttt{clarification} conflicts with the broader coding guidance: it lists prospective ``must''/``should'' statements as non-attribution, while the codebook's coding guidance treats prospective attributions as credit. The revision (\textsc{des\_revise}) appends a single discriminator to the \texttt{clarification}: credit is decided by agency (who is credited), not tense. The \texttt{definition}, \texttt{negative\_clarification}, and examples are unchanged.

\paragraph{Original.}
\begin{description}
\item[\texttt{definition}] DESCRIPTIVE\_NO\_ATTRIBUTION refers to statements where none of the above is present. These will largely be descriptive statements, where there is no attribution for credit and blame.
\item[\texttt{clarification}] This includes, but is not limited to:
\begin{itemize}
\item National leaders listing the policy issues and items that were discussed and/or noted in the EUCO meeting;
\item The announcement of a policy or decision without clear attribution for responsibility, nor an explanation of how the decision will benefit Europe or a member state;
\item Expressions of desired actions or desired policy outcomes, rather than ones that have actually been taken (e.g.\ Europe ``must'' or ``should'' do something);
\item National leaders announcing that the European Council has discussed, made a statement on a policy issue or reaffirmed its stance on said issue.
\end{itemize}
\end{description}

\paragraph{Revised.} The \texttt{definition} above is unchanged; the four
\texttt{clarification} bullets are unchanged and the revision appends the following two sentences to the \texttt{clarification}:
\begin{quote}
A statement expressing desired actions or outcomes (``must''/``should'') is descriptive only when it credits no specific actor. A future-oriented statement that credits the speaker's own government or country is credit claiming, and one that credits the EU, its institutions, or partners is credit sharing, not descriptive.
\end{quote}

\subsection{Credit-claim- Credit-sharing: the same-outcome discriminator}

The Gemma annotator's diagnosis nominated Credit-sharing through the asymmetry screen (precision $0.73$, recall $0.54$, the opposite asymmetry from the Llama annotator's Descriptive), with reference Credit-sharing statements flowing mainly to Credit-claiming and Descriptive. Document review traced the pattern to jointly framed credit. The codebook specifies how to code credit attributed to the speaker’s own government and credit attributed to the EU or other member states, but not how to code statements that attribute the same outcome to both. The candidate revision ($C^{(2)}$) retains the revised Descriptive entry and edits only the CREDIT\_SHARING entry. This clarification was developed from the Gemma annotator's error pattern and evaluated with both annotators as a cross-annotator transfer test; it is not part of the deployed Llama revision sequence.

\paragraph{Revised ($C^{2}$)} The \texttt{definition} and the five \texttt{clarification} bullets above are unchanged; the revision appends the following two sentences to the \texttt{clarification}:

\begin{quote}
A statement that credits the same positive decision, agreement, or outcome both to the speaker's own government or country and to the EU, its institutions, or other member states is credit sharing, not credit claiming, even when the speaker also highlights their own proposals, negotiations, or contribution to that outcome. Credit sharing of this joint kind is not limited to sharing with one specific other member state; it includes joint credit with the EU, its institutions, or member states collectively.
\end{quote}

It also adds the following \texttt{negative\_clarification} (a new field for this entry):
\begin{quote}
When a statement credits the positive outcome to the speaker's own side alone --- such as an outcome secured specifically for their country, or an idea they have long defended that has become EU policy --- and the EU, the Council, or other member states appear only as the venue, the audience, or the body the speaker acted upon, the statement is credit claiming; mentioning the EU, the Council, or member states in that role is not crediting them.
\end{quote}

\section{Boundary cases for the two unrevised confusions}
\label{app:goldcases}

\subsection{\textsc{gov\_attack\_on\_civilians} vs.\ \textsc{gov\_attack\_on\_nonstate\_combatants}}

Near-identical reports of security forces destroying suspected militants' property receive opposite reference labels. None of the six documents reports a casualty or a two-sided engagement.

\begin{center}
  \small
  \begin{tabular}{p{1.6cm}p{8.6cm}p{2.6cm}}
  \hline
  Document & Text & Reference label \\
  \hline
  bfrs\_02549 & ``Police and Elite Force destroyed 4 hideouts of TTP commander Taj Jalal of
  the Maulvi Hanafi group.'' & \textsc{civilians} \\
  bfrs\_01375 & ``At least three hideouts of militants were destroyed by helicopter gunships
  in an operation by the security forces in the areas of Mera Sar and Baz Garha.'' &
  \textsc{nonstate} \\
  \hline
  bfrs\_01980 & ``Several houses of suspected militants were demolished as security forces
  continued operation in different areas of Baizai tehsil of Mohmand Agency.'' &
  \textsc{civilians} \\
  bfrs\_00158 & ``Jet fighters and helicopter gunships pounded suspected militant hideouts
  in the Bezai tehsil of Mohmand tribal region.'' & \textsc{nonstate} \\
  \hline
  bfrs\_02648 & ``The district polcie demolished hideouts belonging to outlaws' in a ten
  hour operation in Salar Bhattian village.'' & \textsc{civilians} \\
  bfrs\_04480 & ``helicopter gunships bombed several hideouts in South Waziristan but there
  were no casualties.'' & \textsc{nonstate} \\
  \hline
  \end{tabular}
  \end{center}

The model predicts \textsc{gov\_attack\_on\_nonstate\_combatants} for all six. The manual error analysis distributed with \textcite{Halterman2025codebook} flags a further document of the same pattern (``Security forces, backed by tanks and artillery, destroyed eight houses of suspected militants during a search operation\ldots,'' reference label \textsc{gov\_attack\_on\_civilians}; bfrs\_01902 in our development sample) with the note that the reference label is also suspect.''

\section{Behavior Across LLM Annotators: Gemma Results}
\label{app:gemma}
This appendix reports the full Gemma-annotator results summarized in the main text, following the same steps as the Llama-annotator audit. The Gemma annotator, Gemma-4-31B, runs the identical codebook ladder, prompts, and samples as the Llama annotator.

\subsection{BFRS}

\begin{figure}[!htbp]
\centering
\includegraphics[width=1\linewidth]{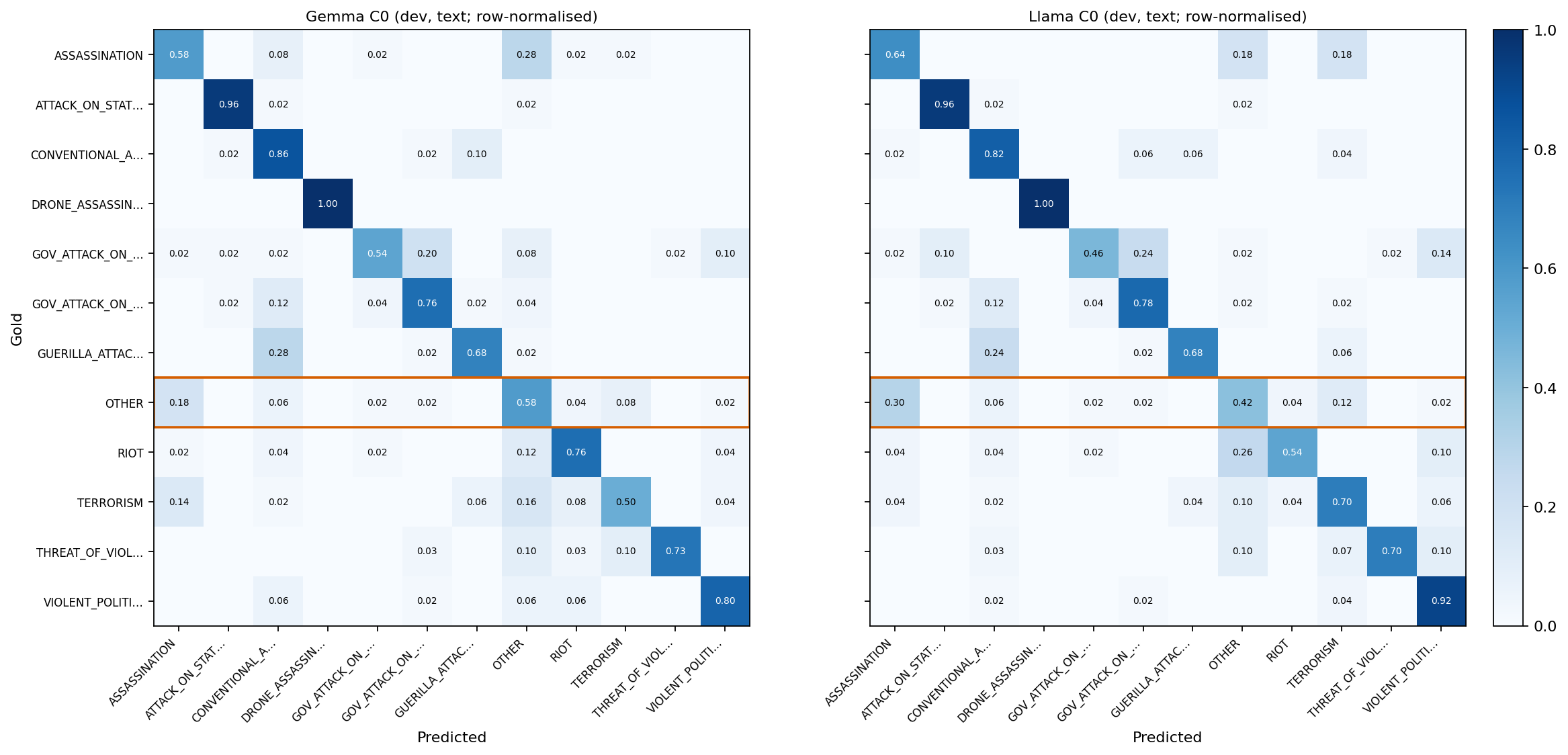}
\caption{BFRS confusion matrices for Llama and Gemma under the original codebook.}
\notes{Row-normalized confusion matrices on $D_{\mathrm{dev}}$ under the original BFRS codebook: Gemma annotator (Gemma, left) and Llama annotator (Llama, right). Rows denote reference labels and columns denote model predictions; diagonal entries are class-specific recall; the highlighted row is the lowest F1 class, \textsc{other}.}
\label{fig:gemma-bfrs-confusion}
\end{figure}

\paragraph{LLM annotation and class-wise performance.} The Gemma annotator annotates $D_{\mathrm{dev}}$ under the original codebook $C^{(0)}$ with the same retrieval configuration as the Llama annotator. \textsc{other} is again the weakest class. Figure~\ref{fig:gemma-bfrs-confusion} places the Gemma annotator's confusion matrix beside the Llama annotator's. 

\paragraph{Problematic classes and common confusions.}
In both panels the largest confusions involving \textsc{Other} are with \textsc{Assassination}. The \textsc{Riot}$\rightarrow$\textsc{Other} confusion, by contrast, is model-specific. The Gemma-annotator does not assign many \textsc{riot} cases to \textsc{other}. Full per-class precision, recall is in Table~\ref{tab:gemma-perclass-dev}. 

\begin{table}[!htbp]
\centering
\caption{Per-class and aggregate F1 on $D_{\mathrm{dev}}$ and the held-out validation sample}
\label{tab:gemma-bfrs-aggregate}
\resizebox{\textwidth}{!}{%
\begin{tabular}{llll}
\toprule
F1 & \textsc{base} & \texttt{other\_rev} ($\Delta$) & \texttt{other\_assass} ($\Delta$) \\
\midrule
\multicolumn{4}{l}{Dev ($n=573$, $50$/class)}\\
\textsc{other} & $0.479$ & $0.500$ ($+0.021$ $[-0.067,+0.110]$) & $0.504$ ($+0.025$ $[-0.077,+0.126]$) \\
\textsc{riot} & $0.768$ & $0.781$ ($+0.014$ $[-0.044,+0.076]$) & $0.781$ ($+0.014$ $[-0.044,+0.076]$) \\
\textsc{assassination} & $0.598$ & $0.553$ ($-0.045$ $[-0.110,+0.011]$) & $0.594$ ($-0.005$ $[-0.097,+0.088]$) \\
macro & $0.736$ & $0.730$ ($-0.005$ $[-0.021,+0.011]$) & $0.733$ ($-0.002$ $[-0.021,+0.017]$) \\
weighted & $0.729$ & $0.724$ ($-0.005$ $[-0.022,+0.011]$) & $0.726$ ($-0.002$ $[-0.022,+0.018]$) \\
\midrule
\multicolumn{4}{l}{Held-out ($n=1{,}045$, ${\sim}100$/class; \textsc{drone} $35$, \textsc{threat} $32$, \textsc{gov\_civ} $78$)}\\
\textsc{other} & $0.456$ & $0.508$ ($\mathbf{+0.052}$ $[+0.002,+0.105]$) & $0.509$ ($\mathbf{+0.052}$ $[+0.003,+0.103]$) \\
\textsc{riot} & $0.760$ & $0.832$ ($\mathbf{+0.072}$ $[+0.029,+0.119]$) & $0.840$ ($\mathbf{+0.080}$ $[+0.038,+0.126]$) \\
\textsc{assassination} & $0.685$ & $0.674$ ($-0.011$ $[-0.061,+0.036]$) & $0.688$ ($+0.003$ $[-0.059,+0.065]$) \\
macro & $0.734$ & $0.746$ ($\mathbf{+0.012}$ $[+0.001,+0.023]$) & $0.745$ ($+0.011$ $[-0.001,+0.023]$) \\
weighted & $0.723$ & $0.738$ ($\mathbf{+0.015}$ $[+0.003,+0.027]$) & $0.737$ ($\mathbf{+0.014}$ $[+0.001,+0.027]$) \\
\bottomrule
\end{tabular}%
}
\notes{Paired with-replacement bootstrap $\Delta$ against \textsc{base}, $B=5000$, $95\%$ CI, \textbf{bold} where the CI clears zero. Gemma uses the identical codebook versions, prompts, and samples. The held-out effects provide a cross-annotator transfer check.}
\end{table}

\paragraph{Error-source attribution.}
The overlap also appears at the document level, not only in aggregate rates. Consider one direction of the boundary. Among
the $N=50$ reference \textsc{assassination} documents in $D_{\mathrm{dev}}$,
the Llama annotator annotates $n=9$ as \textsc{other} and the Gemma annotator annotates $n=14$ as \textsc{other}. Eight of the Llama annotator's nine misclassified documents are also misclassified by the Gemma annotator. In addition, of the $N=50$ reference \textsc{other} documents, the Llama annotator annotates
$n=15$ as \textsc{assassination} and the Gemma annotator $n=9$, and all nine of the Gemma annotator's misclassifications are among the Llama annotator's fifteen. 

The overlap shows that many errors occur on the same documents, not merely at similar aggregate rates. Because the annotators receive the same documents, codebook, prompts, and reference labels, this recurrence is consistent with a source shared across annotators rather than a purely model-specific error. Document-level review is needed to attribute that source more specifically.

\begin{figure}[!htbp]
    \centering
    \includegraphics[width=1\linewidth,trim = 0cm 0cm 0cm 0cm, clip=true]{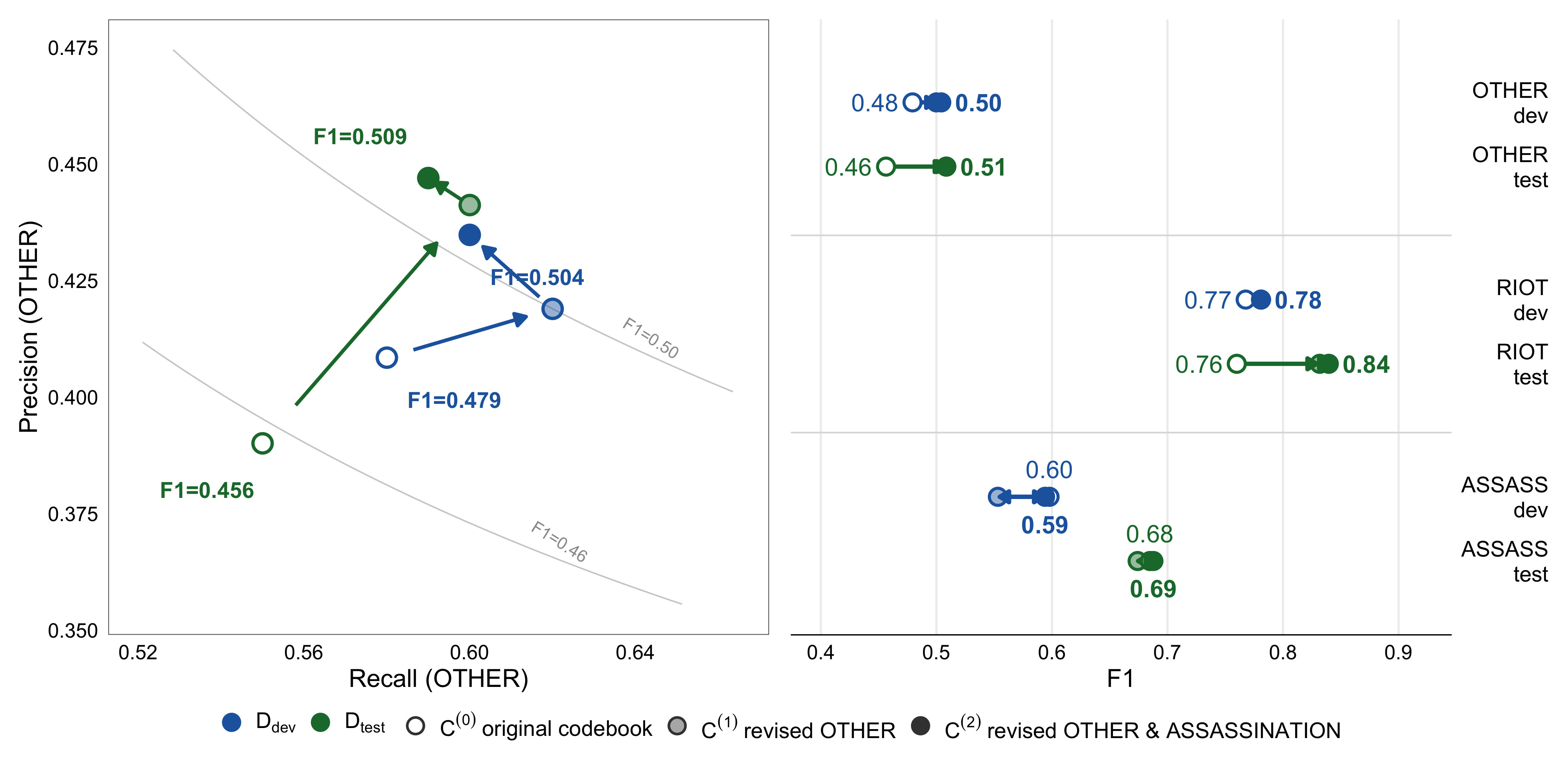}
    \caption{BFRS revision effects for the Gemma annotator}
    \notes{Blue indicates $D_{\mathrm{dev}}$; green indicates the held-out validation ($n=1{,}045$). Open, half-shaded, and filled points denote the original codebook ($C^{(0)}$), the revised \textsc{other} codebook ($C^{(1)}$),  and the revised \textsc{other} plus \textsc{assassination} codebook
    ($C^{(2)}$). Left panel: precision and recall for \textsc{other}; gray
    curves show constant-F1 contours, and arrows trace the revision
    sequence. Right panel: F1 for \textsc{other} and its neighboring classes
    \textsc{riot} and \textsc{assassination} across the same three
    codebooks.}
\label{fig:gemma-other-decomp}
\end{figure}

\paragraph{Held-out evaluation.}
The revisions evaluated are the Llama annotator's $C^{(1)}$ and $C^{(2)}$, unchanged (Figure~\ref{fig:gemma-other-decomp}). On $D_{\mathrm{dev}}$ every version moves in the predicted direction. For example, \textsc{other} $0.479\rightarrow0.500\rightarrow0.504$, driven by recall; \textsc{riot} $0.768\rightarrow0.781\rightarrow0.781$ with no edit to its entry; and \textsc{assassination} $0.598\rightarrow0.553\rightarrow0.594$. On the held-out validation set ($n=1{,}045$), the effects move in the same general direction, although the magnitudes differ from those for Llama. The \textsc{other} gain is $+0.052$ under both revisions, \textsc{riot} improves by $+0.072$ and $+0.080$ although no revision touches its entry, and \textsc{assassination} holds (Table~\ref{tab:gemma-bfrs-aggregate}). Full per-class precision, recall under every codebook version is in Table~\ref{tab:gemma-perclass-test}.

\begin{table}[!htbp]
\centering
\caption{BFRS per-class precision, recall, and F1 across codebook versions, $D_{\mathrm{dev}}$ ($n=573$)}
\label{tab:gemma-perclass-dev}
\resizebox{\textwidth}{!}{%
\begin{tabular}{lrrrrrrrrrr}
\toprule
 & & \multicolumn{3}{c}{$C^{(0)}$} & \multicolumn{3}{c}{$C^{(1)}$} & \multicolumn{3}{c}{$C^{(2)}$} \\
\cmidrule(lr){3-5} \cmidrule(lr){6-8} \cmidrule(lr){9-11}
Class & $n$ & P & R & F1 & P & R & F1 & P & R & F1 \\
\midrule
\textsc{drone\_assassination} & 43 & 1.00 & 1.00 & 1.000 & 1.00 & 1.00 & 1.000 & 1.00 & 1.00 & 1.000 \\
\textsc{attack\_on\_state} & 50 & 0.94 & 0.96 & 0.950 & 0.94 & 0.96 & 0.950 & 0.94 & 0.96 & 0.950 \\
\textsc{threat\_of\_violence} & 30 & 0.96 & 0.73 & 0.830 & 0.96 & 0.73 & 0.830 & 0.96 & 0.73 & 0.830 \\
\textsc{violent\_political\_demonstration} & 50 & 0.80 & 0.80 & 0.800 & 0.80 & 0.78 & 0.788 & 0.80 & 0.78 & 0.788 \\
\textsc{riot} & 50 & 0.78 & 0.76 & 0.768 & 0.75 & 0.82 & 0.781 & 0.75 & 0.82 & 0.781 \\
\textsc{gov\_attack\_on\_nonstate\_combatants} & 50 & 0.72 & 0.76 & 0.738 & 0.74 & 0.78 & 0.757 & 0.74 & 0.78 & 0.757 \\
\textsc{guerilla\_attack\_on\_gov\_forces} & 50 & 0.79 & 0.68 & 0.731 & 0.79 & 0.68 & 0.731 & 0.76 & 0.68 & 0.716 \\
\textsc{conventional\_attack\_on\_gov\_forces} & 50 & 0.55 & 0.86 & 0.672 & 0.54 & 0.86 & 0.667 & 0.54 & 0.86 & 0.667 \\
\textsc{gov\_attack\_on\_civilians} & 50 & 0.84 & 0.54 & 0.659 & 0.84 & 0.52 & 0.642 & 0.87 & 0.52 & 0.650 \\
\textsc{terrorism} & 50 & 0.76 & 0.50 & 0.602 & 0.79 & 0.44 & 0.564 & 0.84 & 0.42 & 0.560 \\
\textsc{assassination} & 50 & 0.62 & 0.58 & 0.598 & 0.59 & 0.52 & 0.553 & 0.59 & 0.60 & 0.594 \\
\textsc{other} & 50 & 0.41 & 0.58 & 0.479 & 0.42 & 0.62 & 0.500 & 0.43 & 0.60 & 0.504 \\
\bottomrule
\end{tabular}%
}
\end{table}

\begin{table}[!htbp]
\centering
\caption{BFRS per-class precision, recall, and F1 across codebook versions on the held-out validation sample ($n=1{,}045$)}
\label{tab:gemma-perclass-test}
\resizebox{\textwidth}{!}{%
\begin{tabular}{lrrrrrrrrrr}
\toprule
 & & \multicolumn{3}{c}{$C^{(0)}$} & \multicolumn{3}{c}{$C^{(1)}$} & \multicolumn{3}{c}{$C^{(2)}$} \\
\cmidrule(lr){3-5} \cmidrule(lr){6-8} \cmidrule(lr){9-11}
Class & $n$ & P & R & F1 & P & R & F1 & P & R & F1 \\
\midrule
\textsc{drone\_assassination} & 35 & 1.00 & 0.97 & 0.986 & 0.97 & 0.97 & 0.971 & 0.97 & 0.97 & 0.971 \\
\textsc{attack\_on\_state} & 100 & 0.93 & 0.96 & 0.946 & 0.93 & 0.96 & 0.946 & 0.93 & 0.96 & 0.946 \\
\textsc{violent\_political\_demonstration} & 100 & 0.78 & 0.83 & 0.802 & 0.79 & 0.85 & 0.821 & 0.79 & 0.85 & 0.817 \\
\textsc{gov\_attack\_on\_nonstate\_combatants} & 100 & 0.75 & 0.82 & 0.781 & 0.75 & 0.85 & 0.798 & 0.75 & 0.86 & 0.800 \\
\textsc{riot} & 100 & 0.76 & 0.76 & 0.760 & 0.78 & 0.89 & 0.832 & 0.79 & 0.89 & 0.840 \\
\textsc{terrorism} & 100 & 0.87 & 0.67 & 0.757 & 0.88 & 0.65 & 0.747 & 0.89 & 0.62 & 0.729 \\
\textsc{threat\_of\_violence} & 32 & 1.00 & 0.56 & 0.720 & 1.00 & 0.56 & 0.720 & 1.00 & 0.56 & 0.720 \\
\textsc{guerilla\_attack\_on\_gov\_forces} & 100 & 0.78 & 0.61 & 0.685 & 0.78 & 0.62 & 0.693 & 0.78 & 0.61 & 0.685 \\
\textsc{assassination} & 100 & 0.75 & 0.63 & 0.685 & 0.77 & 0.60 & 0.674 & 0.73 & 0.65 & 0.688 \\
\textsc{conventional\_attack\_on\_gov\_forces} & 100 & 0.56 & 0.85 & 0.677 & 0.58 & 0.86 & 0.691 & 0.58 & 0.86 & 0.694 \\
\textsc{gov\_attack\_on\_civilians} & 78 & 0.79 & 0.42 & 0.550 & 0.82 & 0.41 & 0.547 & 0.84 & 0.40 & 0.539 \\
\textsc{other} & 100 & 0.39 & 0.55 & 0.456 & 0.44 & 0.60 & 0.508 & 0.45 & 0.59 & 0.509 \\
\bottomrule
\end{tabular}%
}
\notes{The held-out validation sample ($n=1{,}045$, ${\sim}100$/class; \textsc{drone} $35$, \textsc{threat} $32$, \textsc{gov\_civ} $78$), sorted by $C^{(0)}$ F1. Paired bootstrap deltas with confidence intervals for the audited classes are in Table~\ref{tab:gemma-bfrs-aggregate}.}
\end{table}

\subsection{Credit Claiming}

\paragraph{LLM annotation and class-wise performance.}
The same two classes sit at the bottom of the scheme: \textsc{Credit-sharing} ($0.621$) and \textsc{Descriptive} ($0.667$), against $0.674$ and $0.683$ on the Llama annotator. Nomination, however, flips within that pair. The Llama annotator's audit nominates \textsc{Descriptive} on the asymmetry criterion (precision $0.59$ against recall $0.82$), whereas the Gemma annotator's weakest entry is \textsc{Credit-sharing}, with the opposite asymmetry (precision $0.73$ against recall $0.54$).

\begin{figure}[!htbp]
\centering
\includegraphics[width=1\linewidth]{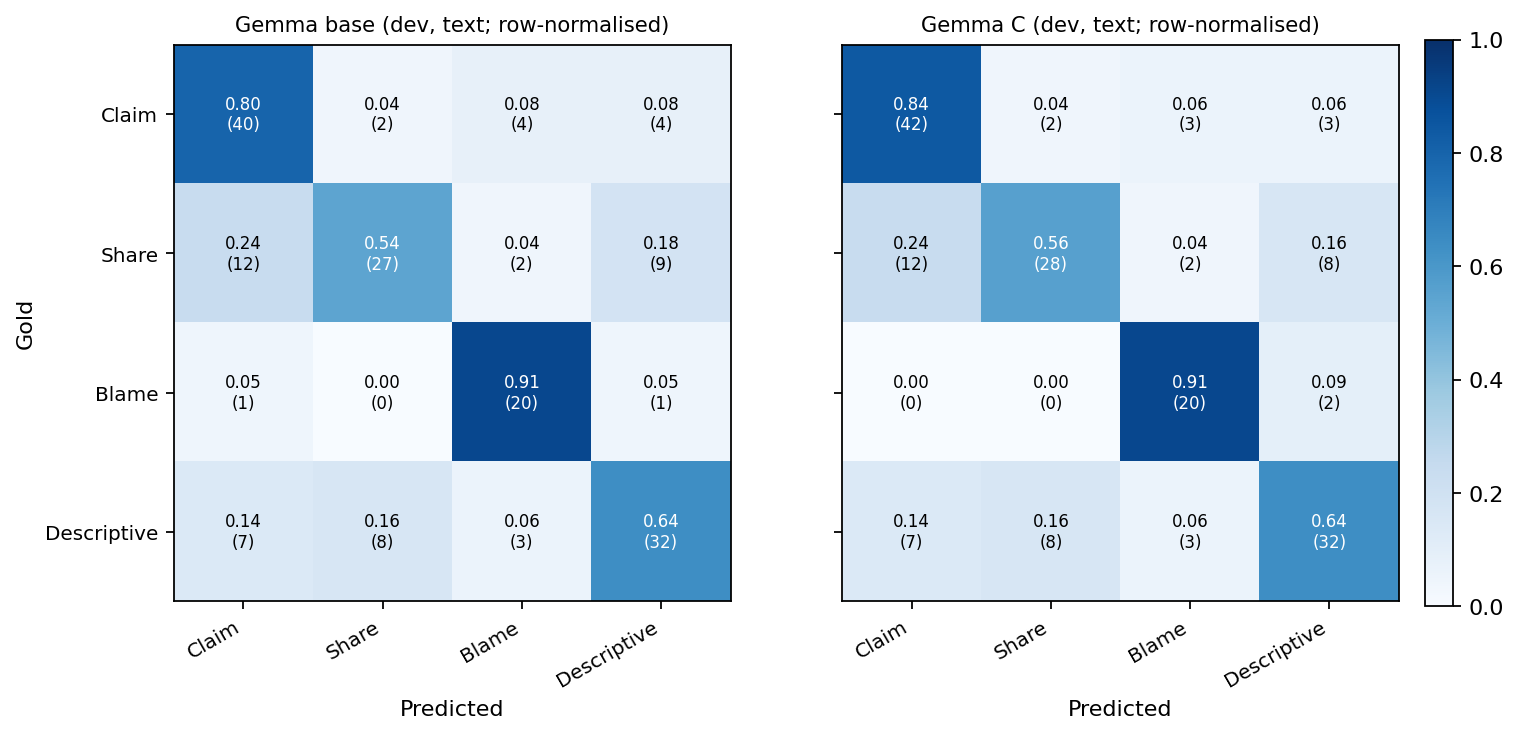}
\caption{Confusion Matrices Across Two Codebook Versions}
\notes{Row-normalized confusion matrices for the Gemma annotator (Gemma) on $D_{\mathrm{dev}}$ under the original codebook $C^{(0)}$ (left) and the Llama-derived revision $C^{(1)}$ (right); document counts in parentheses. Rows denote reference labels and columns denote model
predictions; diagonal entries are class-specific recall.}
\label{fig:gemma-hunter-confusion}
\end{figure}

\paragraph{Problematic classes and common confusions.}
Figure~\ref{fig:gemma-hunter-confusion} shows the confusion matrices under the original and revised codebooks. Under the original codebook, reference \textsc{Credit-sharing} documents flow mainly to \textsc{Credit-claiming} ($12$) and \textsc{Descriptive} ($9$). These mismatches recur across models at the document level. Eight of the twelve \textsc{Credit-sharing}$\rightarrow$\textsc{Credit-claiming} documents and eight of the nine \textsc{Credit-sharing}$\rightarrow$\textsc{Descriptive} documents receive
the same wrong label from the Llama annotator, whose own counts in those two flows are $9$ and $11$. This document-level recurrence points to material shared across the annotators---the written codebook and/or the reference labels---rather than to a purely model-specific error. The \textsc{Descriptive}$\rightarrow$\textsc{Credit-sharing} inflow, by contrast, overlaps on only three of eight documents, a model-dependent pattern. The right panel of Figure~\ref{fig:gemma-hunter-confusion} shows that under the Llama-derived revision $C^{(1)}$ the Gemma annotator's \textsc{Credit-sharing} row is essentially unchanged, which motivates a separate Gemma-specific candidate revision.

\paragraph{Why the nominations differ.}
The two models identify the same difficult boundary region connecting \textsc{Credit-sharing}, \textsc{Credit-claiming}, and \textsc{Descriptive}, but differ in which class absorbs the errors. The BFRS defect is visible in the written text. The \textsc{other} entry states its assignment conditions, and those conditions are under-specified, so the audit can point to the defective sentences and rewrite them. 

The credit-claiming defect is different in kind. The codebook defines each category by the target of the attribution and illustrates typical cases, but it does not state the test for recognizing a credit attribution in the first place, whether describing a favorable outcome the speaker helped produce counts as claiming credit, and whether jointly framed achievements credit the speaker or the collective. The written text leaves that judgment unspecified, and the comparison suggests that the two models resolve it differently \parencite{casanova2026}.

The Llama annotator treats attribution statements as \textsc{Descriptive} unless credit is explicit, applying the fallback class too broadly; the Gemma annotator treats jointly claimed credit as self-credit,  missing \textsc{Credit-sharing} documents. When the codebook leaves a judgment unstated rather than stating a condition inconsistently, different models can resolve that ambiguity differently. Nomination is therefore model-conditional in the credit-claiming case, whereas it is model-stable in BFRS.

\begin{figure}[!htbp]
    \centering
    \includegraphics[width=1\linewidth,trim = 0cm 0cm 0cm 0cm, clip=true]{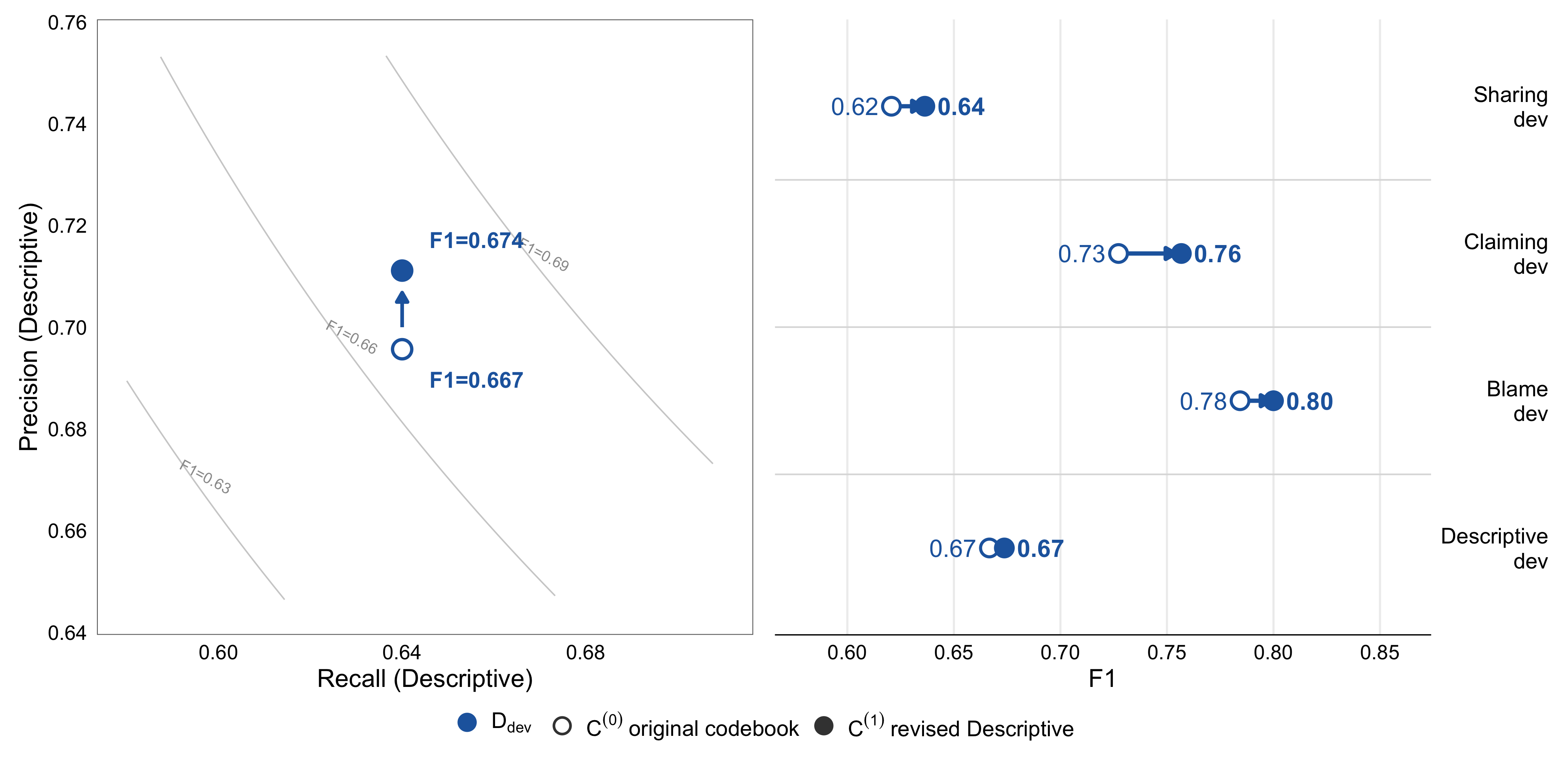}
    \caption{Effects of Revision, $D_{\mathrm{dev}}$.}
    \notes{ $D_{\mathrm{dev}}$ ($n=172$). Open and filled points denote the original codebook  ($C^{(0)}$) and the revised \textsc{Descriptive} codebook ($C^{(1)}$).
    Left panel: precision and recall for contours. The revision raises \textsc{Descriptive}  precision with recall unchanged, the designed narrowing, in weaker form than on the Llama annotator. Right panel: F1 for all four classes across the two codebooks.}
\label{fig:gemma-hunter-decomp}
\end{figure}

\paragraph{Held-out evaluation.}
The Llama annotator's revision $C^{(1)}$ transfers in direction: on $D_{\mathrm{dev}}$ the Gemma annotator's macro F1 rises $0.700\rightarrow0.717$ and every class moves the way it moves on the Llama annotator (Figure~\ref{fig:gemma-hunter-decomp}; Table~\ref{tab:gemma-hunter-dev}).

\begin{table}[!htbp]
\centering
\small
\caption{Credit-claiming per-class precision, recall, and F1, original codebook against revision $C^{(1)}$, $D_{\mathrm{dev}}$ ($n=172$)}
\label{tab:gemma-hunter-dev}
\resizebox{\textwidth}{!}{%
\begin{tabular}{lrrrrrrrl}
\toprule
 & & \multicolumn{3}{c}{$C^{(0)}$} & \multicolumn{3}{c}{$C^{(1)}$} & \\
\cmidrule(lr){3-5} \cmidrule(lr){6-8}
Class & $n$ & P & R & F1 & P & R & F1 & $\Delta$F1 ($95\%$ CI) \\
\midrule
Blame-shifting & 22 & 0.69 & 0.91 & 0.784 & 0.71 & 0.91 & 0.800 & $+0.016$ $[-0.060,+0.095]$ \\
Claiming & 50 & 0.67 & 0.80 & 0.727 & 0.69 & 0.84 & 0.757 & $+0.029$ $[-0.007,+0.071]$ \\
Descriptive & 50 & 0.70 & 0.64 & 0.667 & 0.71 & 0.64 & 0.674 & $+0.007$ $[-0.016,+0.033]$ \\
Sharing & 50 & 0.73 & 0.54 & 0.621 & 0.74 & 0.56 & 0.636 & $+0.016$ $[+0.000,+0.054]$ \\
\bottomrule
\end{tabular}%
}
\notes{Per-class metrics under the original and revised credit-claiming codebooks, sorted by $C^{(0)}$ F1. $\Delta$F1 is the paired document bootstrap of $C^{(1)}$ against $C^{(0)}$ on the same documents ($B=5000$), $95\%$ CI in brackets. The Gemma annotator reproduces the \emph{direction} of every class the Llama annotator's revision moved.}
\end{table}

Model-conditional diagnosis can also produce annotator-specific revision effects. Because Gemma's diagnosis centered on \textsc{Credit-sharing}, we constructed a separate candidate clarification, $C^{(2)}$, for joint-credit cases and evaluated the same wording with both annotators.

On the Gemma annotator's held-out validation sample, the candidate $C^{(2)}$ has the intended effect: \textsc{Credit-sharing} rises by $0.044$, \textsc{Credit-claiming} does not decrease ($+0.022$), and six of the nine recovered documents belong to the diagnosed \textsc{Credit-claiming}$\rightarrow$\textsc{Credit-sharing} pattern. Applied to the Llama annotator, however, the same wording moves performance in the opposite direction: \textsc{Credit-sharing} falls by $0.029$, \textsc{Credit-claiming} by $0.054$, and macro F1 by $0.035$ $[-0.061,-0.013]$ (Table~\ref{tab:hunter-cs}). This analysis is a cross-annotator transfer test rather than an additional stage of the Llama revision sequence. The clause was written to counteract Gemma's tendency to read jointly claimed credit as self-credit; Llama did not show the same tendency and instead interpreted the added wording more broadly, creating new \textsc{Credit-sharing} errors. A clarification motivated by one annotator's reading of an underspecified boundary therefore need not transfer to another. Candidate revisions should be evaluated on the codebook--annotator system in which they are intended to be used.

\begin{table}[!htbp]
\centering
\small
\caption{Credit-claiming revision $C^{(1)}$ against the original codebook $C^{(0)}$, and the candidate revision $C^{(2)}$ against its baseline $C^{(1)}$, both models}
\label{tab:hunter-cs}
\resizebox{\textwidth}{!}{%
\begin{tabular}{lllll}
\toprule
F1 & $C^{(0)}$ & $C^{(1)}$ ($\Delta$ vs.\ $C^{(0)}$) & \textsc{$C^{(2)}$} ($\Delta$ vs.\ $C^{(1)}$) & $\Delta$ \textsc{$C^{(2)}$} vs.\ $C^{(0)}$ \\
\midrule
\multicolumn{5}{l}{\emph{Llama-3.3-70B (deployment)}}\\
\multicolumn{5}{l}{Dev ($n=172$)}\\
\quad Sharing & $0.674$ & $0.750$ ($\mathbf{+0.077}$ $[+0.015,+0.150]$) & $0.764$ ($+0.014$ $[-0.018,+0.055]$) & $\mathbf{+0.091}$ $[+0.026,+0.168]$ \\
\quad Claiming & $0.735$ & $0.747$ ($+0.013$ $[-0.030,+0.057]$) & $0.755$ ($+0.008$ $[+0.000,+0.026]$) & $+0.021$ $[-0.024,+0.070]$ \\
\quad Blame-shifting & $0.750$ & $0.750$ ($+0.000$ $[+0.000,+0.000]$) & $0.810$ ($+0.061$ $[+0.000,+0.165]$) & $+0.061$ $[+0.000,+0.165]$ \\
\quad Descriptive & $0.683$ & $0.684$ ($+0.001$ $[-0.051,+0.052]$) & $0.696$ ($+0.012$ $[-0.023,+0.049]$) & $+0.013$ $[-0.042,+0.068]$ \\
\quad macro & $0.711$ & $0.733$ ($+0.023$ $[-0.008,+0.055]$) & $0.756$ ($+0.024$ $[-0.004,+0.059]$) & $\mathbf{+0.046}$ $[+0.007,+0.090]$ \\
\multicolumn{5}{l}{Held-out ($n=321$, ${\sim}100$/class; Blame-shifting $21$)}\\
\quad Sharing & $0.589$ & $0.653$ ($\mathbf{+0.064}$ $[+0.017,+0.115]$) & $0.624$ ($\mathbf{-0.029}$ $[-0.061,-0.001]$) & $+0.035$ $[-0.014,+0.085]$ \\
\quad Claiming & $0.709$ & $0.748$ ($+0.038$ $[+0.000,+0.079]$) & $0.693$ ($\mathbf{-0.054}$ $[-0.096,-0.018]$) & $-0.016$ $[-0.052,+0.017]$ \\
\quad Blame-shifting & $0.682$ & $0.711$ ($+0.030$ $[+0.000,+0.101]$) & $0.682$ ($-0.030$ $[-0.101,+0.000]$) & $+0.000$ $[+0.000,+0.000]$ \\
\quad Descriptive & $0.623$ & $0.605$ ($-0.018$ $[-0.063,+0.023]$) & $0.579$ ($-0.027$ $[-0.059,+0.004]$) & $\mathbf{-0.045}$ $[-0.092,-0.003]$ \\
\quad macro & $0.651$ & $0.679$ ($\mathbf{+0.029}$ $[+0.000,+0.060]$) & $0.644$ ($\mathbf{-0.035}$ $[-0.061,-0.013]$) & $-0.007$ $[-0.031,+0.018]$ \\
\midrule
\multicolumn{5}{l}{\emph{Gemma (second annotator)}}\\
\multicolumn{5}{l}{Dev ($n=172$)}\\
\quad Sharing & $0.621$ & $0.636$ ($+0.016$ $[+0.000,+0.054]$) & $0.637$ ($+0.001$ $[-0.053,+0.060]$) & $+0.017$ $[-0.045,+0.084]$ \\
\quad Claiming & $0.727$ & $0.757$ ($+0.029$ $[-0.007,+0.071]$) & $0.712$ ($-0.045$ $[-0.108,+0.010]$) & $-0.016$ $[-0.066,+0.028]$ \\
\quad Blame-shifting & $0.784$ & $0.800$ ($+0.015$ $[-0.060,+0.095]$) & $0.745$ ($-0.056$ $[-0.134,+0.000]$) & $-0.040$ $[-0.108,+0.000]$ \\
\quad Descriptive & $0.667$ & $0.674$ ($+0.007$ $[-0.016,+0.033]$) & $0.653$ ($-0.021$ $[-0.047,+0.000]$) & $-0.014$ $[-0.036,+0.000]$ \\
\quad macro & $0.700$ & $0.717$ ($+0.017$ $[-0.011,+0.047]$) & $0.687$ ($-0.030$ $[-0.066,+0.003]$) & $-0.013$ $[-0.046,+0.017]$ \\
\multicolumn{5}{l}{Held-out ($n=321$, ${\sim}100$/class; Blame-shifting $21$)}\\
\quad Sharing & $0.607$ & $0.601$ ($-0.006$ $[-0.037,+0.027]$) & $0.645$ ($\mathbf{+0.044}$ $[+0.005,+0.090]$) & $+0.038$ $[-0.005,+0.087]$ \\
\quad Claiming & $0.753$ & $0.750$ ($-0.003$ $[-0.015,+0.007]$) & $0.772$ ($+0.022$ $[-0.004,+0.047]$) & $+0.019$ $[-0.006,+0.043]$ \\
\quad Blame-shifting & $0.630$ & $0.642$ ($+0.012$ $[+0.000,+0.040]$) & $0.642$ ($+0.000$ $[+0.000,+0.000]$) & $+0.012$ $[+0.000,+0.040]$ \\
\quad Descriptive & $0.641$ & $0.625$ ($-0.016$ $[-0.053,+0.017]$) & $0.629$ ($+0.005$ $[-0.024,+0.035]$) & $-0.012$ $[-0.047,+0.022]$ \\
\quad macro & $0.658$ & $0.654$ ($-0.004$ $[-0.021,+0.015]$) & $0.672$ ($+0.018$ $[-0.000,+0.036]$) & $+0.014$ $[-0.006,+0.037]$ \\
\bottomrule
\end{tabular}%
}
\notes{Paired with-replacement bootstrap $\Delta$ ($B=5000$) of $C^{(1)}$ against $C^{(0)}$, $C^{(2)}$ against $C^{(1)}$ and $C^{(2)}$ against $C^{(0)}$ on the same documents, $95\%$ CI in brackets, \textbf{bold} where the CI clears zero.}
\end{table}

\end{document}